\documentclass[lettersize,journal]{IEEEtran}
\usepackage{amsmath,amsfonts}
\usepackage{algorithmic}
\usepackage{array}
\usepackage[caption=false,font=normalsize,labelfont=sf,textfont=sf]{subfig}
\usepackage{textcomp}
\usepackage{stfloats}
\usepackage{url}
\usepackage{verbatim}
\usepackage{graphicx}
\usepackage{balance}
\usepackage{caption}
\usepackage{subfig} 
\usepackage{tabularx}

\usepackage{hyperref}

\usepackage{color,soul}
\usepackage{float}
\usepackage{booktabs}
\usepackage{multirow}
\usepackage{color}
\usepackage[usenames,dvipsnames,svgnames,table]{xcolor}
\usepackage{listings}
\hypersetup{
    colorlinks=true,
    filecolor=magenta,      
    urlcolor=magenta,
}

\usepackage{array}
\newcolumntype{L}{>{\RaggedRight\arraybackslash}m{3cm}}
\begin{document}

% \title{Measuring Driver Cognitive Load with EEG with Additional Modalities}

% \title{Measuring Driver Cognitive Load with EEG and Additional Wearable Signals: Dataset and Baselines}

% \title{Measuring Driver Cognitive Load with EEG and Auxillary Wearable Signals}

% \title{Multimodal Brain-Computer Interface for In-Vehicle Driver Cognitive Load Measurement}

\title{Multimodal Brain-Computer Interface for In-Vehicle Driver Cognitive Load Measurement: Dataset and Baselines}


% \author{Prithila Angkan, Anubhav Bhatti, Behnam Behinaein, Zunayed Mahmud, Dirk Rodenburg, Heather Braund, Jim Mclellan, Aaron Ruberto, Geoffery Harris, Daryl Wilson, Adam Szulewski, Dan Howes,\\Ali Etemad,~\IEEEmembership{Senior Member,~IEEE}, Paul Hungler}

% \author{Prithila Angkan, Ali Etemad,~\IEEEmembership{Senior Member,~IEEE}, Behnam Behinaein, Zunayed Mahmud, Anubhav Bhatti, Dirk Rodenburg, Heather Braund, James Mclellan, Aaron Ruberto, Geoffrey W. Harris, Daryl Wilson, Adam Szulewski, Dan Howes, Paul Hungler}

\author{Prithila Angkan, Behnam Behinaein, Zunayed Mahmud, Anubhav Bhatti, Dirk Rodenburg, Paul Hungler, Ali Etemad,~\IEEEmembership{Senior Member,~IEEE}}

% Prithila Angkan, Anubhav Bhatti, Behnam Behinaein, Zunayed Mahmud,Dirk Rodenburg, Heather Braund, Jim Mclellan, Aaron Ruberto, Geoffery Harris, Daryl Wilson, Adam Szulewski, Dan Howes, Paul Hungler, Ali Etemad, Senior Member, IEEE

% \title{Measuring Driver Cognitive Load with Multimodal Brain-Computer Interface}

% \title{Measuring Driver Cognitive Load with Multimodal Physiological Signal Representation Learning}

% \markboth{IEEE Transactions on Intelligent Vehicles}%
% {How to Use the IEEEtran \LaTeX \ Templates}

% \markboth{}

\maketitle

\begin{abstract}
Through this paper, we introduce a novel driver cognitive load assessment dataset, CL-Drive, which contains Electroencephalogram (EEG) signals along with other physiological signals such as Electrocardiography (ECG) and Electrodermal Activity (EDA) as well as eye tracking data. The data was collected from 21 subjects while driving in an immersive vehicle simulator, in various driving conditions, to induce different levels of cognitive load in the subjects. The tasks consisted of 9 complexity levels for 3 minutes each. Each driver reported their subjective cognitive load every 10 seconds throughout the experiment. The dataset contains the subjective cognitive load recorded as ground truth. In this paper, we also provide benchmark classification results for different machine learning and deep learning models for both binary and ternary label distributions. We followed 2 evaluation criteria namely 10-fold and leave-one-subject-out (LOSO). We have trained our models on both hand-crafted features as well as on raw data. We make our dataset public to contribute to the field.  
\end{abstract}


\section{Introduction}
% According to Transport Canada’s National Collision Database (NCDB), there were 1,745 fatal accident cases in Canada. 
A large number of accidents and collisions occur on the roads every year. While many of these accidents are caused by distracted drivers, for instance due to distraction or drowsiness \cite{miyaji2009driver}. Distraction, meanwhile, can be caused by a number of personal or ambient factors, including high cognitive load due to engagement with secondary tasks \cite{palinko2010estimating, fridman2018cognitive, yuce2016action}. 
\textcolor{black}{
Over the past several years, much research has been conducted to investigate the effects of cognitive load and cognitive fatigue. Studies have demonstrated that prolonged engagement in cognitively demanding tasks may lead to cognitive fatigue, a condition that could pose risks \cite{wu2020nonparametric, wu2021scalable}.
% Over the past years, many research has been done to investigate the effect of cognitive load as well as cognitive fatigue. Research has shown that engaging in a cognitively demanding task for a long period of time might cause cognitive fatigue \cite{wu2020nonparametric, wu2021scalable} that can be dangerous.
}
Cognitive load refers to the quantity of information our working memory can process at a given time. In other words, it is the amount of cognitive resources required to accomplish a task. In general, two categories of cognitive load, intrinsic and extraneous, have been described in the literature \cite{sweller2011cognitive}. While intrinsic cognitive load is defined as the inherent complexity of a given task, extraneous cognitive load refers to the cognitive resources demanded by environmental factors that are task-irrelevant \cite{sweller2011cognitive}. The success or failure in performance toward a particular task, on the other hand, is influenced by the amount of cognitive load experienced by the person performing the task
% The cognitive load is directly related to the task performance 
\cite{das2014cognitive}. 
If the cognitive load increases beyond a certain point, the individual's performance will degrade, which in case of driving may increase the likelihood of road accidents.

% \cite{iqbal2004understanding,iqbal2004task,stuyven2000effect}. 

In order to reduce the number of road accidents caused by high cognitive load, recent intelligent technologies integrated into vehicles should possess the ability to measure cognitive load and alarm the user should dangerously high amounts of it be detected. Brain-computer interfaces (BCI) have recently gained traction in providing advanced means of communication between humans and machines. In particular, head-worn Electroencephalogram (EEG) devices allow for non-invasive yet accurate human-machine interactions. To this end, machine learning and deep learning techniques can be used to learn from datasets with various types of driver-related signals (including EEG). Additionally, these datasets require quantitative cognitive load scores to be measured and provided at frequent intervals, so that they could be used to train the machine learning models. While a number of relevant datasets have been collected and published in recent years \cite{schneegass2013data, mijic2019mmod, markova2019clas, gjoreski2020datasets, kalatzis2021database}, a number of problems persist. 
% \textcolor{black}{
First, while a number of datasets for cognitive load do exist, they have often been captured in non-vehicle scenarios. In fact, to our knowledge, only \cite{schneegass2013data} has studied cognitive load in the context of driving.% fridman2018cognitive 
Second, the cognitive load ground-truth scores in most existing datasets are generally sparse, and have been measured several minutes apart or upon task completion \cite{gjoreski2020datasets,kalatzis2021database}. This in turn makes training of machine learning models more difficult and less accurate. 
Third, while most existing datasets on cognitive load are in fact `multimodal', the notion of BCI with auxiliary wearable signals has not been widely explored 
% do not contain a diverse set of signals obtained from the drivers. While multimodal datasets for cognitive load have been popular, a few informative modalities, namely EEG and gaze have not been widely explored
\cite{schneegass2013data, mijic2019mmod, markova2019clas, gjoreski2020datasets, kalatzis2021database}. 
Lastly, in most existing works in the area, the focus has been solely on cognitive load or distraction caused by task-irrelevant activities, overlooking the fact that performing the main task itself (in our case, driving) can be a strong source of high cognitive load.
% }

% 1- labels are generally sparse, few minutes apart
% 2- lack of multimodal datasets
% 3- having distraction... task-irrelevant factors

% While advances in machine learning and deep learning techniques  allow us to utilize EEG data along with other physiological signals to assess cognitive load, the lack of multimodal cognitive load datasets including EEG has created a barrier in research progress. Moreover, most of the datasets for cognitive load lack EEG as well gaze data, despite these modalities being proven to be directly affected by changes in cognitive load \cite{schneegass2013data, mijic2019mmod, markova2019clas, gjoreski2020datasets, kalatzis2021database}. Another drawback that some of the current datasets have is that they do not provide frequent subjective ratings or take the ratings at the end of the session \cite{schneegass2013data}. Subjective ratings can vary every few seconds so it is important to take frequent ratings in order to capture changes in cognitive load.

In this paper, we introduce a novel driver cognitive load assessment dataset containing EEG signals along with other physiological signals such as Electrocardiography (ECG) and Electrodermal Activity (EDA) as well as eye tracking data. This dataset, which we name CL-Drive, is collected from 21 subjects while driving in an immersive vehicle simulator in diverse situations capable of inducing various levels of cognitive load in the subjects. Each subject performs driving tasks in 9 complexity levels for 3 minutes each and reports their subjective cognitive load every 10 seconds throughout the experiment as ground-truth cognitive load labels. 
% The dataset contains the subjective cognitive load recorded as ground truth. 
% Next, we perform a detailed analysis of 
In this paper, we also provide benchmark classification results for different machine learning and deep learning models. Both raw signals as well as popular features supported by the literature have been used as inputs.
% for both binary and ternary label distributions. 
We follow two important evaluation criteria, namely 10-fold and leave-one-subject-out (LOSO). Our benchmarking demonstrates that cognitive load induced by driving can be measured with reasonable accuracy using EEG and auxiliary wearable signals.
%We have trained our models on both hand-crafted features as well as on raw data.   

% \textcolor{black}{
Our contributions in this paper are summarized as follows:
\begin{itemize}
	\item We collect and release a dataset, CL-Drive, that can allow researchers to evaluate driving-induced cognitive load, which can be useful for developing automated alarm systems for intelligent vehicles.  
% 	to understand variations in intrinsic cognitive load.
	\item CL-Drive provides data from various modalities including EEG, ECG, EDA and Gaze, which is a rich source for training machine learning systems capable of performing cognitive load assessment. To the best of our knowledge, this is the first and only dataset to collect driver cognitive load ratings along with bio-signals.
	\item CL-Drive contains dense and frequent subjective ratings which are spread only 10 seconds, allowing for more reliable and frequent automated cognitive load measurement by learned models.
\end{itemize}
% }

The rest of this paper is summarized as follows. In Section \ref{section:related_work}, we first provide a study of cognitive load followed by an overview of the publicly available cognitive load datasets that contain physiological signals. Section \ref{section:experiment_setup} explains the experimental setup, including sensor configurations, driving simulator details, cognitive load assessment, and data collection protocol. Next, we discuss the data pre-processing, feature extraction, normalization, and baseline classifiers in section \ref{section:data_processing}. Lastly, in Section \ref{section:results} we provide the results and discussions.

\section{Related Work} \label{section:related_work}

\subsection{Cognitive Load Measurement}

Prior research has shown that measuring cognitive load from physiological signals \cite{sarkar2019classification, ross2019toward} continues to be a challenging task \cite{brunken2003direct, mayer2002aids}. 
% Some prior works have classified high vs. low cognitive load using physiological signals \cite{sarkar2019classification, ross2019toward}.
There are both subjective and objective measures that are commonly used to evaluate cognitive load levels that involve: (\textit{i}) self-reporting, (\textit{ii}) dual-task measures, and (\textit{iii}) physiological measures \cite{klepsch2017development}. The PAAS scale \cite{paas1992training}, shown in Table \ref{table_paas_level} is most commonly used for self-reported subjective cognitive load labels. The National Aeronautics and Space Association Task Load Index (NASA-TLX) \cite{hart1988development} is also commonly used as a self-reporting tool. Dual-task measurement involves the individual performing two tasks at the same time. One way of designing this is to measure knowledge gain from one task and response time for the other task \cite{brunken2002assessment}. In \cite{park2015rhythm}, another way of implementing dual-task measurement was explored, which was by performing a continuous secondary task while learning the primary task. There are several physiological parameters that have also been used as cognitive load measures in the past. This includes variation in pupil diameter and blink rate \cite{van2004memory, chen2014using}, heart rate variability \cite{paas1994variability}, and electrocardiogram (ECG) \cite{antonenko2010using} to name a few.

\begin{table}[!t]
\caption{PAAS subjective cognitive load scores used in this study.}
\centering
\begin{tabular}{l l}
\hline
\textbf{PAAS Subjective} & \multirow{2}{*}{\textbf{Description}} \\
\textbf{Cognitive Load Scores} &  \\
 \hline \hline
1 & Very, very low \\ 
2 & Very low \\
3 & Low \\
4 & Rather low\\
5 & Neither low nor high\\
6 & High\\
7 & Rather high\\
8 & Very high\\
9 & Very, very high\\
\hline
\end{tabular}
\label{table_paas_level}
\end{table}
% }

\subsection{Cognitive Load in Driving}

In the area of driving, prior works have studied cognitive load mainly in the context of the driver being engaged by secondary tasks such as using mobile phones or performing some other in-vehicle activities \cite{palinko2010estimating, chisholm2008effects, he2019high, fridman2018cognitive, barua2017classifying}. In \cite{palinko2010estimating}, the cognitive load of drivers was measured when the drivers were involved in verbal conversation and word games while driving. As a result, the cognitive load induced was due to the combination of both primary as well as secondary tasks. A remote eye tracker was used to measure the pupil size of all 32 participants which in turn was used to estimate the cognitive load of the participants.
% This experiment was carried out in a real life driving environment while the participants were engaged in playing word games verbally during driving. 
The ground truth was the performance measures which they calculated using lane position and degree of rotation of the steering wheel, and subsequently evaluated the relationship between the change in pupil diameter and driving performance.

\textcolor{black}{
In \cite{yusof2022reading}, the non-driving task of reading was performed by 18 participants in a fully automated vehicle. Two peripheral information systems, one utilizing the visual modality and the other the haptic modality, were assessed to examine their impact on situational awareness, mental workload, viewing behavior, and reading performance. In
\cite{jiang2022understanding}, a three-phase framework was proposed that allows effective diagnosis of driver's visual and comprehension loads in traffic scenes. Drivers from diverse backgrounds were assessed for visual and comprehension load while driving simulations across various traffic scenarios. 
% and determined to indicate visual load and comprehension load. 
Another paper, \cite{ayoub2022predicting}, studied the effect of safe takeover transition in conditionally automated driving and used XGBoost to evaluate their work using a dataset from a meta-analysis study.
}

Driving performance while interacting with a portable music player
% an iPod\footnote{https://www.apple.com/ca/}
% during first session vs. sixth session 
was evaluated in \cite{chisholm2008effects}. 
It was observed during a multi-session setup, that the cognitive load of the participants during the first sessions was higher, hence the driving performance (e.g., perception response time (PRT) while braking and overall control of the vehicle), was lower in comparison to the later sessions. 
The experiment was carried out on 19 participants using simulated vehicle. 
%secondary task performance were measures were used as ground truth. 
Next, in \cite{he2019high}, the high cognitive load of drivers was evaluated using EEG signals in 3 different driving conditions, namely: no secondary task (baseline), low cognitive load task, and high cognitive load task. The low and high cognitive load tasks were based on N-back tasks used in \cite{mehler2009impact, mehler2012sensitivity}. GSR, eye tracking, respiration rate (RR), and accelerator release time (ART) data were collected during the experiments. The data was collected from 37 participants in a vehicle simulator. The NASA-TLX was used to collect the participants subjective rating at the end of the experiment. In \cite{fridman2018cognitive}, the cognitive load of participants was evaluated using eye video data extracted from facial videos during driving. Three different N-back tasks were used as secondary tasks, which were also used to quantify the ground truth levels. Hidden Markov models and 3D-CNN were then used to evaluate the result. In another paper \cite{barua2017classifying}, the cognitive load of participants was evaluated while driving and performing a 1-back task. EEG data was collected from 36 participants. To evaluate the performance, case-based reasoning classifiers were used \cite{kolodner1992introduction}. A few other prior works such as \cite{palinko2010estimating, chisholm2008effects, he2019high} have used more simple approaches based on predefined metrics (e.g., required time to break, degree of motion of the steering wheel, etc.) to measure cognitive load from physiological signals. 
% The data was collected from 92 subjects while driving real car.
% In addition to video data, ECG, EDA, and audio data were also used. 

% In \cite{fridman2018cognitive}, the cognitive load of the participants was evaluated using eye video data extracted from facial video data during driving. 0-back, 1-back and 2-back tasks were used as secondary task while driving as well as the ground truth levels. Data was collected from 92 subjects while driving real car. In addition to video data, ECG, EDA and audio data were also recorded. 

Besides cognitive load, other factors such as driver emotions \cite{braun2019improving, nass2005improving, zepf2020driver} and vigilance \cite{zhang2021capsule,bergasa2006real, lin2014wireless}, have been widely studies in the literature. While these works may maintain some similarities to works on cognitive load, they are in fact different driver attributes which are outside the scope of this study. Moreover, the notions of affect and distraction have been more widely studied for drivers, as opposed to cognitive load which is a less explored area.

% Driver vigilance estimation has also been explored in the past with the ultimate goal of preventing accidents \cite{zhang2021capsule}. 
% ...\\...\\...\\...\\...\\...\\...\\...\\...\\...\\...\\...

% Although, cognitive load imposed by secondary tasks is important to evaluate, it is equally important to investigate the cognitive load induced by the primary task. 

% Having prior knowledge and experience about a certain task has been shown to reduce cognitive load, which means novices will experience higher intrinsic cognitive load compared to experts while performing the same task \cite{leppink2014effects}. Expert vs. novice cognitive load assessment strategies have been investigated in the past to portray the importance of adaptive simulated learning \cite{sarkar2019classification, ross2019toward, ruberto2021future}. In \cite{lavie2004load}, it is explained that tasks with a high perceptual load occupy more cognitive resources, hence leaving less room for processing irrelevant tasks. Therefore it is important to balance the cognitive load demand of the primary task to ensure less distraction as well as better task performance. 

{\renewcommand{\arraystretch}{1.4}
\begin{table*}[!t]
    % \footnotesize
    \centering
    \caption{Existing datasets in the literature that study cognitive load using physiological signals.}
    \begin{tabularx}{\textwidth}{p{0.15\textwidth}llXXXX}
    \hline
     \textbf{Dataset} & \textbf{Year} & \textbf{Sub.} & \textbf{Mental State} & \textbf{Modalities}   & \textbf{Stimuli} \\
        \hline\hline
        % Driver Workload \cite{schneegass2013data} & 2013 & 10 & Mental workload of the driver & ECG, BTemp, SCR & Watching driving videos, Driving in real environment \\
        % \hline 
        Driver Workload \cite{schneegass2013data} & 2013 & 10 & Mental workload of the driver & ECG, BTemp, SCR & Watching driving videos, Driving in real environment \\
        \hline
        % \cite{fridman2018cognitive} & 2018 & 92 & Cognitive load of driver & Video (eye), ECG, EDA, audio data & Different cognitive load tasks (0-back, 2-back and 3-back tasks)\\
        % \hline
        MMOD-COG \cite{mijic2019mmod}  & 2019 & 40   & Cognitive load & ECG, EDA,  Speech  & Arithmetic, Reading \\
        \hline
        CLAS \cite{markova2019clas} & 2019 & 62  & Cognitive Load, negative emotion and mental stress & ECG, PPG and EDA & Math problems, Logic problems and Stroop test \\
        \hline
        CogLoad \cite{gjoreski2020datasets}  & 2020 & 23  &  Cognitive load, Personality traits & heart rate, beat to beat interval, EDA, ST, and ACC & Different cognitive load tasks (2-back and 3-back tasks, visual cue task etc.) \\
        \hline
        Snake \cite{gjoreski2020datasets}  & 2020 & 23 & Cognitive load & heart rate, beat to beat interval, EDA, ST, and ACC  & Snake game on a smartphone \\
        \hline
        Kalatzis et al. \cite{kalatzis2021database}  & 2021 & 26 & Cognitive load & ECG, RR & MATB-II \\
        \hline
        \textcolor{black}{CL-Drive (ours)} & \textcolor{black}{2023} & \textcolor{black}{21} & \textcolor{black}{Cognitive load} & \textcolor{black}{EEG, ECG, EDA, Gaze} & \textcolor{black}{Driving a simulated vehicle in scenarios with various complexity levels}\\
        \hline
    \end{tabularx}
    \label{table:public_datasets}
\end{table*}
}

\subsection{BCI}

BCI systems can communicate the neural activities in the brain directly with an external device \cite{das2014cognitive, emami2020effects}. Research has shown that BCI can play a vital role in interpreting the cognitive load induced while driving \cite{das2014cognitive, emami2020effects}. This is due to the fact that the fronto-parietal brain regions along with sub-cortical regions can be engaged while experiencing varying amounts of cognitive load \cite{chai2018working}. EEG, is a non-invasive method which measures the potential difference caused by the electrical activity in the brain \cite{siuly2016eeg, klimesch1999eeg}. This property of EEG allows it to capture changes in brain activity while experiencing variations in cognitive load, which makes it a very good candidate for cognitive load evaluation. Multimodal approaches have proven effective at magnifying the accuracy of cognitive load assessment in the past \cite{ross2021unsupervised}. Apart from EEG, research has shown that both the sympathetic nervous system (SNS) which controls the skin conductance response and automatic nervous system which controls the heart rate variability (HRV) are impacted by cognitive load \cite{xiong2020pattern, hughes2019cardiac, johannessen2020psychophysiologic}. Prior research has also shown significant correlation between changes in pupil size, blink rate, saccade, and fixation with cognitive load \cite{zagermann2016measuring, iqbal2004understanding}.

\subsection{Public Cognitive Load Datasets}
Previous research has examined the induction of cognitive load in drivers with a subsequent evaluation of driving performance under varying cognitive load levels \cite{engstrom2017effects, sena2013studying}. Some other publications have studied cognitive load under a variety of  different experimental setups \cite{cabanero2019analysis, antonenko2010using}. 
There is evidence that affect and cognitive load are interrelated and affect has a significant impact on cognitive load \cite{volman2011anterior}. Though a wide range of studies have been done to study the impact of affect on EEG \cite{zhang2021deep, zhang2022parse}, the available datasets for cognitive load are indeed quite limited. In this section, we provide an overview of the publicly available datasets for cognitive load with physiological signals. Table \ref{table:public_datasets} presents a summary of these datasets.% the public datasets we will be examining. 

The Driver Workload dataset \cite{schneegass2013data}, provides multimodal data to evaluate driver workload using ECG, body temperature, and skin conductance response (SCR). The dataset was collected from 10 participants with the goal of evaluating cognitive load of drivers on different types of roads and in different driving environments. In addition to the physiological signals, two cameras were also used to record the driving route as well as the participant's facial videos. The video data were not made public for privacy purposes. The data was collected as the participants drove the car for 30 minutes. Participants subjective ratings were collected by watching videos of their own driving at the end of the activity. 

The MMOD-COG \cite{mijic2019mmod} dataset was recorded from 40 different subjects for cognitive load assessment during reading and arithmetic tasks. ECG and EDA were recorded from the subjects in addition to speech. The experiment was divided into reading and arithmatic segments where in the reading segment, two separate digits were shown for 5 seconds and repeated with different digits 20 times. The arithmetic segment was divided into high and low cognitive load levels and a total of 40 problems were asked to be solved by each participant. The CLAS dataset \cite{markova2019clas} was collected from 62 participants and was obtained by recording various cognitive load levels induced by tasks such as mathematics and logic problems as well as the Stroop test \cite{stroop1935studies}. In addition to cognitive load, audio-visual stimuli were used to induce emotional variations in the participants. Physiological data was collected using ECG, Plethysmography (PPG), EDA, and accelerometer (ACC) data. The next dataset, CogLoad \cite{gjoreski2020datasets} was also a multimodal dataset with 23 participants who performed 6 different computer tasks during the collection process. The physiological data collected was heart rate, beat to beat interval, EDA, skin temperature (ST), and ACC. The task was divided into two segments where the first segment involved understanding the participant's degree of cognitive resources and their personality traits using two N-back tasks \cite{gjoreski2020datasets, schmiedek2014task}. Whereas, in the second segment, 6 different tasks that required varying levels of cognitive load were performed. In addition, in order to completely occupy the cognitive resources of the participants, a secondary task was also given to them to perform. 

Another multimodal dataset consisting of heart rate, EDA, ST, and ACC was collected from 23 participants while they played the Snake game on a smartphone \cite{gjoreski2020datasets}. This dataset named Snake \cite{gjoreski2020datasets}, was collected with varying cognitive load levels where the amount of cognitive load experienced by the participants was controlled by the changing speed of the game. The task consisted of 3 complexity levels, high, medium and low, which lasted 2 minutes each. After the task completion, the participants answered the NASA TLX \cite{hart1988development} questionnaire along with two other 7-point Likert scale questions. Finally, Kalatzis et al. \cite{kalatzis2021database} presented a dataset that has been collected from 26 participants. In this dataset, the cognitive load of the participants was assessed using ECG and respiration rate (RR) data. High and low cognitive load data were collected as the participants used the MATB-II software \cite{santiago2011multi} while the NASA TLX \cite{hart1988development} questionnaire was used to collect ground truth values for the two cognitive load levels.

\textcolor{black}{In contrast to the above, our dataset considers driving to be the primary task and evaluates cognitive load induced while driving. Moreover, we record frequent subjective cognitive load scores which is not the case in existing cognitive load datasets. This allows us to perform more accurate evaluations and train better machine learning models. Finally, CL-Drive contains several modalities that enable multi-modal studies on cognitive load.}

\section{Experiment Setup and Data Collection}
% ???? give an overview in 1 paragraph
\label{section:experiment_setup}

In this section, we discuss the experimental protocol used in the study. This includes specifics on the setup for the sensors and driving simulator as well as details of the participants, diving scenarios, and cognitive load assessment.  

\subsection{Sensors}
\label{section:sensors}
\begin{figure*}
\centering
\subfloat[\small EEG headband]{\includegraphics[width=0.2\textwidth]{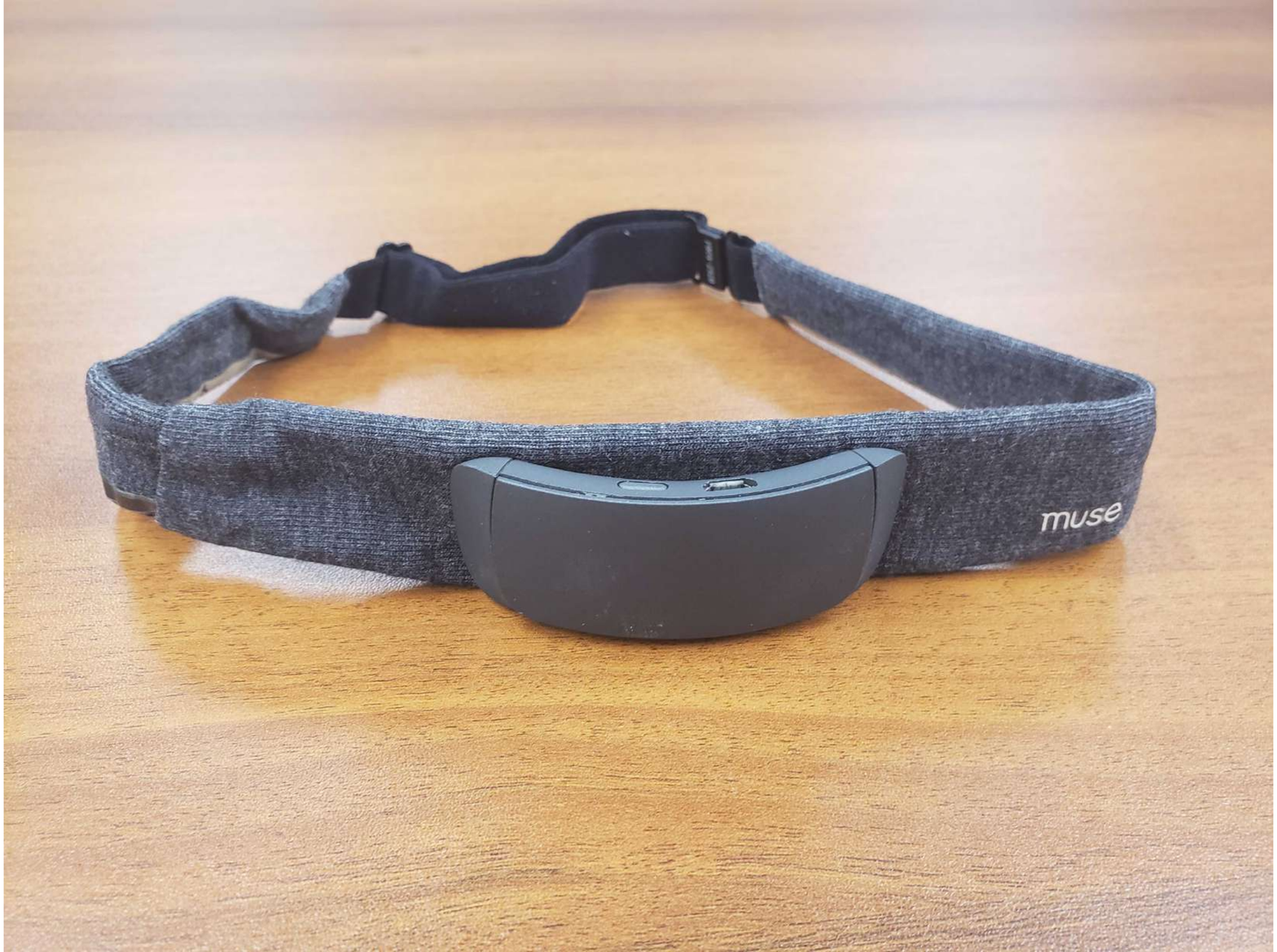}\label{fig:eeg_device}}\hskip1ex
\subfloat[\small ECG device]{\includegraphics[width=0.2\textwidth]{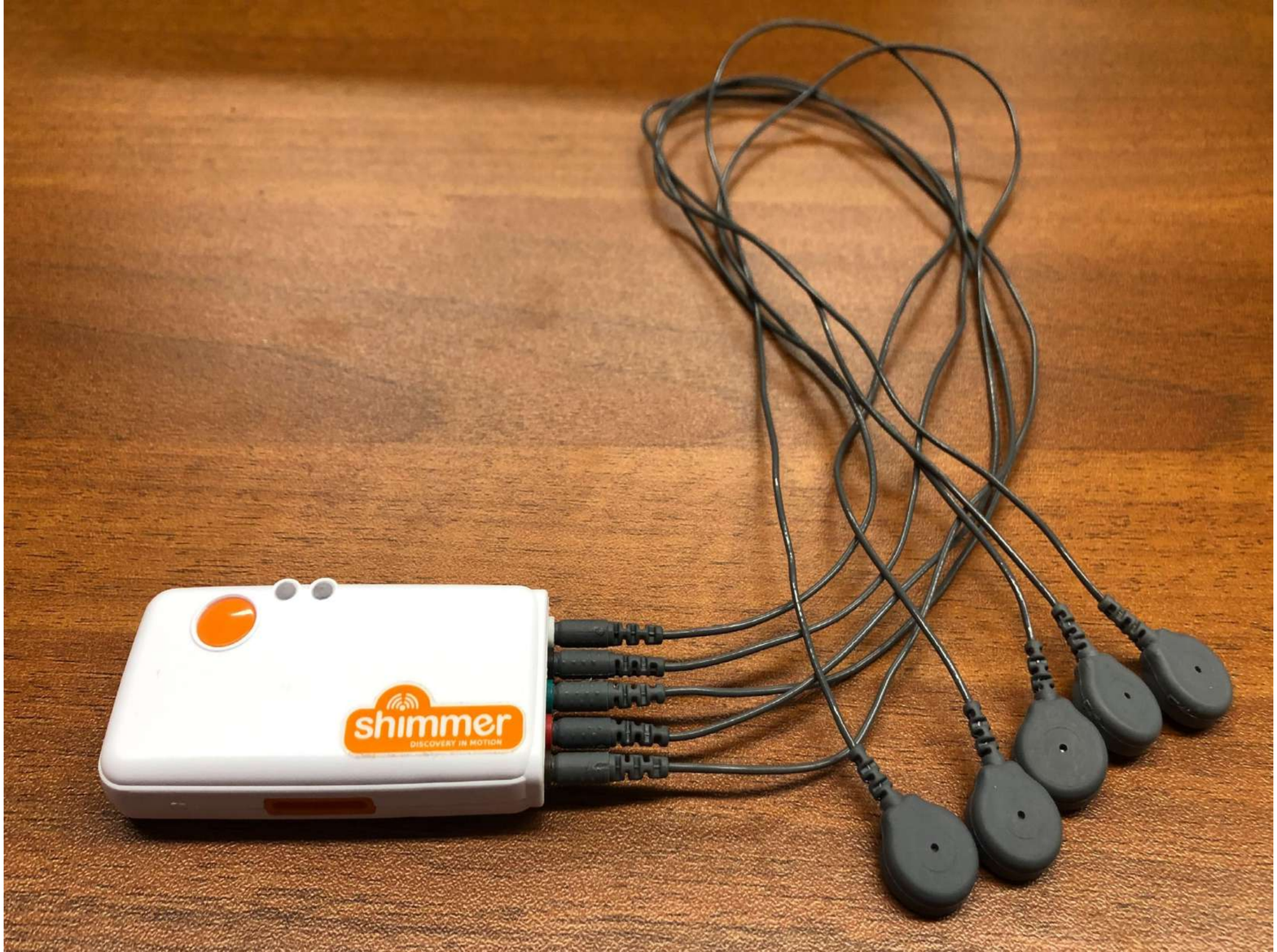}\label{fig:shimmer_ecg_pic}}\hskip1ex
\subfloat[\small EDA device]{\includegraphics[width=0.2\textwidth]{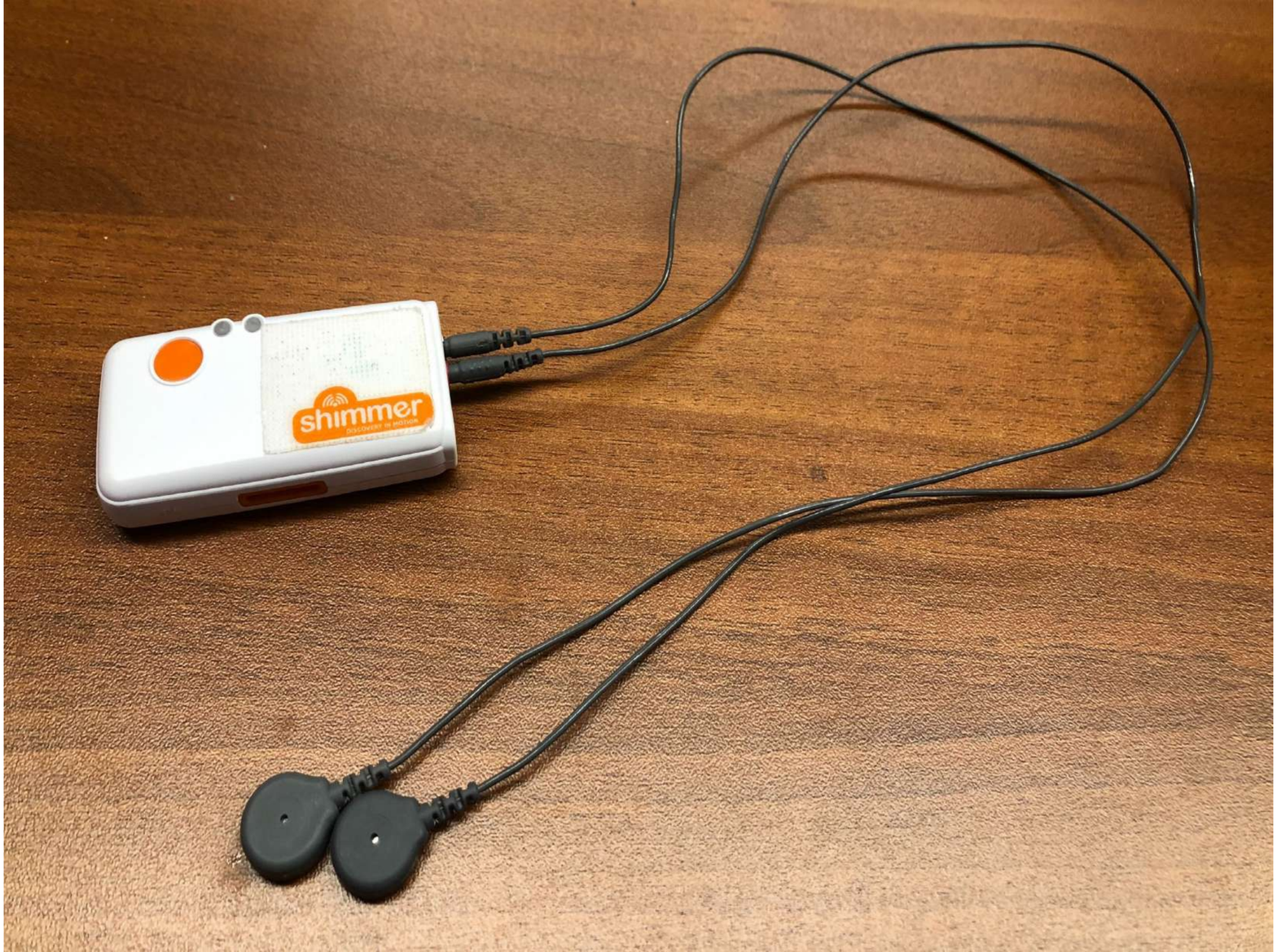}\label{fig:shimmer_eda_pic}}\hskip1ex
\subfloat[\small Eye tracker]{\includegraphics[width=0.2\textwidth]{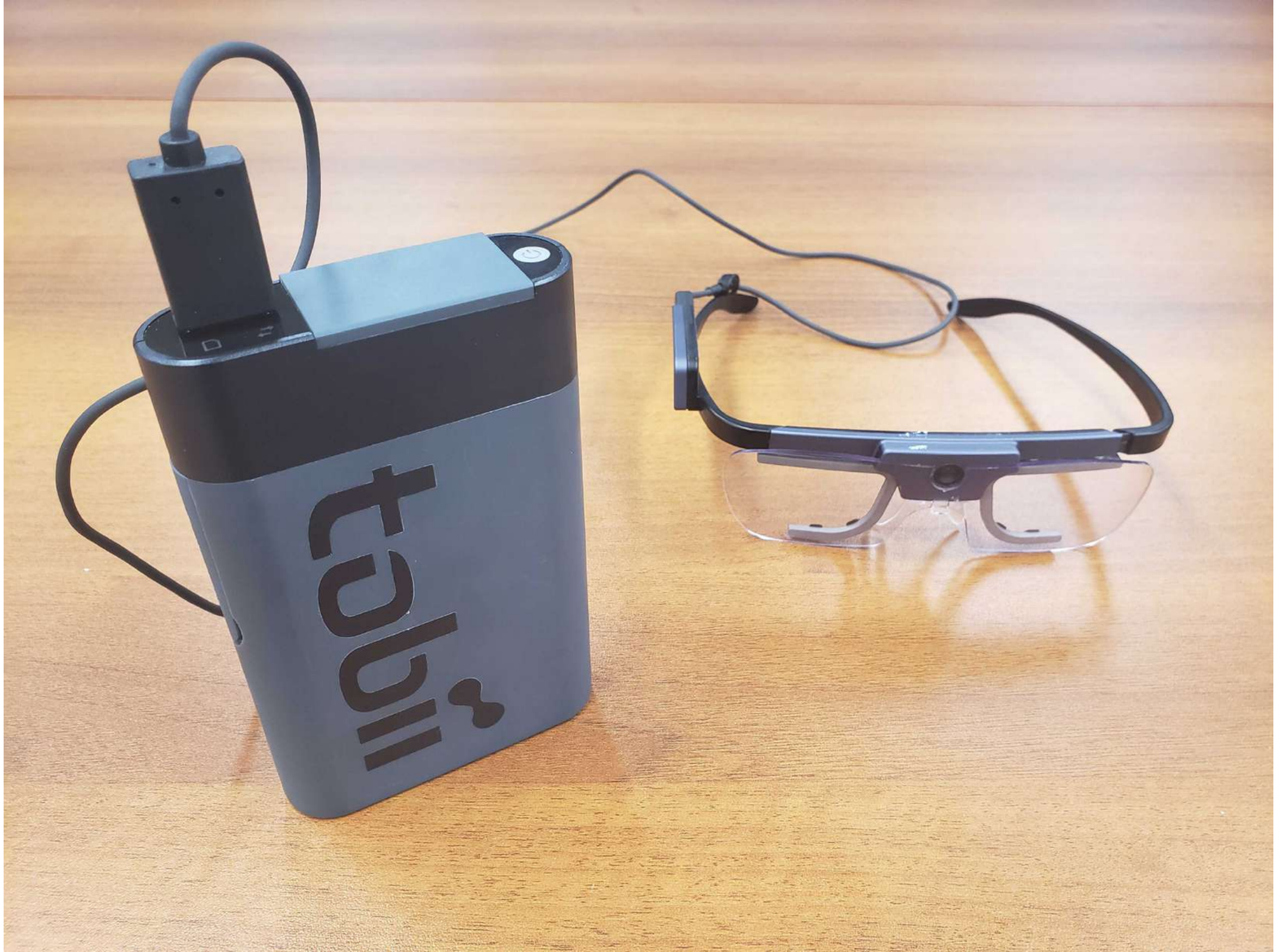}\label{fig:tobii_pic}}
% \subfloat[EEG and Gaze device placement]{\includegraphics[width=0.33\textwidth]{Figures/muse_tobii_sensor_placement2.pdf}\label{fig:eeg_gaze_placement}}
\caption{Wearable EEG, ECG, EDA, and Gaze devices.}
\label{fig:all_sensors}
\end{figure*}

\begin{figure*}
\centering
\subfloat[\small EEG]{\includegraphics[width=0.2\textwidth]{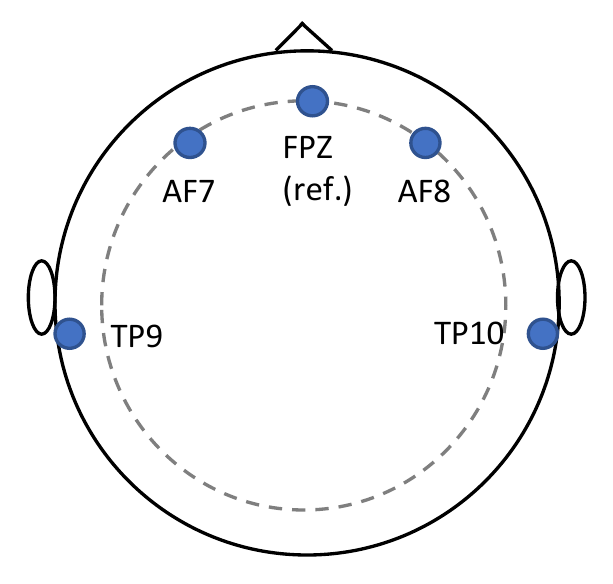}\label{fig:muse_loc}}\hskip1ex
\subfloat[
\small ECG
]{\includegraphics[width=0.2\textwidth]{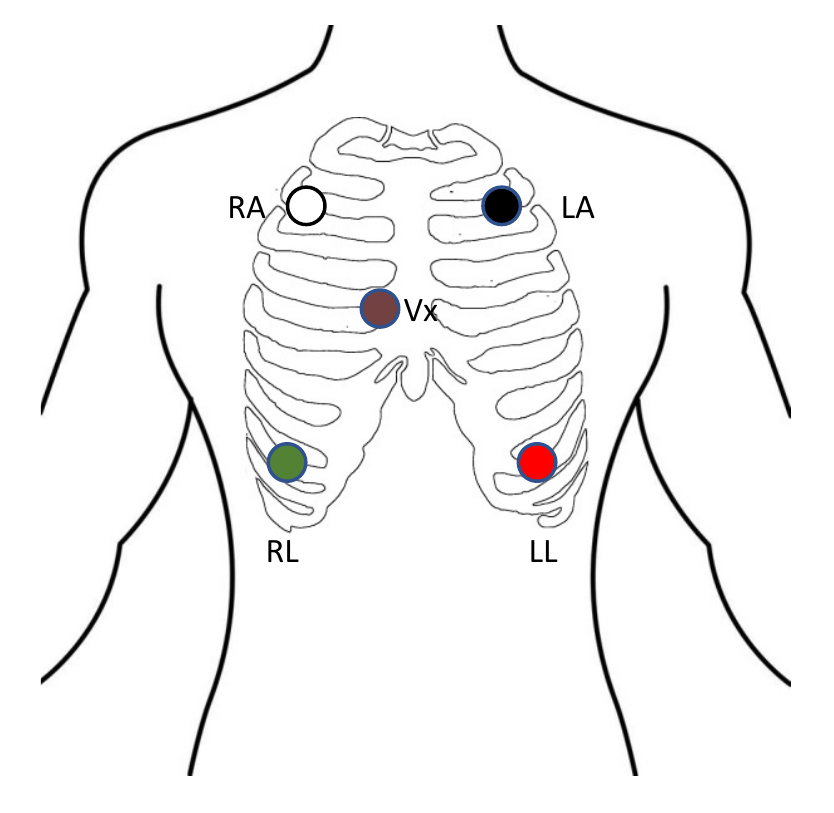}\label{fig:ecg_loc}}\hskip1ex
\subfloat[\small EDA]{\includegraphics[width=0.2\textwidth]{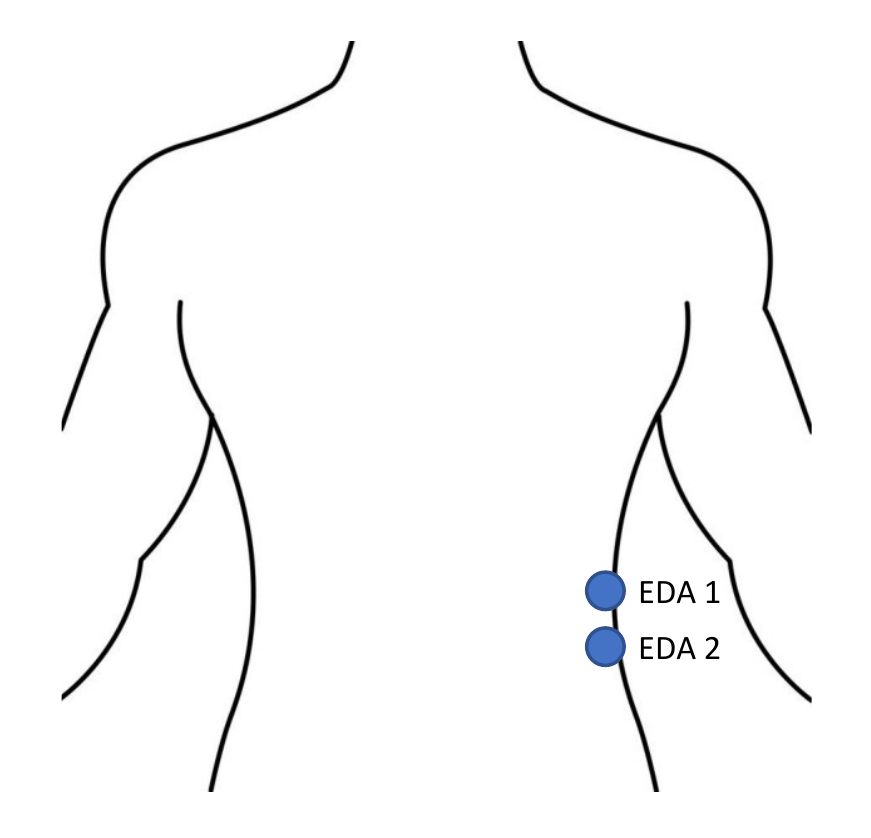}\label{fig:eda_loc}}\hskip1ex
\subfloat[\small EEG and Gaze] {\includegraphics[width=0.3\textwidth]{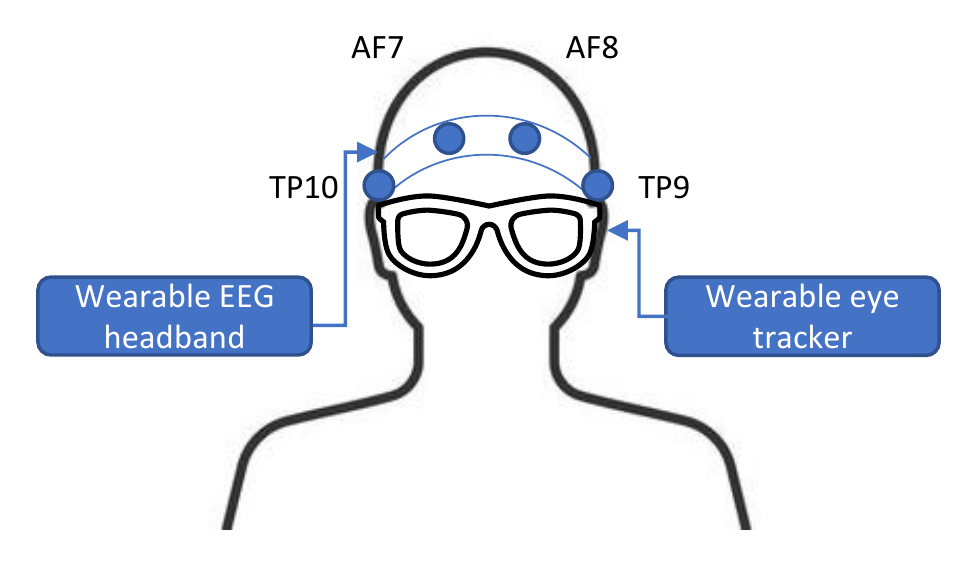}\label{fig:eeg_gaze_placement}}
\caption{EEG, ECG, EDA, and Gaze electrode placements.}
\label{fig:all_sensor_loc}
\end{figure*}

During the experiments which will be described in Section \ref{section:experiment_protocol}, we use four different sensors to collect physiological signals from which to measure cognitive load. Following is a description of each sensor type, namely EEG, ECG, EDA, and Gaze, in detail. Figure \ref{fig:all_sensors} shows the sensors used in our study, while Figure \ref{fig:all_sensor_loc} shows their detailed sensors placement.  

% To collect physiological signals from the participants a separate sensor is used for each of the five modalities: 

% \begin{figure*}
%     \begin{center}
%     \includegraphics[width=1.9\columnwidth]{Figures/sensor_placements_eeg_ecg_eda.pdf} 
%     \caption{(a) EEG electrode placement on the human scalp according to the international $10-20$ system (reproduced directly from \cite{eeglab}), (b) ECG electrode placement, (c) EDA electrode placement}
%     \label{fig:muse_loc}
%     \end{center}
% \end{figure*}

% \begin{figure*}
% \centering
% \subfloat[EEG (international 10-20 system)]{\includegraphics[width=0.3\textwidth]{Figures/eeg_electrode_location.pdf}\label{fig:muse_loc}}\hskip1ex
% \subfloat[ECG (4 leads) ]{\includegraphics[width=0.3\textwidth]{Figures/ecg_placement2.pdf}\label{fig:ecg_loc}}\hskip1ex
% \subfloat[EDA]{\includegraphics[width=0.3\textwidth]{Figures/eda_placement.pdf}\label{fig:eda_loc}}
% \caption{EEG, ECG and EDA electrode location.}
% \end{figure*}

\noindent \textbf{EEG.} 
For collecting EEG signals the Muse S\footnote{https://choosemuse.com/muse-s/} headband shown in Figure~\ref{fig:eeg_device} is used. The device has 4 channels where 2 of them are frontal electrodes located at the forehead in locations AF7 and AF8 (according to the international 10-20 system \cite{10_20_system1,10_20_system2}) while the remaining 2 are temporal electrodes located behind the ears in locations TP9 and TP10. 
% The electrodes are located according to the universal 10-20 system \cite{10_20_system1,10_20_system2} as shown in
Figure \ref{fig:eeg_device} depicts the Muse EEG device while in Figure \ref{fig:muse_loc}, we present the sensor locations of this EEG headset. As shown in the figure, the reference electrode is located at the middle of the forehead in location FpZ. The sampling rate of the EEG headband is $256$ \textit{Hz}. Conductive gel is used to enhance the conductivity between the electrode and the skin. 
\textcolor{black}{We opt for the Muse S headband to ensure both comfort and compatibility with the gaze device. 
% The Muse S headband can be comfortably worn alongside the gaze device.
}

% \begin{figure}
%     \begin{center}
%     \includegraphics[width=0.7\columnwidth]{Figures/eeg_electrode_location.pdf} 
%     \caption{EEG electrode placement on the human scalp according to the international $10-20$ system (reproduced directly from \cite{eeglab}).}
%     \label{fig:muse_loc}
%     \end{center}
% \end{figure}

% \begin{figure}
%     \begin{center}
%     \includegraphics[width=0.7\columnwidth]{Figures/muse_pic2.pdf} 
% m    \caption{Muse S headband}
%     \label{fig:tobii_pic}
%     \end{center}
% \end{figure}

\noindent \textbf{ECG.}  ECG signals are collected through the  Shimmer\footnote{https://shimmersensing.com/product/shimmer3-ecg-unit-2/} sensors \cite{shimmer}, which is shown in Figure \ref{fig:shimmer_ecg_pic}. As depicted in Figure \ref{fig:ecg_loc}, this wearable device uses 5 standard pre-gelled adhesive electrodes from the chest and abdominal area. Among the 4 electrodes, the Right Arm (RA) and Left Arm (LA) are placed on the left and right sides of the manubrium, while Right Leg (RL) and Left Leg (LL) are placed right above the lower costal margin. The reference electrode denoted by Vx is placed slightly on the right of the sternum. The signals collected are LL-RA, LA-RA, and Vx-RA at a sampling frequency of 512 \textit{Hz}. The Shimmer is worn by the participants using a belt and a cradle. 

% \begin{figure}
%     \begin{center}
%     \includegraphics[width=0.6\columnwidth]{Figures/ecg_placement.pdf} 
%     \caption{ECG electrode placement}
%     \label{fig:ecg_loc}
%     \end{center}
% \end{figure}

\noindent \textbf{EDA.} Similar to ECG, the EDA signal is collected using a Shimmer wearable device \cite{shimmer} as shown in Figure \ref{fig:shimmer_eda_pic}. The data is collected using 2 electrodes placed on the left side of the abdomen which is shown in Figure \ref{fig:eda_loc}. The sampling frequency of the EDA Shimmer device is 128 \textit{Hz}

% \begin{figure}
%     \begin{center}
%     \includegraphics[width=0.5\columnwidth]{Figures/eda_placement.pdf} 
%     \caption{EDA electrode placement}
%     \label{fig:eda_loc}
%     \end{center}
% \end{figure}

\noindent \textbf{Gaze} Figure \ref{fig:tobii_pic} shows a Tobii device\footnote{https://www.tobiipro.com/product-listing/tobii-pro-glasses-2/} used to collect eye tracking data. The device is comprised of a head unit and a recording unit. Inside the device there are 2 cameras per eye as well as a wide angle scene camera. 
From the eye tracking device we record the 
% raw gaze data which is further processed to extract 
specific eye movement events such as saccade, fixation, and others. 
The sampling frequency is 50 \textit{Hz}.  Figure \ref{fig:eeg_gaze_placement} illustrates the placement of the eye tracking device along with the EEG headset.   

% \begin{figure*}
% \centering
% \subfloat[Wearable EEG headband]{\includegraphics[width=0.257\textwidth]{Figures/muse_pic3.pdf}\label{fig:eeg_device}}\hskip1ex
% \subfloat[Wearable eye tracker]{\includegraphics[width=0.257\textwidth]{Figures/tobii_pic3.pdf}\label{fig:tobii_pic}}\hskip1ex
% \subfloat[EEG and Gaze device placement]{\includegraphics[width=0.33\textwidth]{Figures/muse_tobii_sensor_placement2.pdf}\label{fig:eeg_gaze_placement}}
% \caption{Worn EEG and eye tracking device}
% \end{figure*}

\subsection{Experiment Test-bed} 
\begin{figure}
    \begin{center}
    \includegraphics[width=1.0\columnwidth]{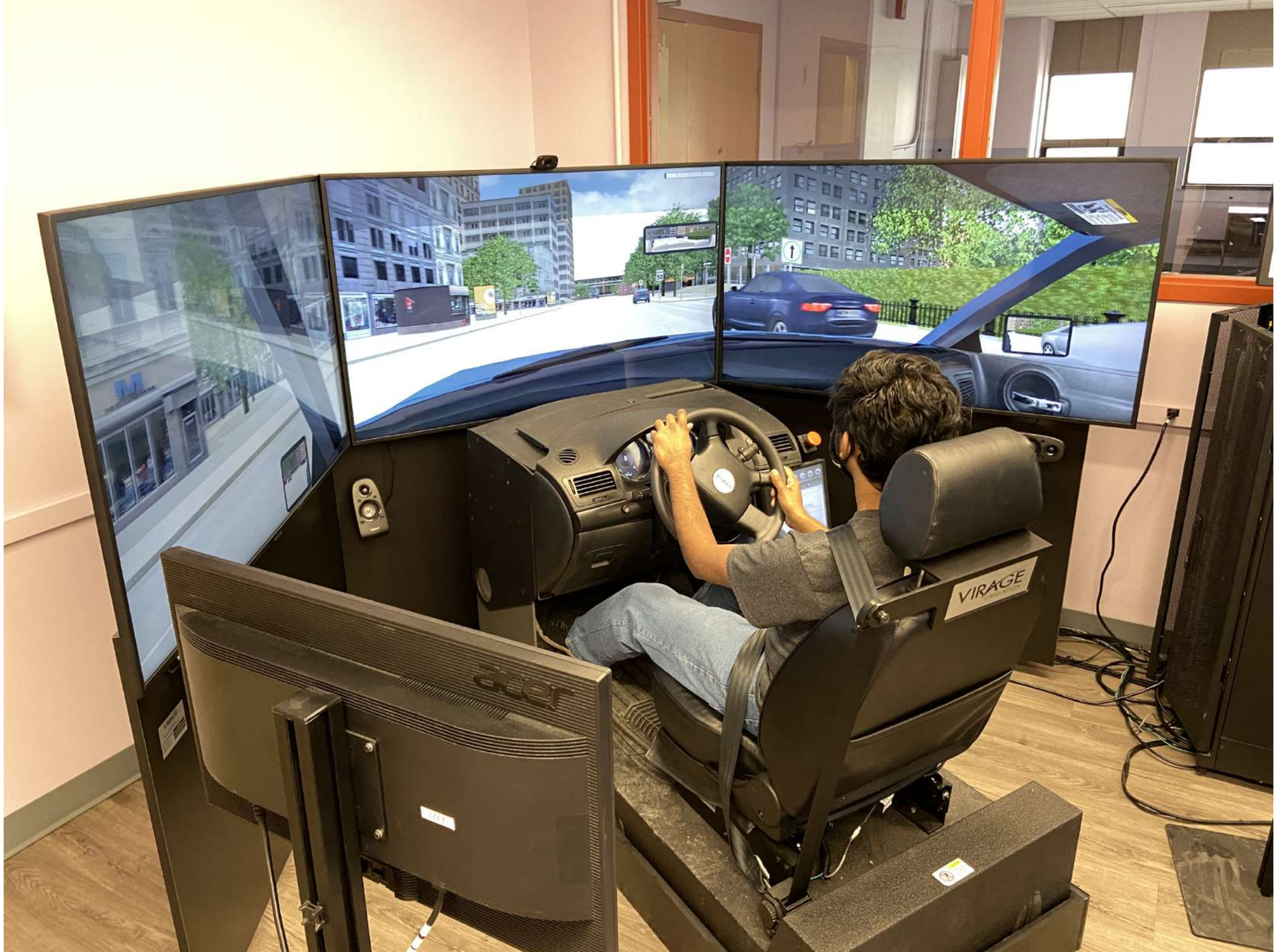} 
    \caption{The immersive vehicle simulator used in this study.}
    \label{fig:virage}
    \end{center}
\end{figure}

% We designed our experiment using the Virage driving simulator\footnote{https://viragesimulation.com/vs500m-car-simulator-training-and-research/} shown in Figure~\ref{fig:virage}. 
In order to simulate driving and be able to control the parameters surrounding the driving experience, we use a driving simulator\footnote{https://viragesimulation.com/vs500m-car-simulator-training-and-research/} shown in Figure~\ref{fig:virage}. The driving simulator includes elements similar to a real car, including steering wheel, dashboard, accelerator, and brake. 
These components combined with a motion system
% which provides haptic feedback to the seat and wheel, 
provides participants with a more realistic driving sensation. 
The motion system can emulate real-life motions up to 100 \textit{Hz} in frequency. This includes vibrations from road texture, acceleration, braking, speeding, and turning along with other essential movements to provide users with engaging haptic feedback. 
Additionally, there are three 55 inch LCD screens which provide a 180 degree view from the front, plus two additional screens for the blind spots, together creating an immersive experience. Each front screen has a display resolution of 1920 $\times$ 1080 pixels. Moreover, directional sound is incorporated using a surround sound system. The sound is intended to mimic typical sounds heard while driving including the sound of the engine, speeding and passing vehicles, and horns, among others. 

A debrief station is designed to provide a complete video of the participant and simulation screens during the experiment along with performance graphs. There is a webcam mounted on the top of the middle frontal screen, which has a resolution of 720p. The camera records video of the participants while driving in the simulator. 
% including facial features. 
The performance information displayed in the debrief station includes driving performance data, including the time required for breaking and acceleration, possible crashes, and others. This data can also be used for performance analysis.
% There is option to include pedestrians, dogs, deers and other vehicles on the road as well. 

% \hl{Data available?
% what it Records?
% what kind of vehicles?
% objects?
% maps?}
% \hl{image of the debrief station}

%%%%% STOPPED HERE

\subsection{Driving Scenarios}
\begin{figure}
    % \begin{center}
    \includegraphics[width=1.1\linewidth]{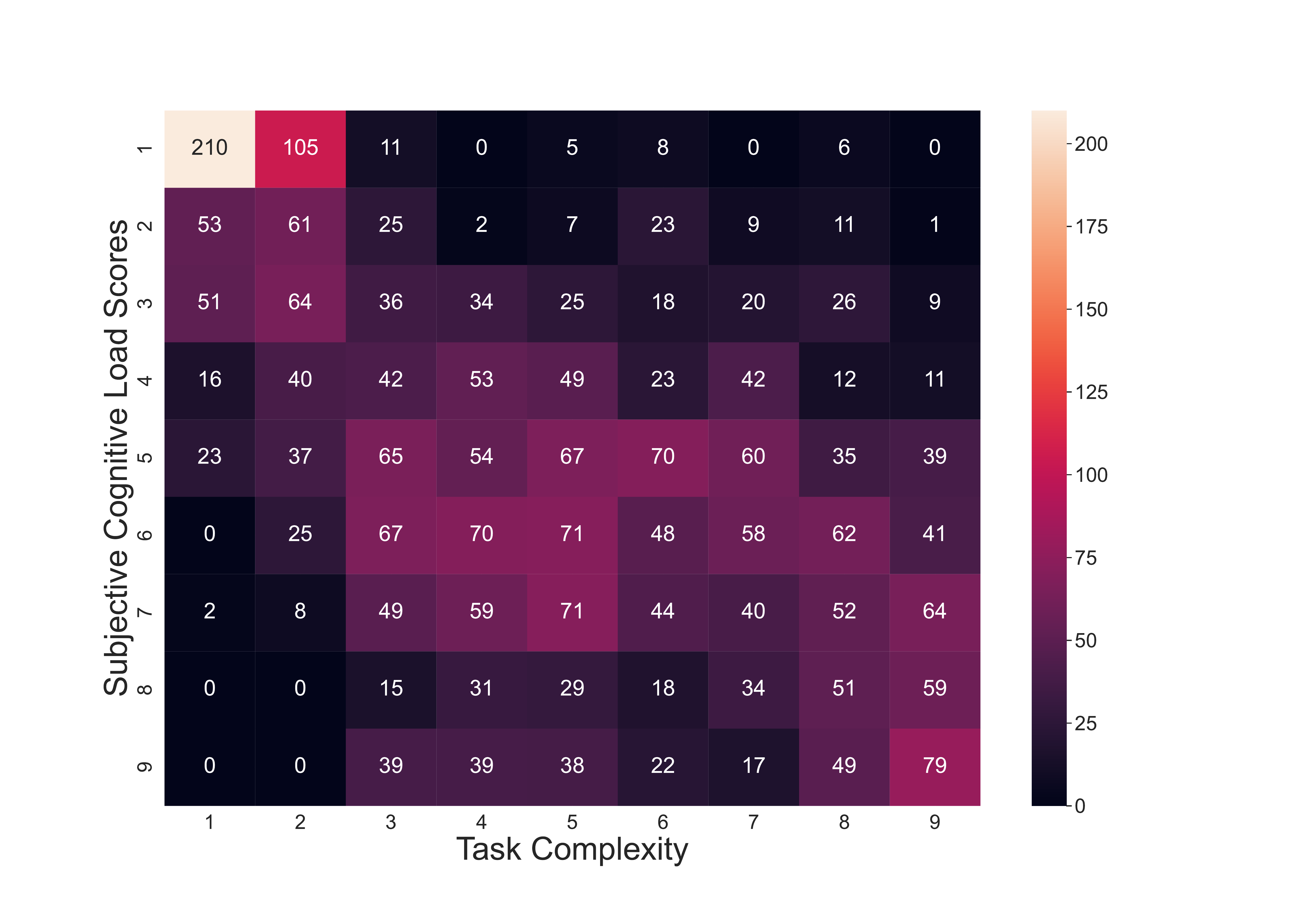}
    \caption{\textcolor{black}{Complexity vs. subjective cognitive load scores. The lighter shade means more sample.}}
    \label{fig:comp_vs_subj}
    % \end{center}
\end{figure}

The simulator comes with a number of pre-built driving scenarios in which the vehicle type, environmental conditions, and other factors can vary. Each scenario consists of a number of tasks that need to be performed, e.g., keeping the speed above a certain threshold. Moreover, each scenario has a designated complexity level. We choose 9 different scenarios, one from each complexity level. 
\textcolor{black}{The scenarios encompass a range of common challenges encountered during everyday driving, such as driving on highways, at night-time, and in snowy conditions (scenarios 1, 2, 3 respectively), maintaining/changing speed levels (scenarios 4, 5, 6, 7 and 9), avoiding accidents (scenarios 4, 5, 6, 7 and 9), back-to-back turns (scenario 7), and turning the car around on a narrow road using 3-point turn (scenario 8). The scenarios induce different levels of cognitive load in the participants based on a number of pilot tests that we carried out to choose these 9 from among a larger pool of possible tasks. 
% Moreover, the scenarios are designed to increase in difficulty level, but still be able to be completed by an average driver.
Moreover, the scenarios are structured to progressively increase in difficulty, yet remain achievable for the average driver. 
In Figure \ref{fig:comp_vs_subj}, we depict the heat map of the frequency of the ratings for each driving scenario. We observe that as the scenarios progress, more and more participants select higher cognitive load scores, indicating that the task complexities do indeed increase as the scenarios progress, especially for scenarios 8 and 9.}

% \textcolor{red}{The scenarios encompass a range of common challenges encountered during everyday driving, such as consecutive turns (scenario 7), executing a 3-point turn in a narrow road (scenario 8), and maintaining varying speeds displayed on the screen (scenarios 4, 5, 6, 7, and 9). Additionally, the participants must drive accurately and avoid collisions, which led us to include challenges that require precision (scenarios 4, 5, 6, 7, and 9). These diverse challenges result in different levels of cognitive load experienced by the participants. The scenarios are intentionally designed to escalate in difficulty, providing increasing challenges as they progress. However, individuals may vary in their response to these challenges. For instance, some participants might encounter high cognitive load while attempting the tennis ball challenge (scenario 4), where they must hit the ball with the tire. On the other hand, others might find higher cognitive load when navigating through the gates challenges. Overall, the scenarios are thoughtfully selected to induce varying levels of cognitive load among all study participants, accounting for individual differences in how they perceive and respond to the challenges presented.}
%  to cover all 9 on the complexity levels
% All the driving tasks were based on the driving simulator's ambulance driving package and modified as necessary. 
% The tasks started with an orientation followed by 9 linear complexity levels in ascending order. 
% The task complexity level was increased from level 1 to level 9 where 
The duration of each scenario is set to 3 minutes.
% with 2 minutes baseline in between. 
An orientation scenario was designed and performed by each participant at the beginning of the session to allow each participant to adapt to the simulator. \textcolor{black}{In the orientation session, we described the system first as the system completes a number of steps to ensure things like the turn signals, motion sensors, brake, accelerator, ignition, emergency stop, seat belt, and others function properly. Participants then began to drive on the highway to become familiar with the speed display, on-screen arrow that provides cues about the direction of driving, and all the other necessary indicators, while we were present to provide help if needed and answer any questions.}
% \textcolor{red}{At the start of the experiment, participants were required to become acquainted with the system and the on-screen instructions to aid their navigation. As a result, each participant underwent an orientation session, during which they drove on the highway and familiarized themselves with the speed display, on-screen arrow indicating turns, and other essential indicators.} 
Table~\ref{table_virage_tasks} shows the description of each driving scenario along with its corresponding complexity level.

{\renewcommand{\arraystretch}{1.0}
\begin{table*}[!t]
    \footnotesize
    \centering
    \caption{Driving scenario details}
    % \begin{tabularx}{\textwidth}{p{0.15\textwidth}llXX}
    \begin{tabular}{lll}
    % \begin{tabular}{l l p{0.40\linewidth}}
    \hline
% 		\rowcolor{lightgray} 
        \textbf{Scenario/} & \multirow{ 2}{*}{\textbf{Simulation}} & \multirow{ 2}{*}{\textbf{Description}} \\
        \textbf{Complexity} & & \\
		% \textbf{Scenario/Complexity} & \textbf{Simulation Tasks} & \textbf{Description} \\
		\hline \hline
		0/Orientation & Highway driving & Maintain centre \\
	  % \hline
		1 & Highway driving & 80km/h\\
% 		\hline
		2 & Night time driving & 80km/h\\
% 		\hline
		3 & Night time driving in the highway w/ snow & 80km/h\\
% 		\hline
		4 & Tennis ball challenge & Hit the tennis ball with the tire, maintain accuracy, try to accelerate\\
% 		\hline
		5 & Slalom challenge & Navigate through gates, maintain accuracy, try to accelerate\\
% 		\hline
		6 & Narrow passage challenge & Navigate through gates, maintain accuracy, try to accelerate\\
% 		\hline
		7 & 90-degree turn challenge & Take 90 degrees left and right turns, maintain accuracy, try to accelerate\\
% 		\hline
		8 & 3-point turn challenge & Drive while following instructions, take 3 point turn\\
% 		\hline
		9 & Narrow alley challenge & Navigate through narrow alley, maintain accuracy, try to accelerate\\
		\hline
    % \end{tabularx}
    \end{tabular}
    \label{table_virage_tasks}
\end{table*}
}

\subsection{Simulation Adaptation Syndrome}
It has been shown in prior research that Simulation Adaptation Syndrome (SAS) can affect different participants \cite{rizzo2003demographic, reed2009comparing}. SAS can range from feeling minor discomfort to severe symptoms such as dry mouth, dizziness, vertigo, vomiting, nausea, and disorientation while taking part in simulations such as driving a vehicle ~\cite{sas_dobie}. The main cause of SAS is the discrepancy between the 
% virtual 
sensory inputs such as visual and vestibular system (which is responsible for our sense of balance ~\cite{sas_cobb, sas_galvez}). One of the challenges we faced during our study was avoiding and minimizing SAS. As recommended in the simulator instructions, 
% According to the driving simulator manual, 
we followed the following 5 steps to manage and reduce SAS as much as possible:
\begin{enumerate}
  \item \textbf{Cool room:} the simulator room must be cool and well ventilated;
  \item \textbf{Confident introduction:} must create a calm and relaxed environment;
  \item \textbf{Cautious alert:} moving on slowly to allow the driver's time to adjust;
  \item \textbf{Careful observation:} must actively look for signs and symptoms of SAS;
  \item \textbf{Cease driving:} must pause immediately on observing the slightest sign of SAS.
\end{enumerate}

SAS can be managed by carefully monitoring the participant's level of discomfort using a Likert scale and asking the participants to do an intermittent self assessment. In our case, the participants were asked to self report on their SAS level every minute using a 9-point Likert scale as shown in Table~\ref{table_sas_level}, which was made based on the Motion Sickness Questionnaire (MSQ)~\cite{msq}, Simulator Sickness Questionnaire (SSQ)~\cite{ssq}, and The Motion Sickness Assessment Questionnaire (MSAQ)~\cite{msaq}. Based on these pre-cautions and careful monitoring, SAS resulted in pausing and discontinuing only 2 participants. 
%The data from these two incomplete sessions were not included 

% {\renewcommand{\arraystretch}{1.8}
\begin{table}[!t]
\caption{SAS levels and their corresponding description.}
\centering
\begin{tabular}{l l}
\hline 
% 		\rowcolor{lightgray} 
		\textbf{SAS Level} & \textbf{Description} \\
		\hline \hline
		1 & Feeling no adverse effects\\
	   % \hline
		2 & Feeling very mild discomfort \\
% 		\hline
		3 & Feeling mild discomfort \\
% 		\hline
		4 & Feeling mild to moderate discomfort  \\
% 		\hline
		5 & Feeling moderate discomfort \\
% 		\hline
		6 & Feeling moderate to pronounced discomfort \\
% 		\hline
		7 & Feeling pronounced discomfort \\
% 		\hline
		8 & Feeling pronounced to severe discomfort \\
% 		\hline
		9 & Feeling severe discomfort (potential for vomiting) \\
		\hline
	\end{tabular}
	\label{table_sas_level}
\end{table}
% }

% \setlength{\arrayrulewidth}{0.08mm}
% \setlength{\tabcolsep}{10pt}
% \renewcommand{\arraystretch}{1.5}
% \begin{table}
% 	\caption{PAAS subjective cognitive load scores of 9 levels}
% 	\begin{tabular}{|l|l|}
% 		\hline
% % 		\rowcolor{lightgray} 
% 		\textbf{PAAS subjective cognitive load scores} & \textbf{Description} \\
% 		\hline
% 		1 & Very, very low\\
% 	    \hline
% 		2 & Very low \\
% 		\hline
% 		3 & Low \\
% 		\hline
% 		4 & Rather low  \\
% 		\hline
% 		5 & Neither low nor high \\
% 		\hline
% 		6 & High \\
% 		\hline
% 		7 & Rather high \\
% 		\hline
% 		8 & Very High \\
% 		\hline
% 		9 & Very, very high \\
% 		\hline
% 	\end{tabular}
% 	\label{table_paas_level}
% \end{table}

% {\renewcommand{\arraystretch}{1.8}

\begin{figure}[!t]
    \begin{center}
    \includegraphics[width=1.0\columnwidth]{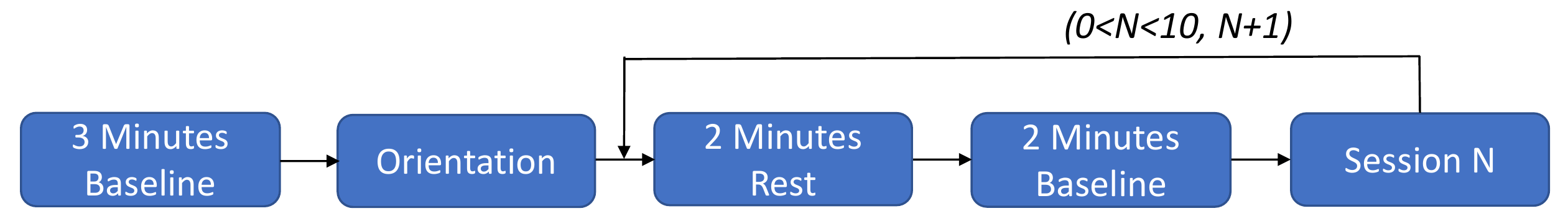} 
    \caption{The experiment flow.}
    \label{fig:exp_flow}
    \end{center}
\end{figure}

\subsection{Participants}
Data was collected from 23 participants including 17 females and 6 males. 
% with an average age of 26.
\textcolor{black}{The participant's gender was not controlled for and we merely included volunteers regardless. We did not actively seek individuals of specific genders, as it was not a prerequisite for our study. Given that the key focus of this study has been cognitive load during driving, we controlled for having a driving license and a few years of driving experience. We also ensured that participants were not under the influence of any substances at the time of data collection. The average age of participants was 26.9. Background health information were not collected in this study.}
% All the participants had prior driving experience. 
% This information decreases variability associated with driving experience and age. 
% We took measures to control for variables such as sleep, physical exertion, general anxiety level, and fitness level. 
% By doing so, we aimed to ensure that our findings were not biased or unduly influenced by these factors. 
% Every participant falls into the category of either a student or a working professional. These factors including others such as health status would affect the cognitive load levels experienced by the participants, however, our goal is to evaluate cognitive load for all the participants in general and not for different groups based on age, health, or other factors.} 

Prior to the simulation, participants were provided with detailed information on the experimental process and the research team received written consent. The study was approved by Queen's University's General Research Ethics Board (GREB). Among the 23 participants, 2 data collection sessions were stopped due to high levels of SAS, while the data from another 3 sessions was incomplete due to device or connectivity issues. Specifically, for participant 13, we only have data for scenarios 3, 4, 5, 6, and 9, while for participant 16, we have data for scenarios 1, 5, 6 and 9. Finally, for participant 18, we have the data for scenarios 1 to 6. While the data from the two sessions that were incomplete due to SAS are not incorporated in the dataset as they were interrupted too early in the process, the data from the three incomplete sessions are incorporated.
% We have included all the participants data in for training the classifiers. 

% The participants were contacted via email 48 hours before the experiment and were informed about the experiment process and all the other relevant information. Informed written consent was received from each participant before the experiment day. Each participant also filled out the Covid-19 screening form on the day of the experiment. Before the experiment day, the participants were asked to participate in an online Attention Network Test (ANT)~\cite{ant_2002, ant_2005}. The study was approved by General Research Ethics Board (GREB) of Queen's University. 

\subsection{Cognitive Load Self-Assessment}
\begin{figure*}
\centering
\subfloat[Subject 1]{\includegraphics[width=0.13\textwidth]{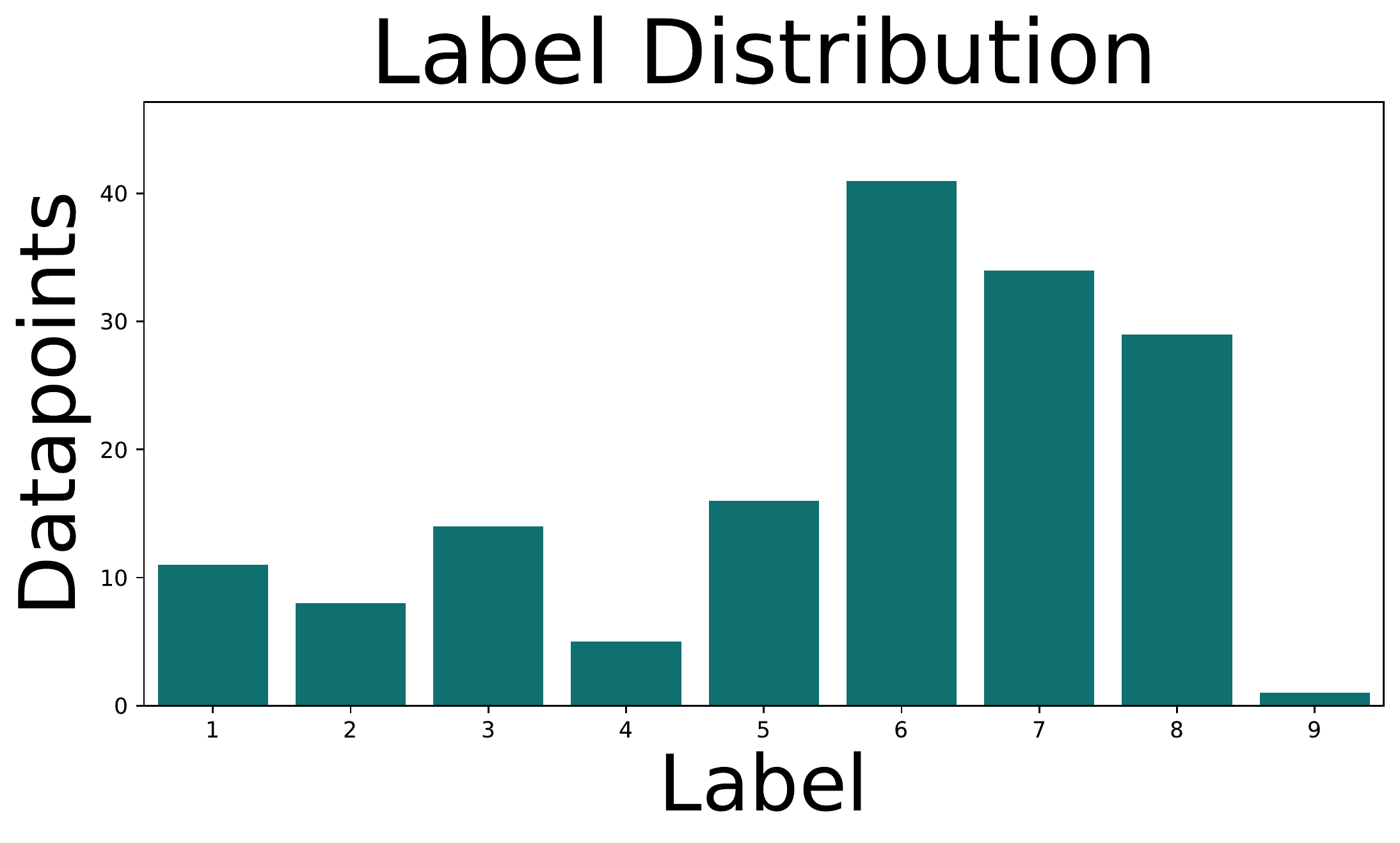}\label{fig:sub1}}\hskip1ex
\subfloat[Subject 2]{\includegraphics[width=0.13\textwidth]{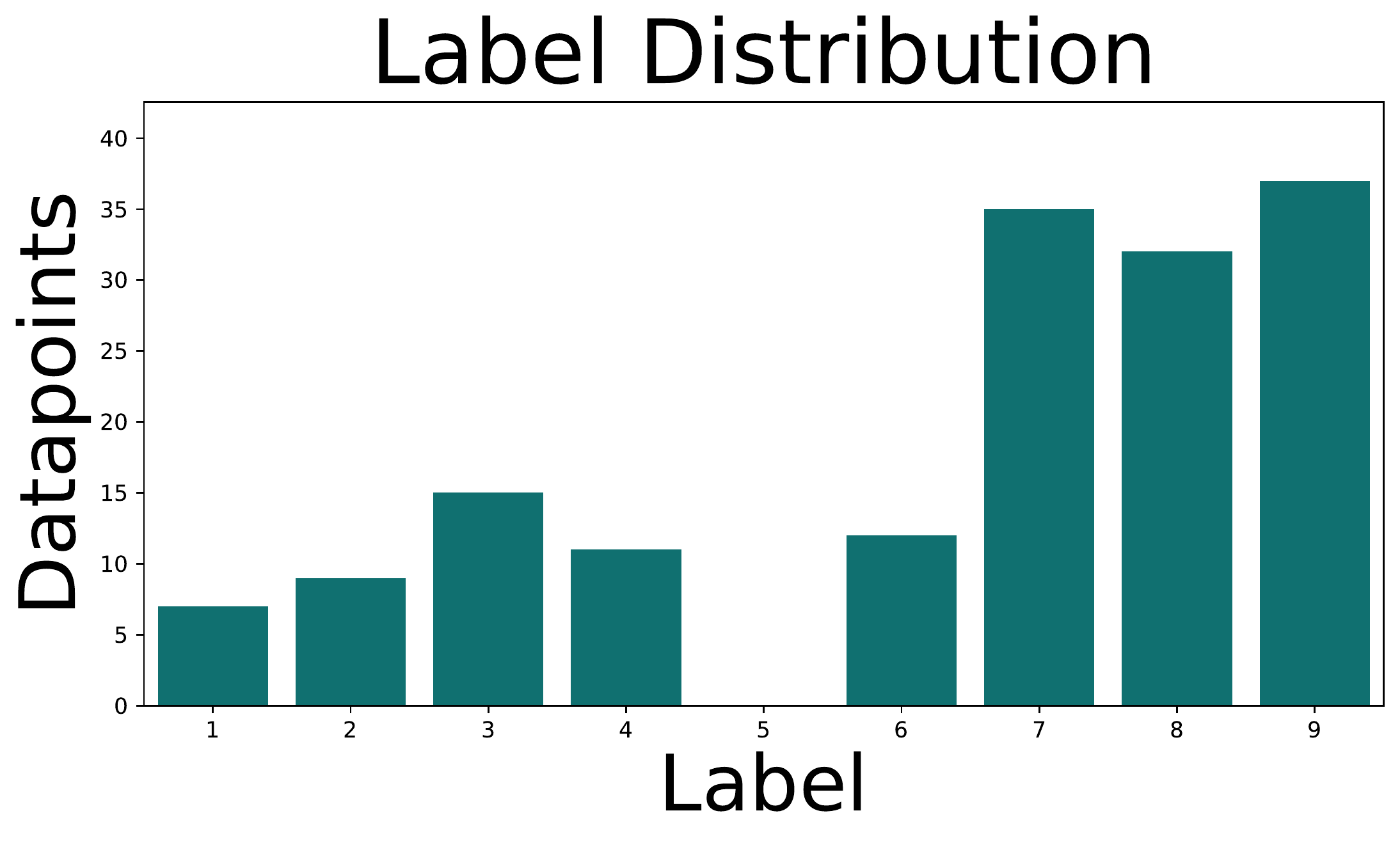}\label{fig:sub2}}\hskip1ex
\subfloat[Subject 3]{\includegraphics[width=0.13\textwidth]{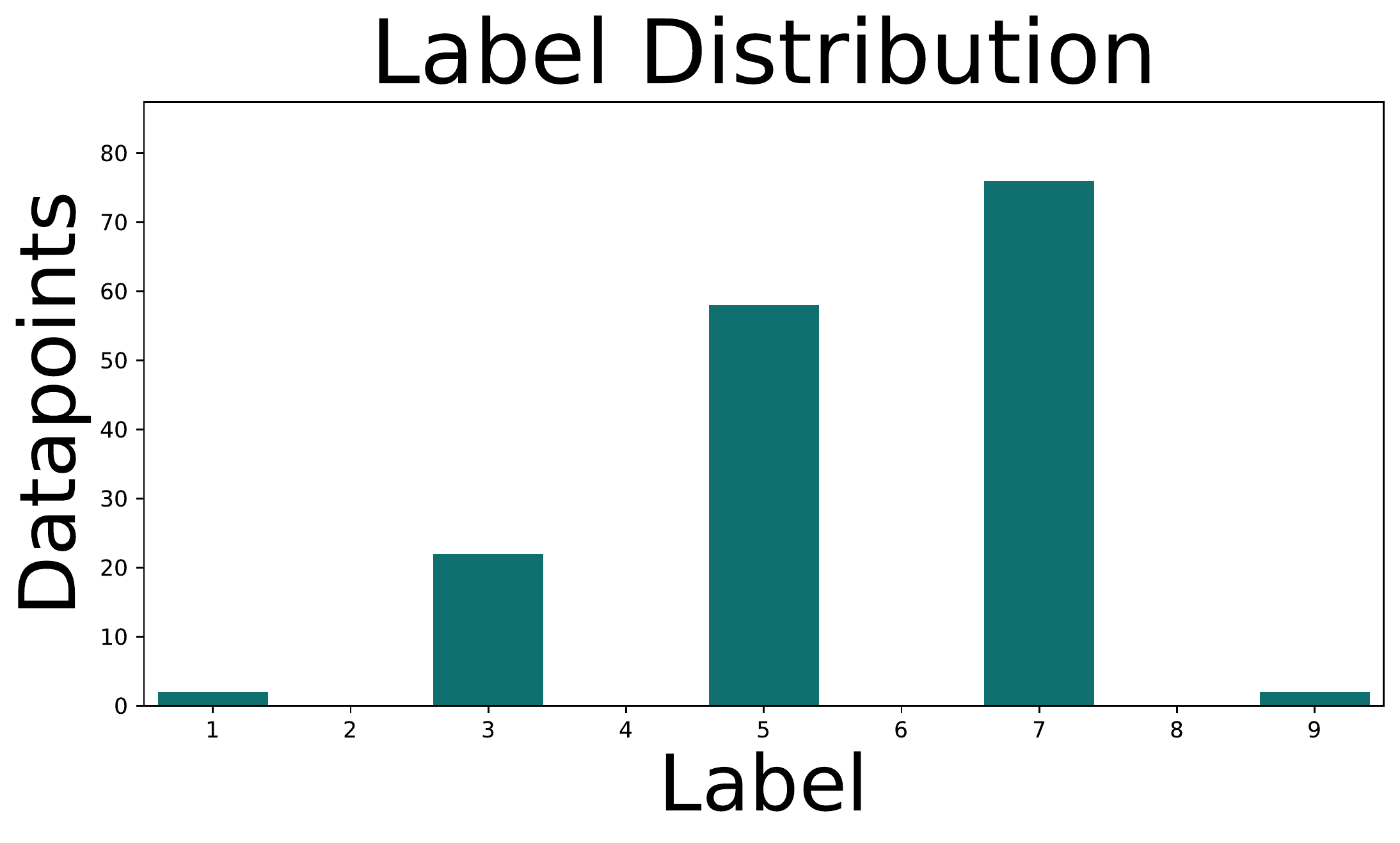}\label{fig:sub3}}\hskip1ex
\subfloat[Subject 4]{\includegraphics[width=0.13\textwidth]{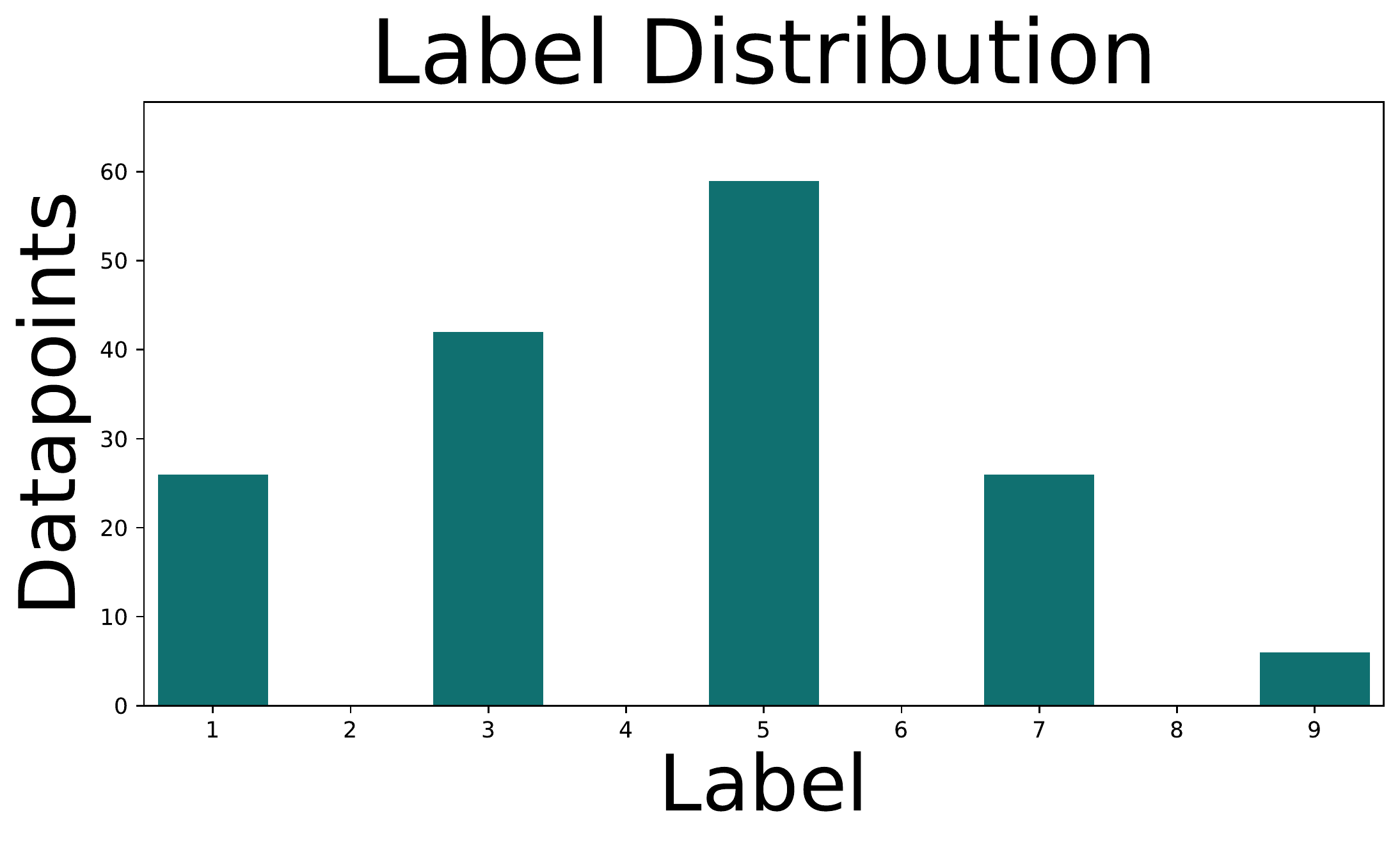}\label{fig:sub1}}\hskip1ex
\subfloat[Subject 5]{\includegraphics[width=0.13\textwidth]{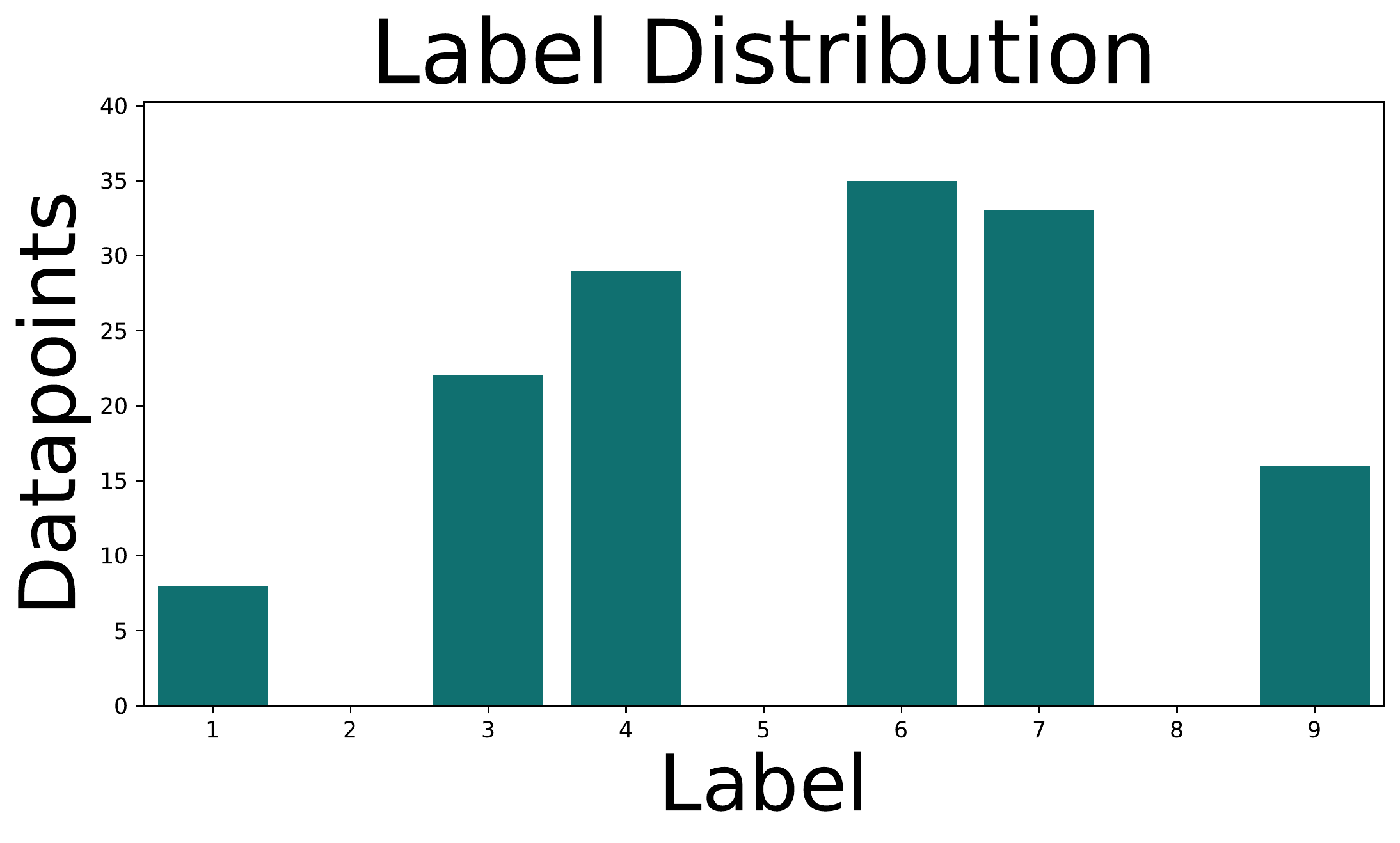}\label{fig:sub2}}\hskip1ex
\subfloat[Subject 6]{\includegraphics[width=0.13\textwidth]{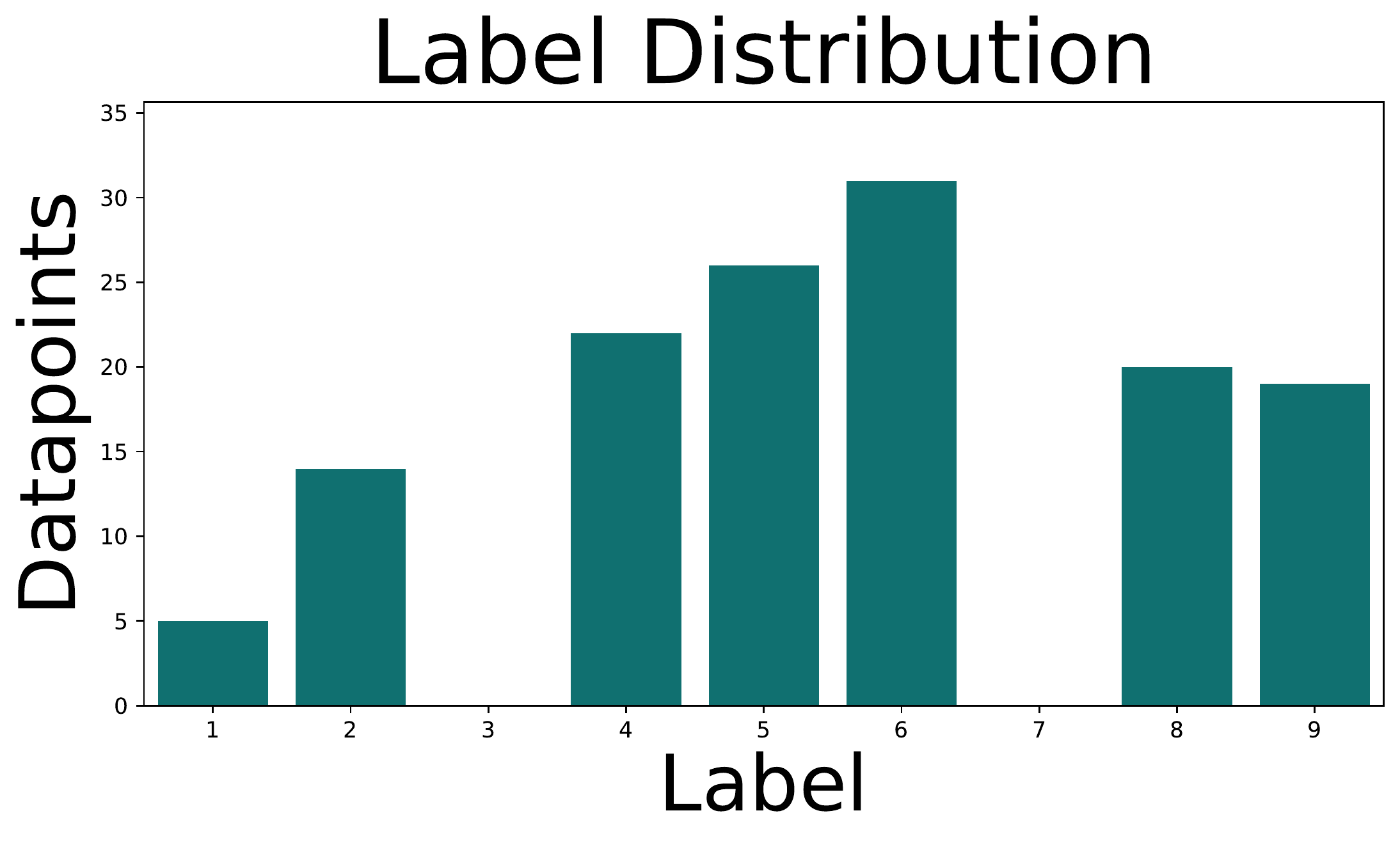}\label{fig:sub3}}\hskip1ex
\subfloat[Subject 7]{\includegraphics[width=0.13\textwidth]{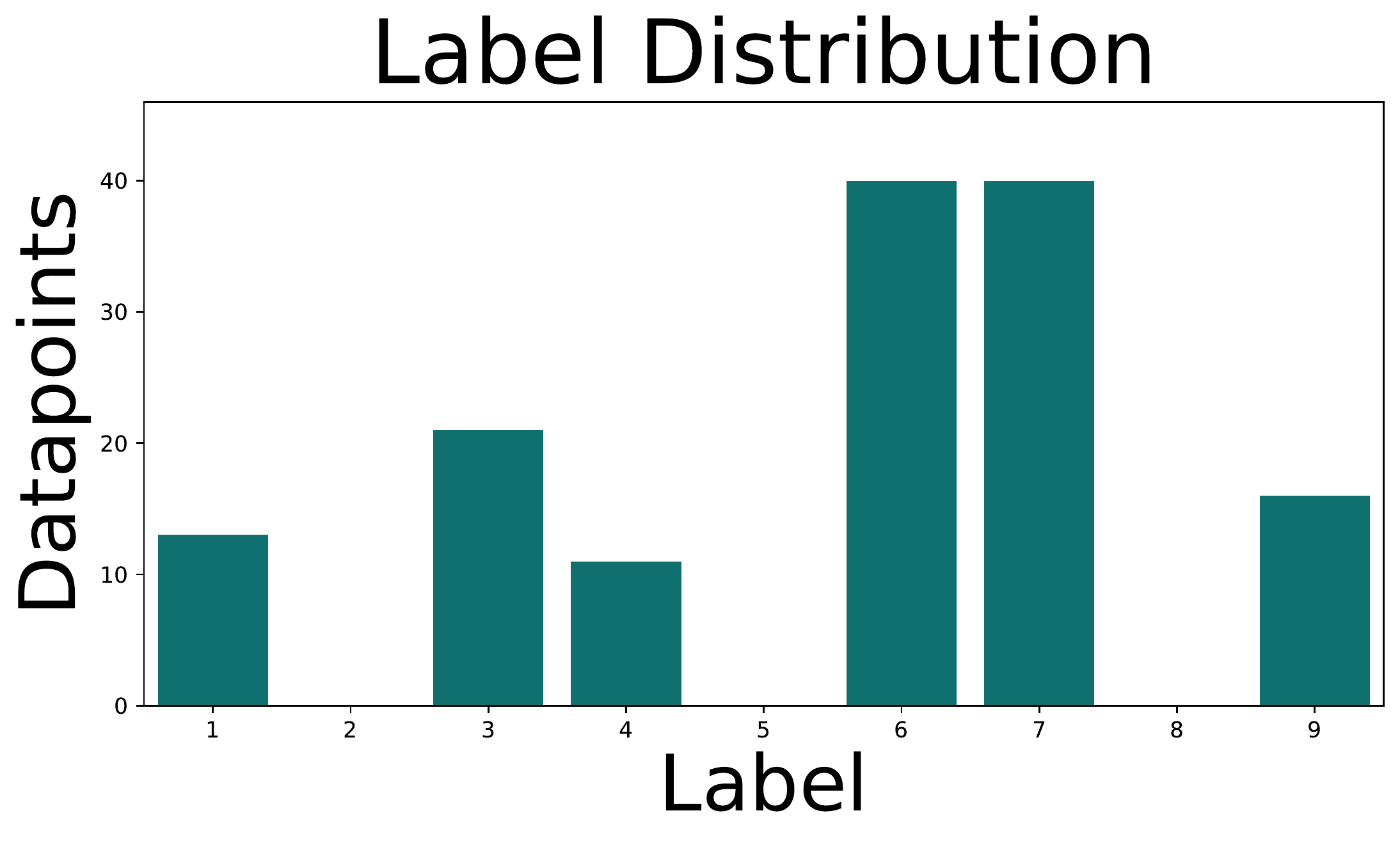}\label{fig:sub1}}\hskip1ex
\subfloat[Subject 8]{\includegraphics[width=0.13\textwidth]{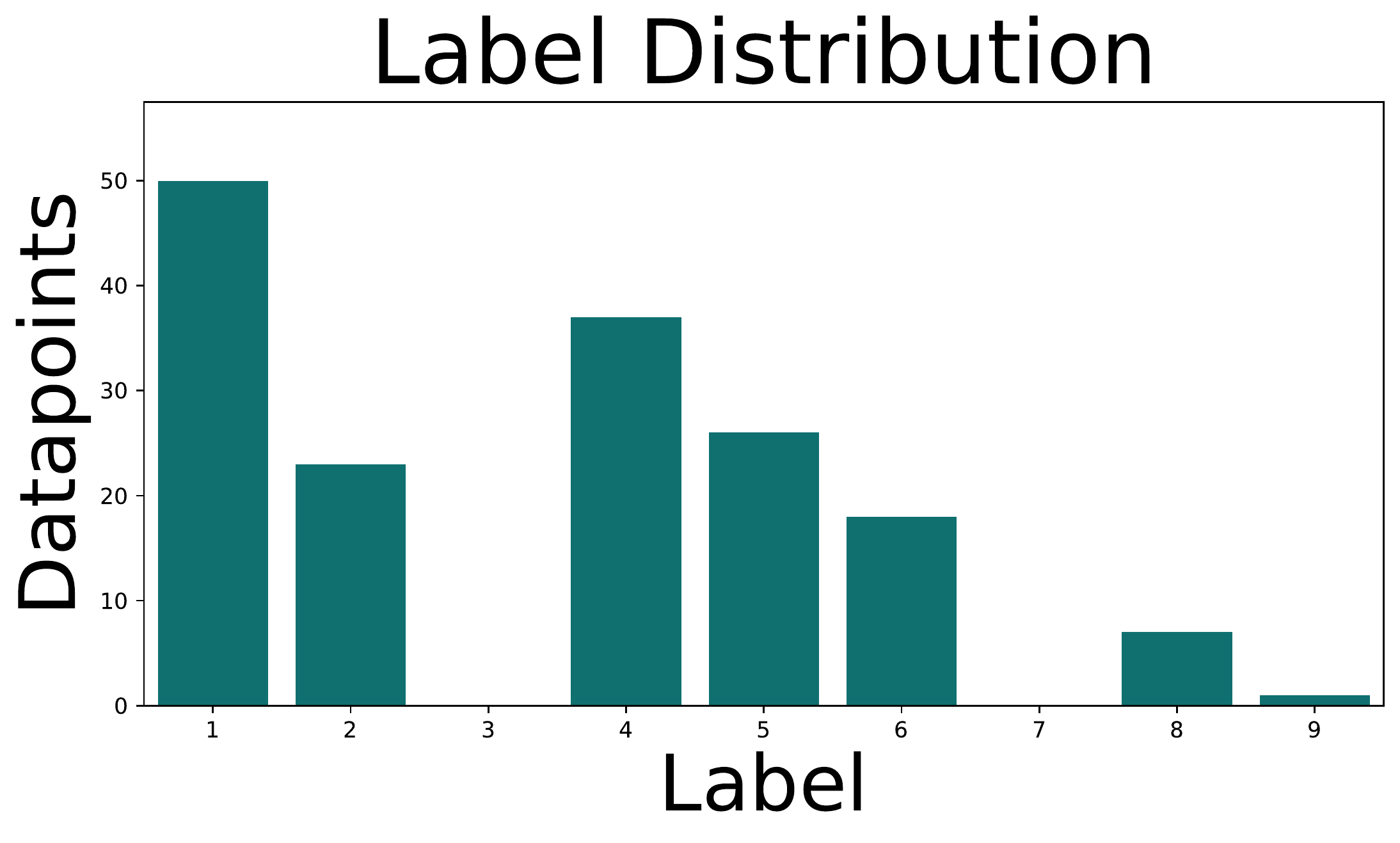}\label{fig:sub2}}\hskip1ex
\subfloat[Subject 9]{\includegraphics[width=0.13\textwidth]{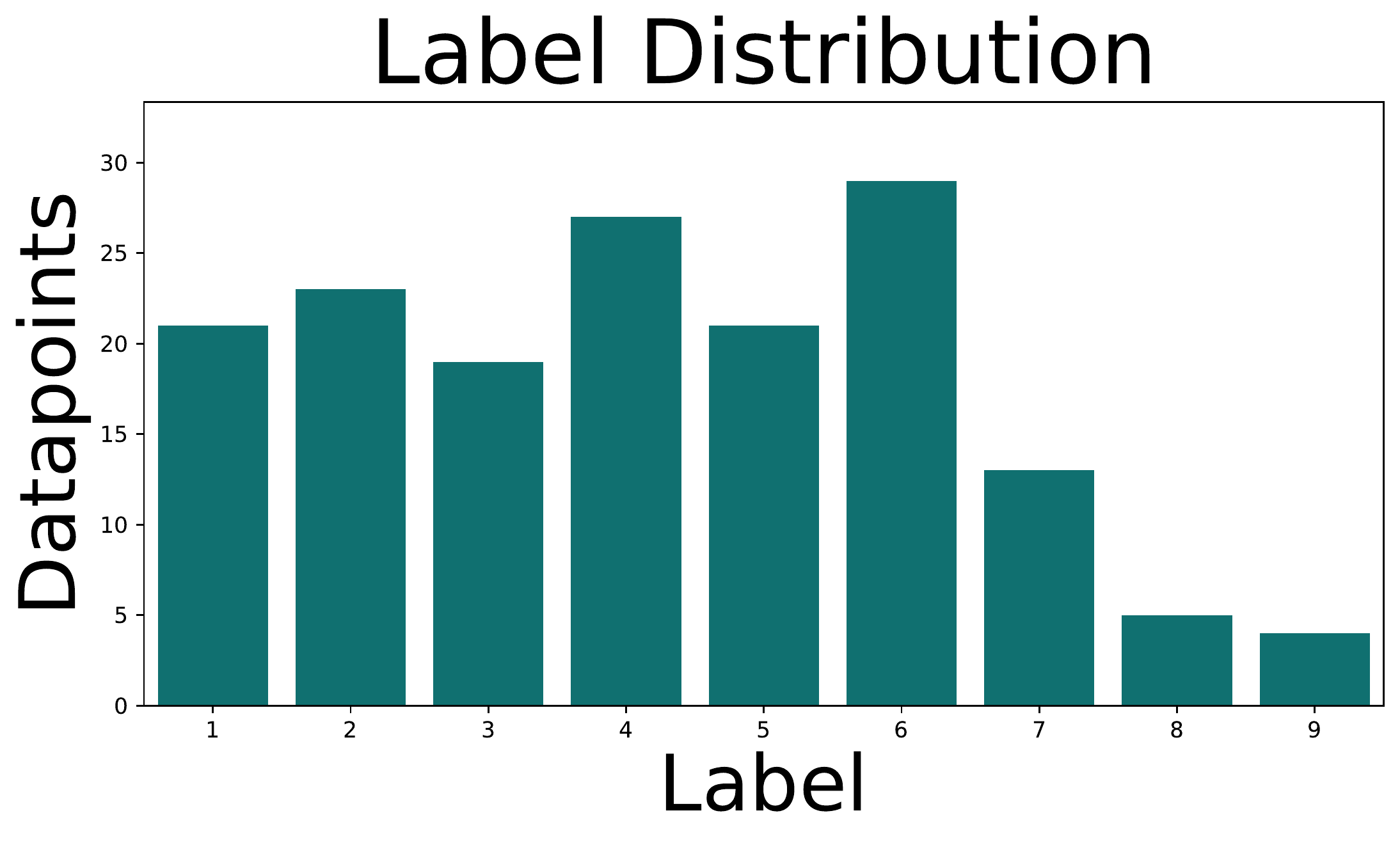}\label{fig:sub3}}\hskip1ex
\subfloat[Subject 10]{\includegraphics[width=0.13\textwidth]{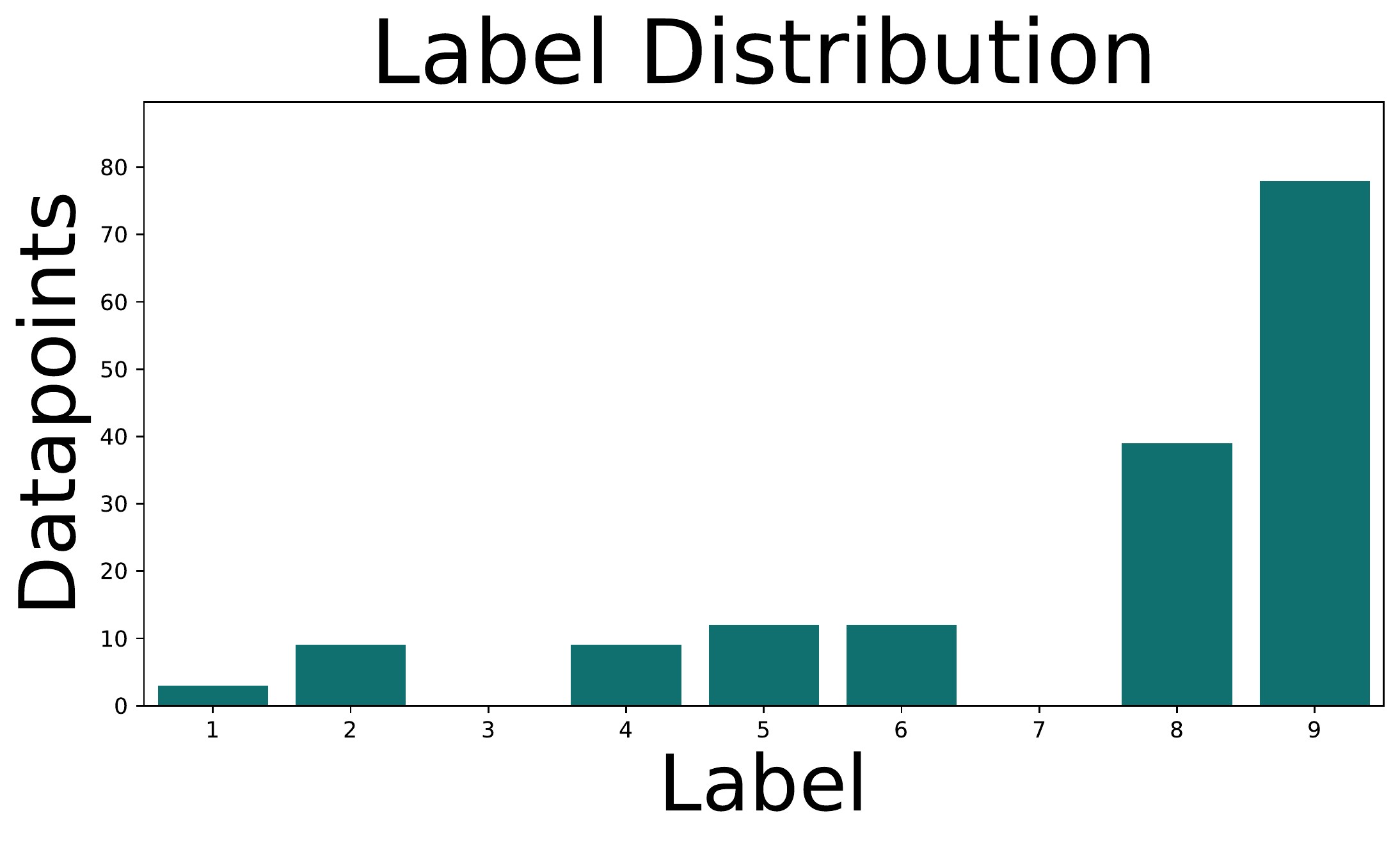}\label{fig:sub1}}\hskip1ex
\subfloat[Subject 11]{\includegraphics[width=0.13\textwidth]{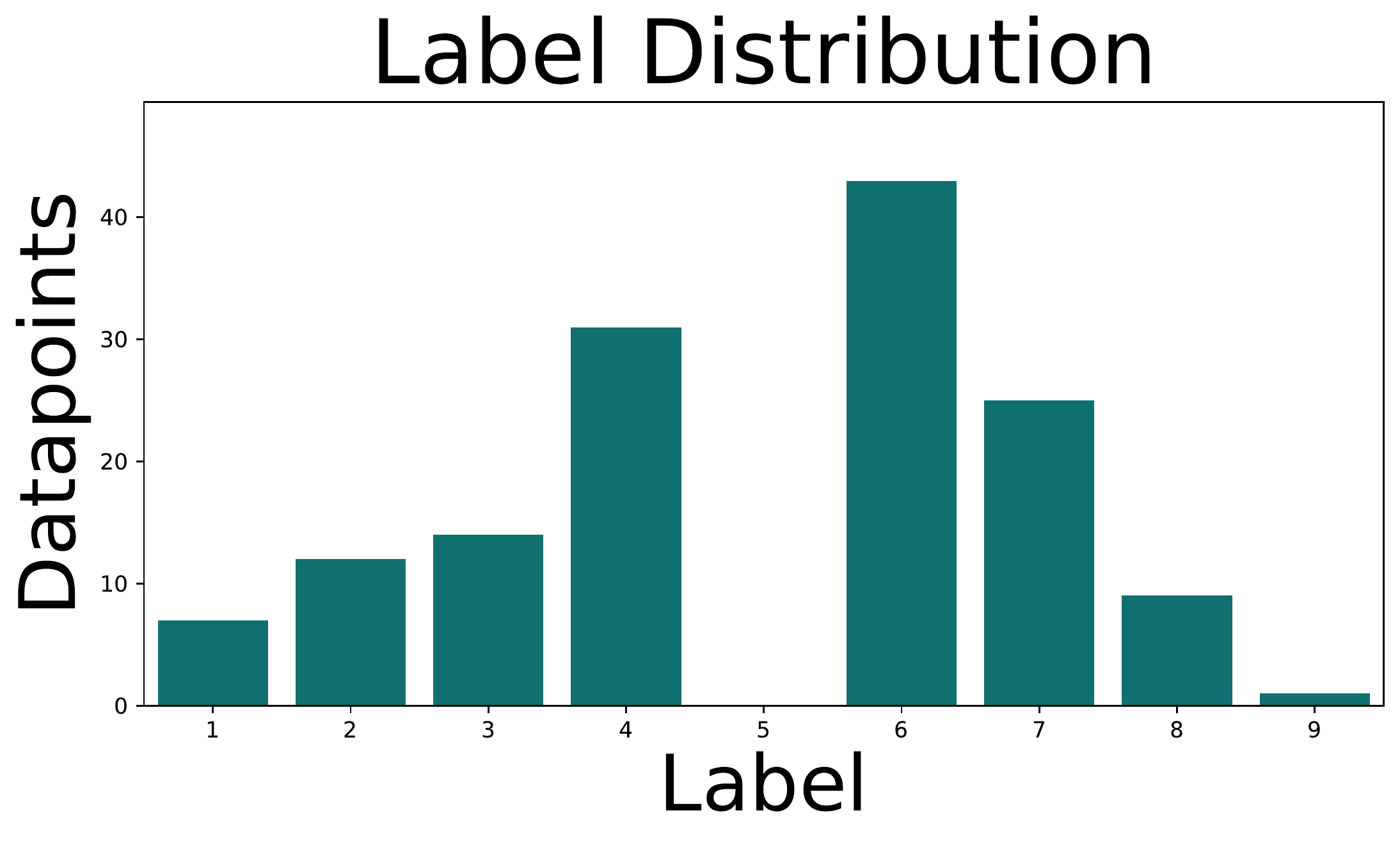}\label{fig:sub2}}\hskip1ex
\subfloat[Subject 12]{\includegraphics[width=0.13\textwidth]{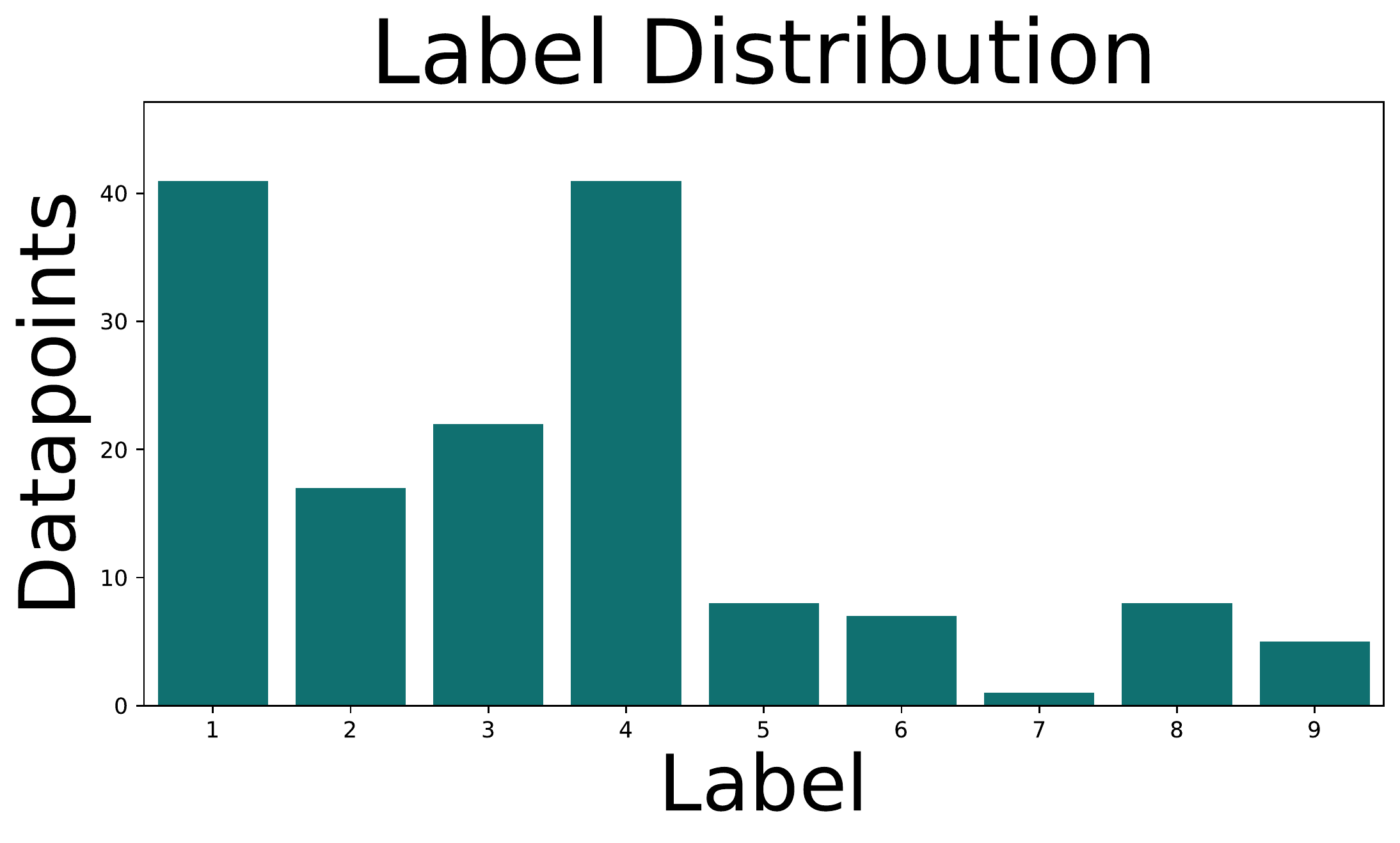}\label{fig:sub3}}\hskip1ex
\subfloat[Subject 13]{\includegraphics[width=0.13\textwidth]{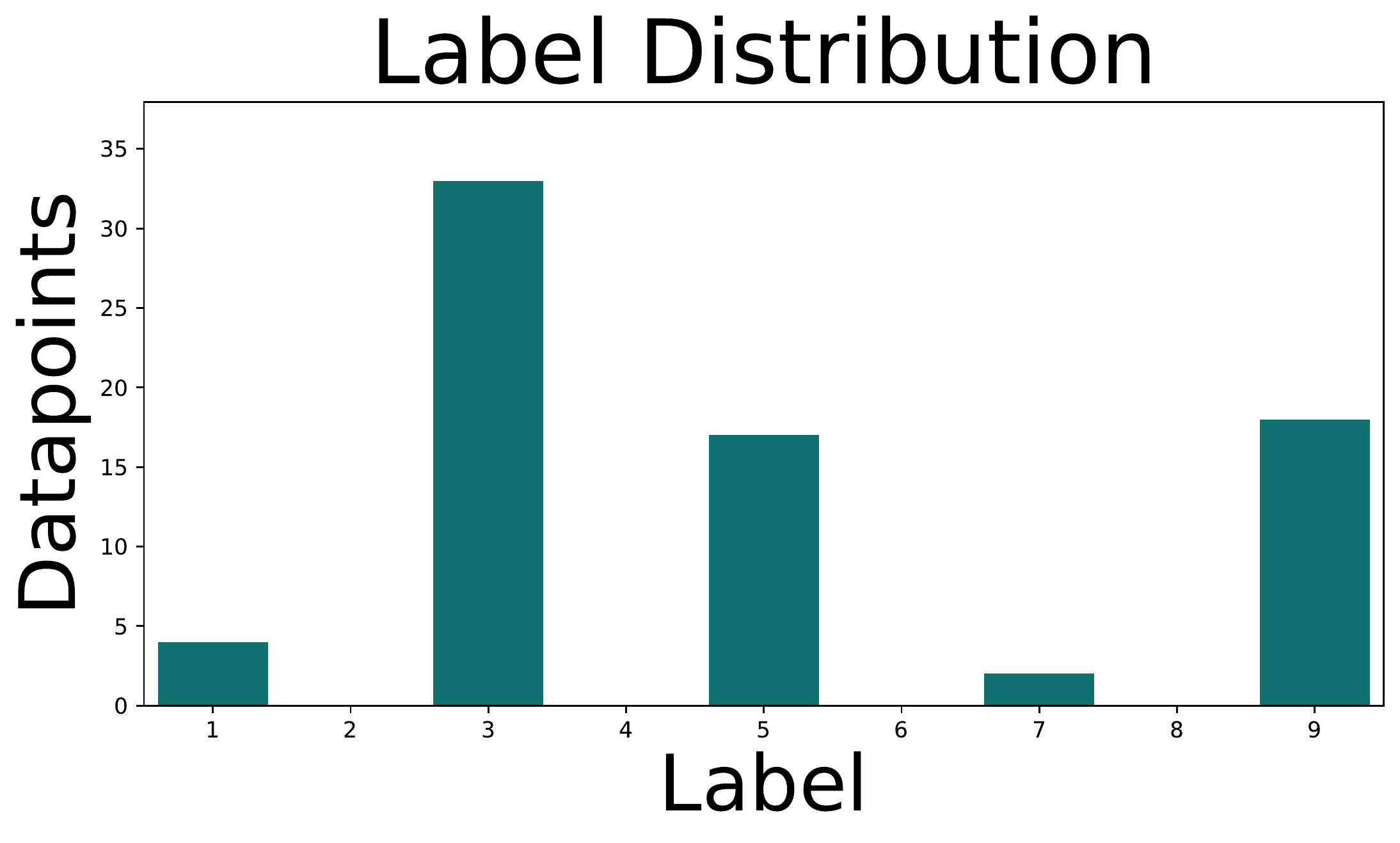}\label{fig:sub1}}\hskip1ex
\subfloat[Subject 14]{\includegraphics[width=0.13\textwidth]{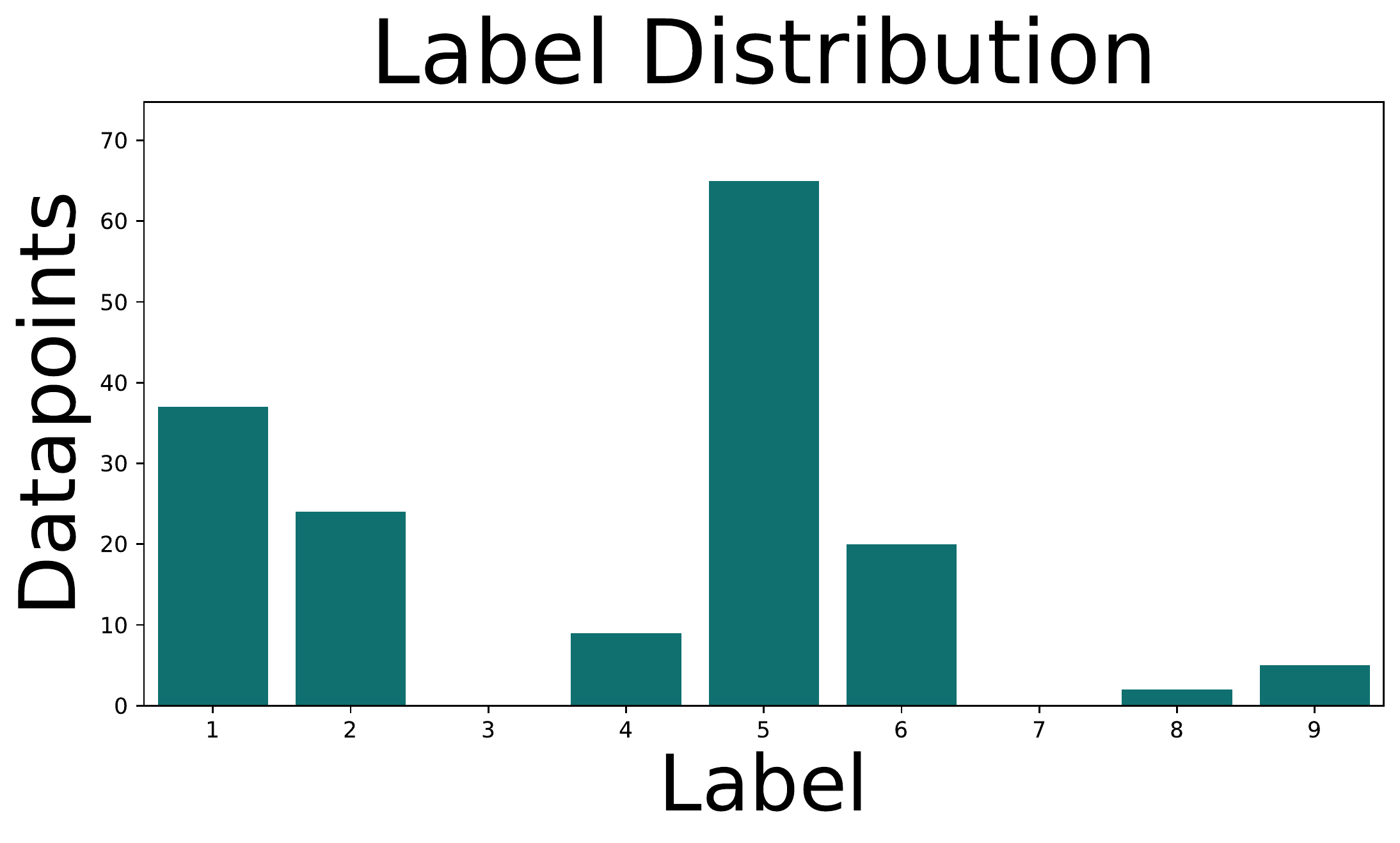}\label{fig:sub2}}\hskip1ex
\subfloat[Subject 15]{\includegraphics[width=0.13\textwidth]{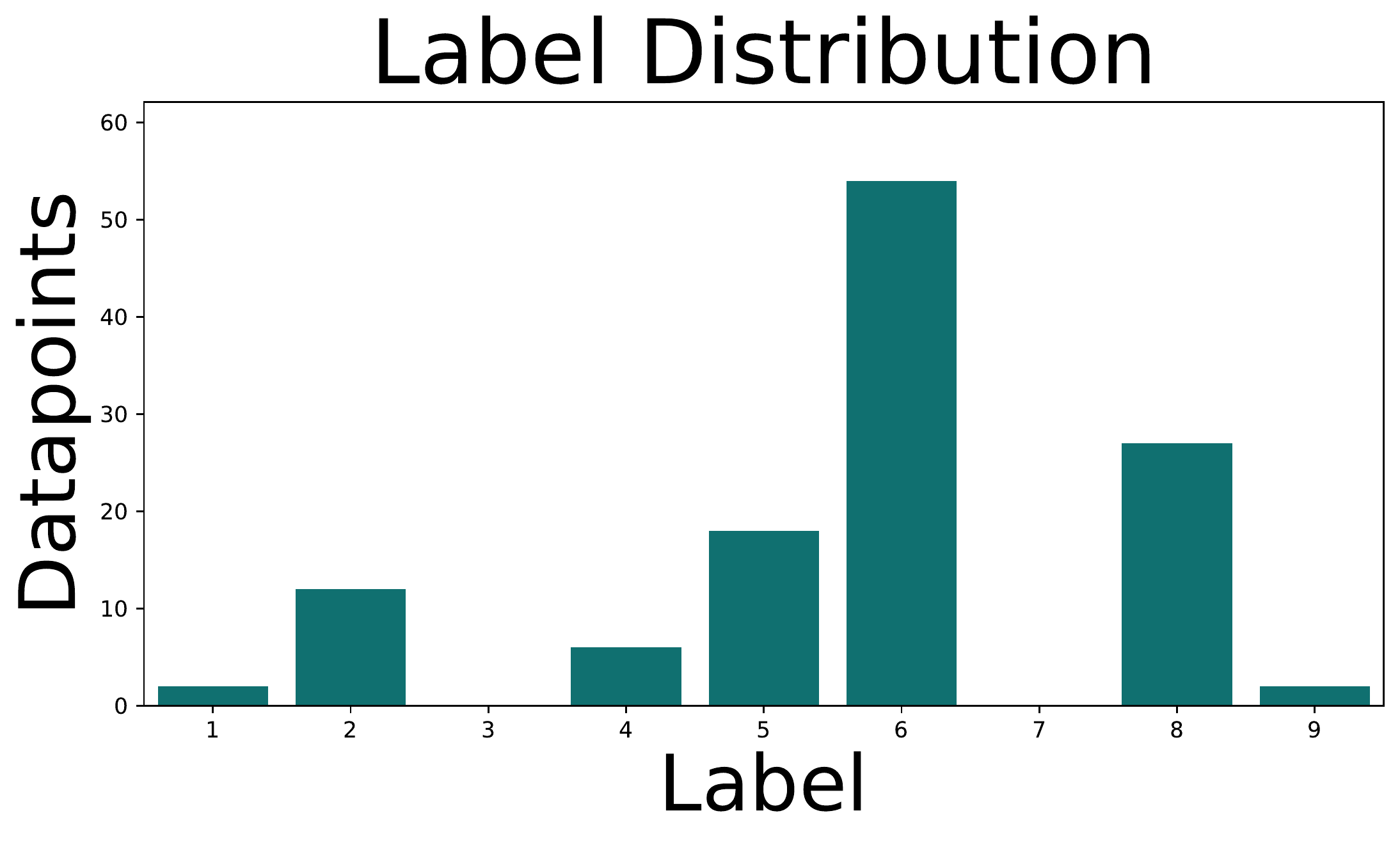}\label{fig:sub3}}\hskip1ex
\subfloat[Subject 16]{\includegraphics[width=0.13\textwidth]{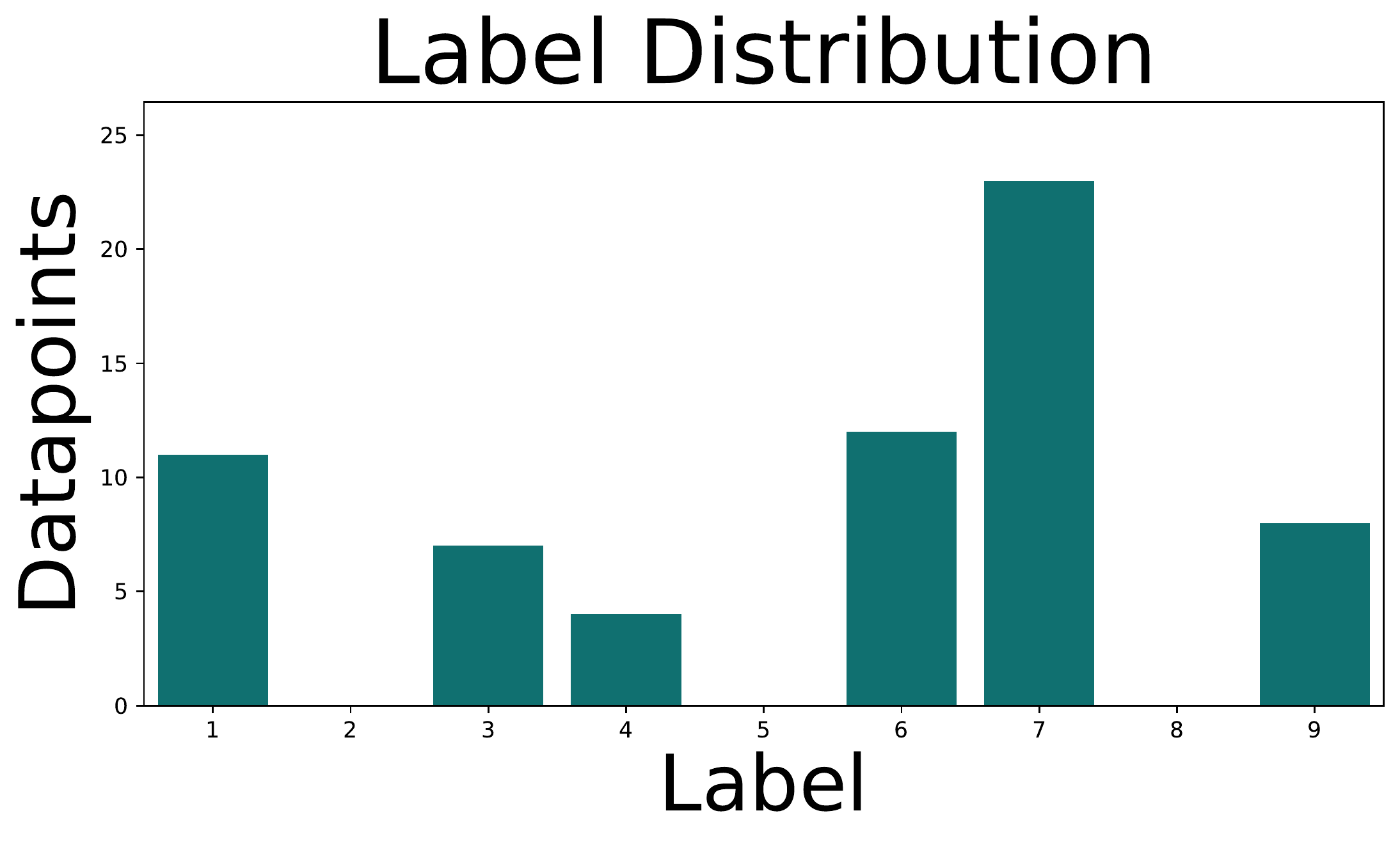}\label{fig:sub1}}\hskip1ex
\subfloat[Subject 17]{\includegraphics[width=0.13\textwidth]{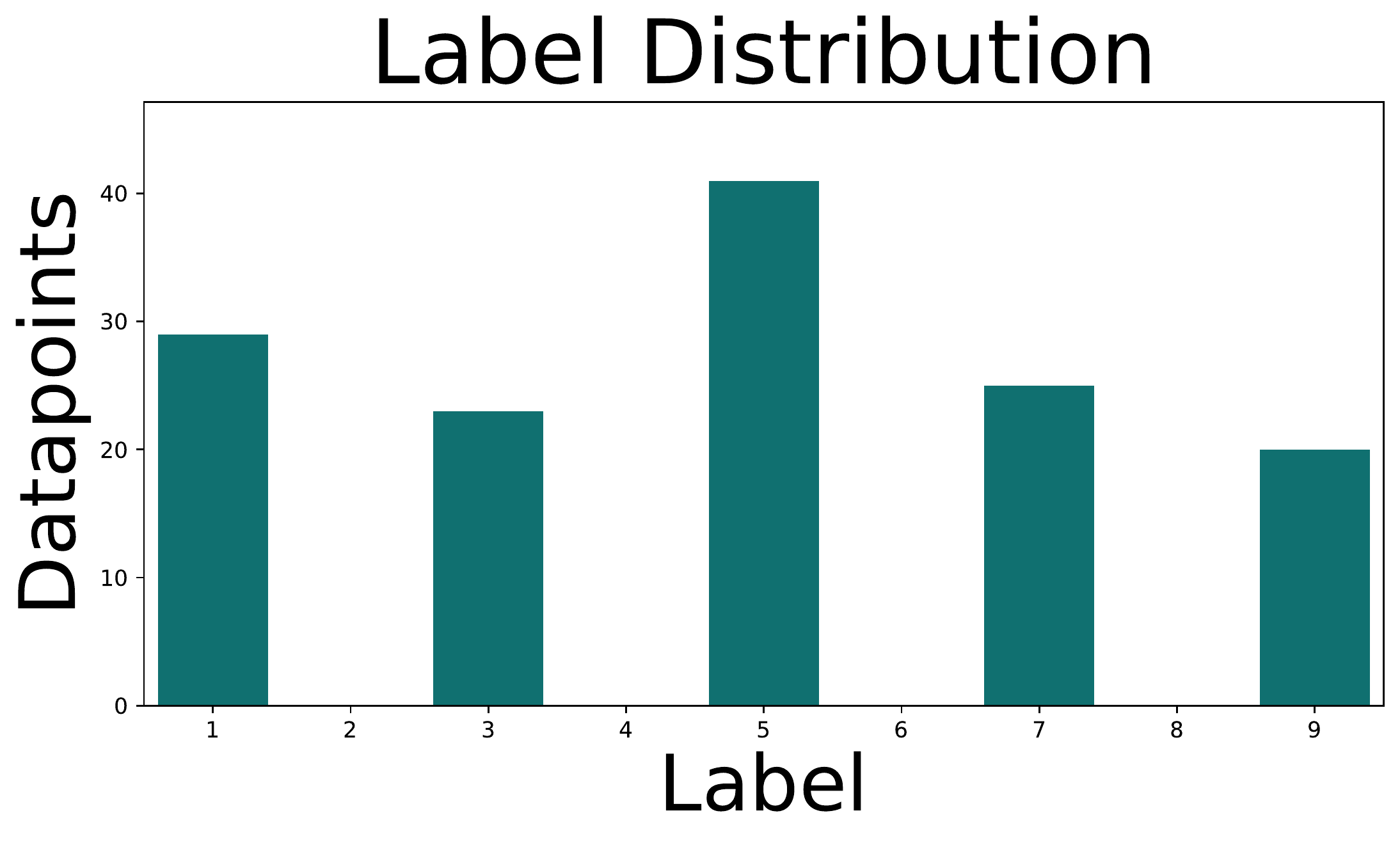}\label{fig:sub2}}\hskip1ex
\subfloat[Subject 18]{\includegraphics[width=0.13\textwidth]{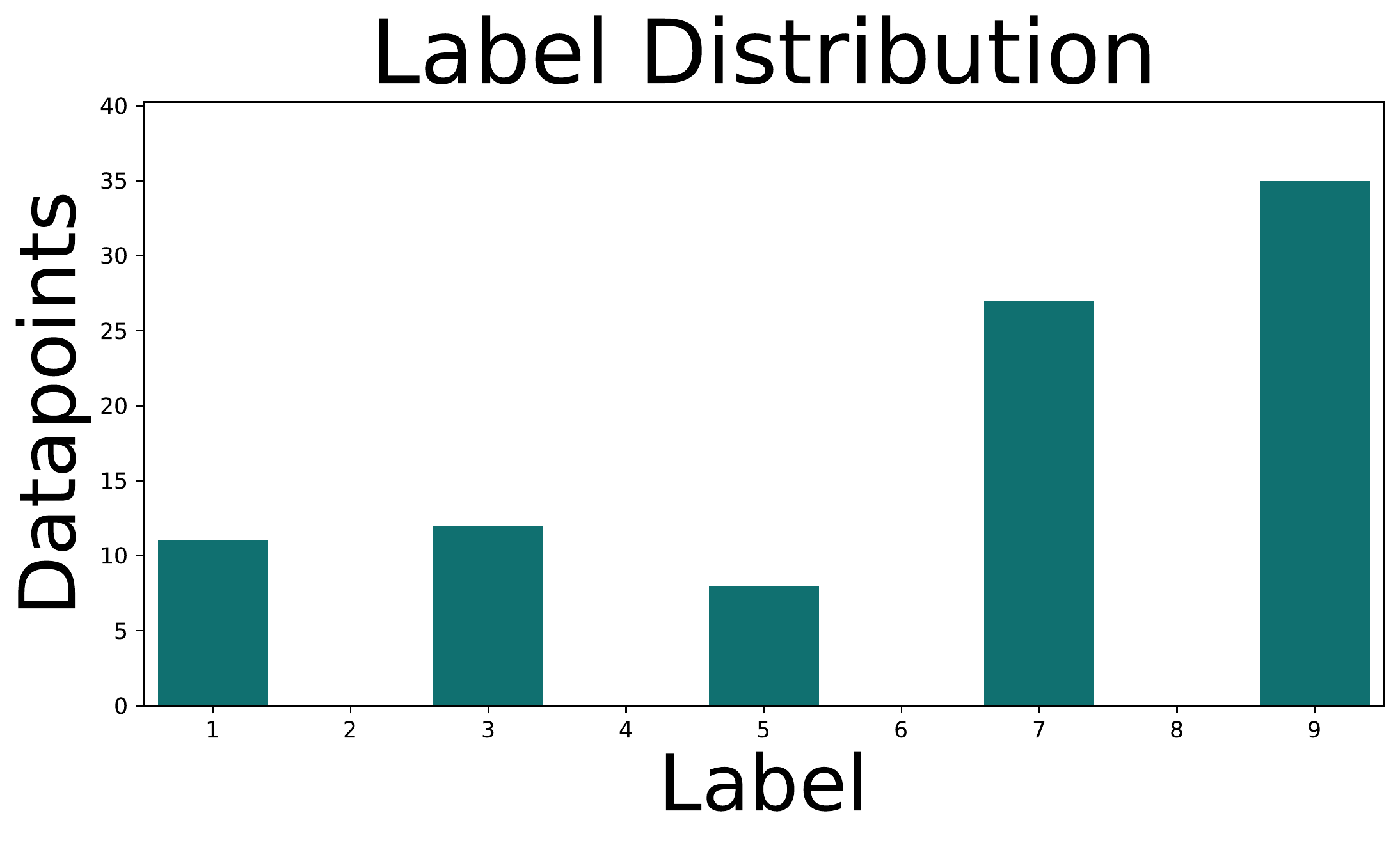}\label{fig:sub3}}\hskip1ex
% \end{figure*}%
% \begin{figure*}\ContinuedFloat
% \centering
\subfloat[Subject 19]{\includegraphics[width=0.13\textwidth]{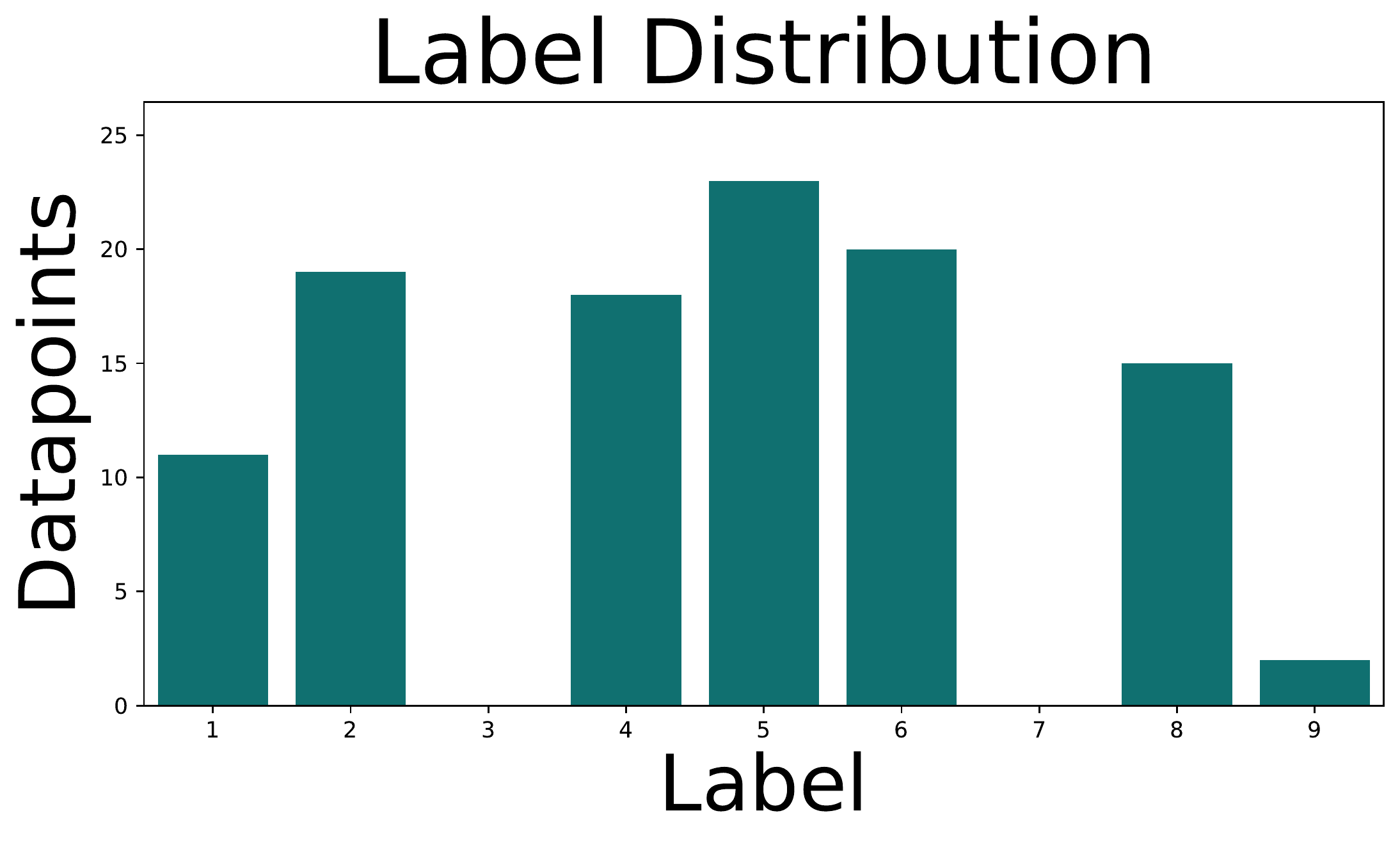}\label{fig:sub1}}\hskip1ex
\subfloat[Subject 20]{\includegraphics[width=0.13\textwidth]{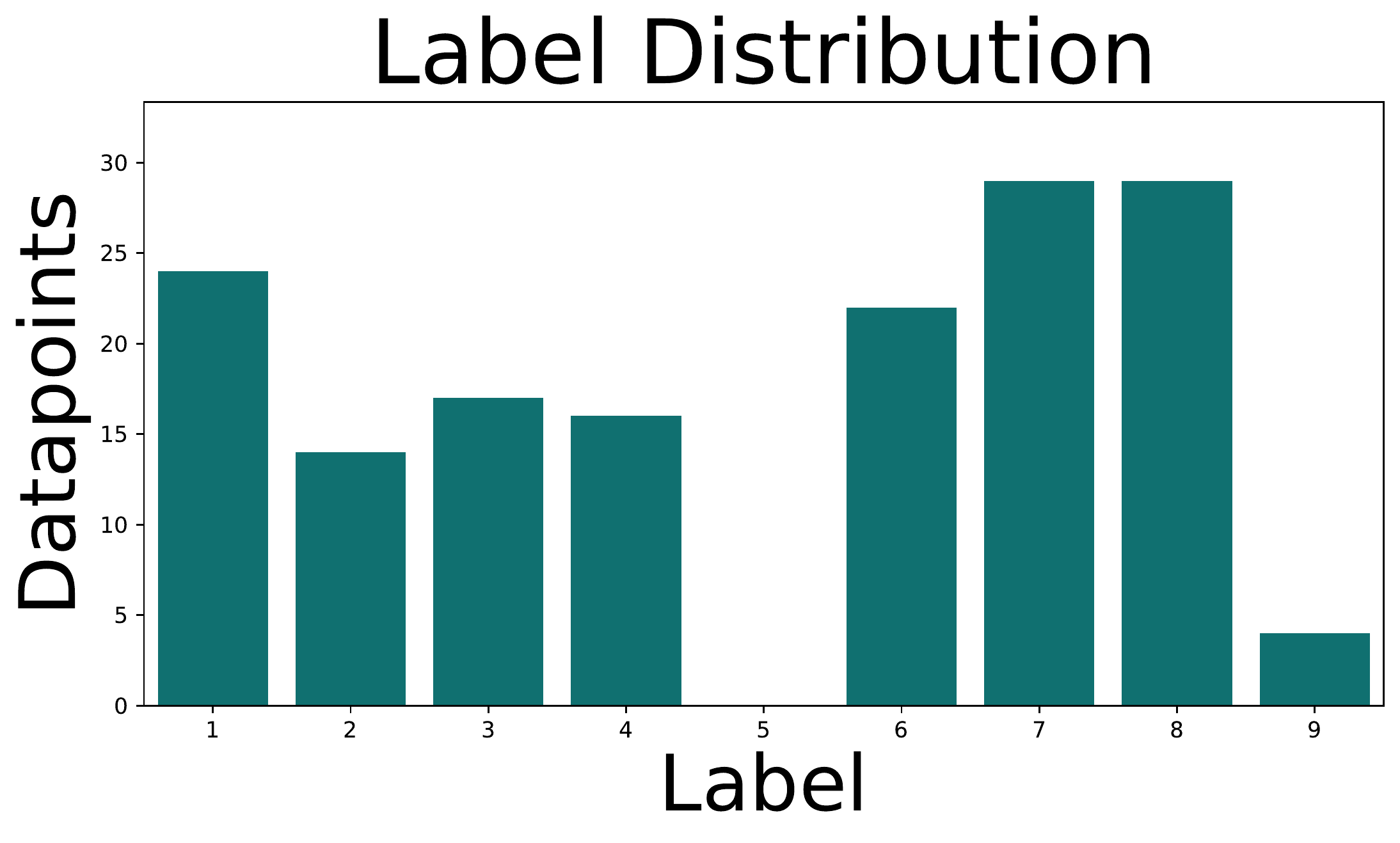}\label{fig:sub2}}\hskip1ex
\subfloat[Subject 21]{\includegraphics[width=0.13\textwidth]{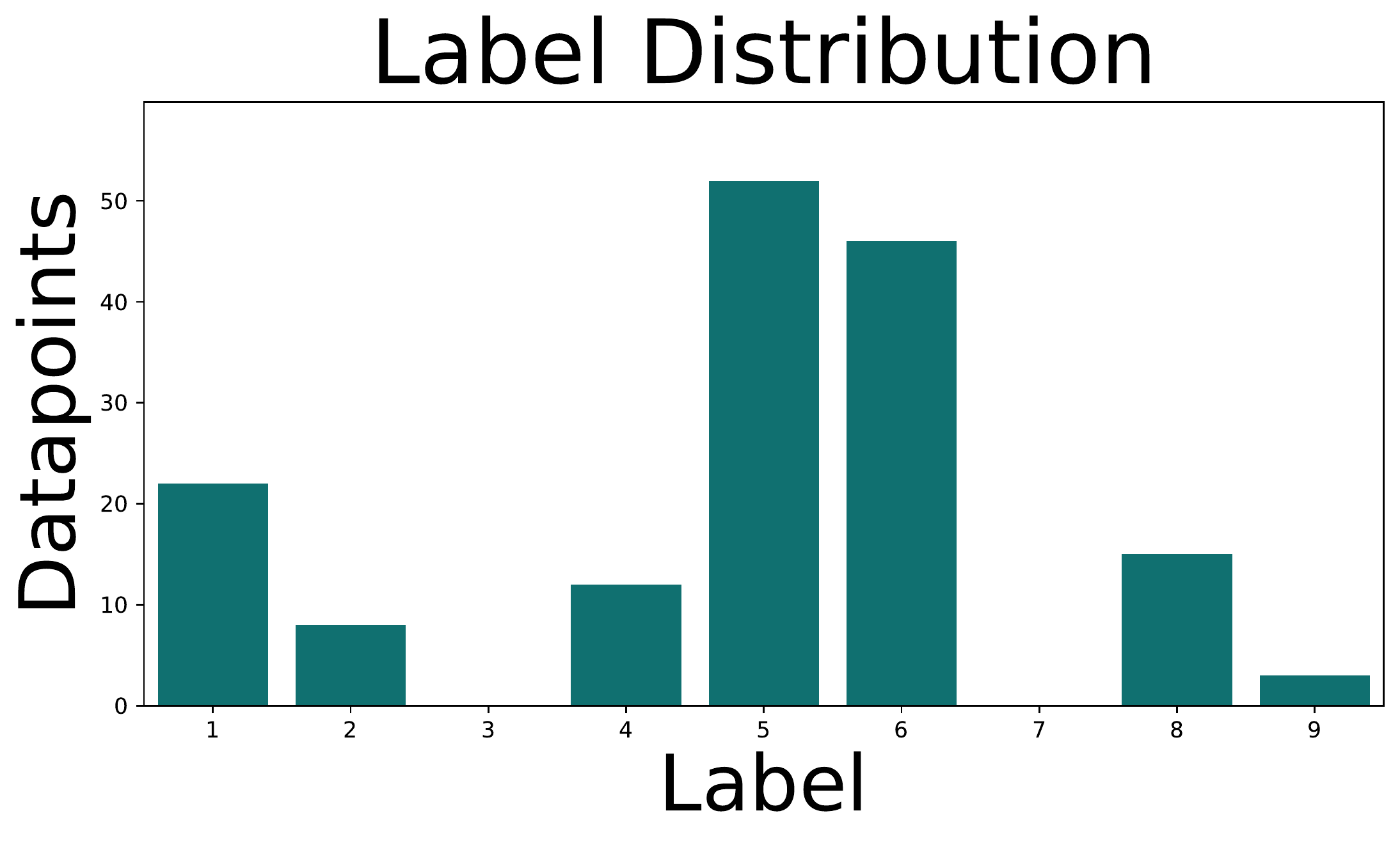}\label{fig:sub3}}
\caption{Self-reported cognitive load level distribution for each participant.}
\label{fig:scaled_label}
\end{figure*}

Participant cognitive load self-assessment ratings were used as ground truth labels in this study. As shown in Table~\ref{table_paas_level}, PAAS subjective cognitive load scores consist of 9 levels~\cite{paas}. A looping audio cue was generated every 10 seconds during the experiments to prompt the participants to verbally report their cognitive load, and a member of the research team recorded the reported scores. Figure \ref{fig:scaled_label} presents the distribution of the recorded output scores for all the participants.

\textcolor{black}{
% Regarding the frequency of recorded responses, we had to make a design choice to balance frequency of labels for the purpose of training machine learning models, and ensuring that the questions themselves do not impact the experiment significantly. Through a few pilot runs we came to a conclusion that 10 seconds is a reasonable interval since longer intervals would result in the participants to forget their state in the earlier parts of that segment, while shorter intervals could impact the experiments themselves.
Regarding the frequency at which the responses were recorded, we aimed to balance the frequency of labels for the purpose of training machine learning models, and ensuring that the questions themselves did not impact the experiment significantly. Through a few pilot trials, 10 seconds was found to be a reasonable interval, as longer intervals could lead participants to forget their earlier experiences during that segment, while shorter intervals could interfere with the experiment itself.}

\begin{figure*}
\centering
% \subfloat[EEG signal for high and low cognitive load]
{\includegraphics[width=0.33\textwidth]{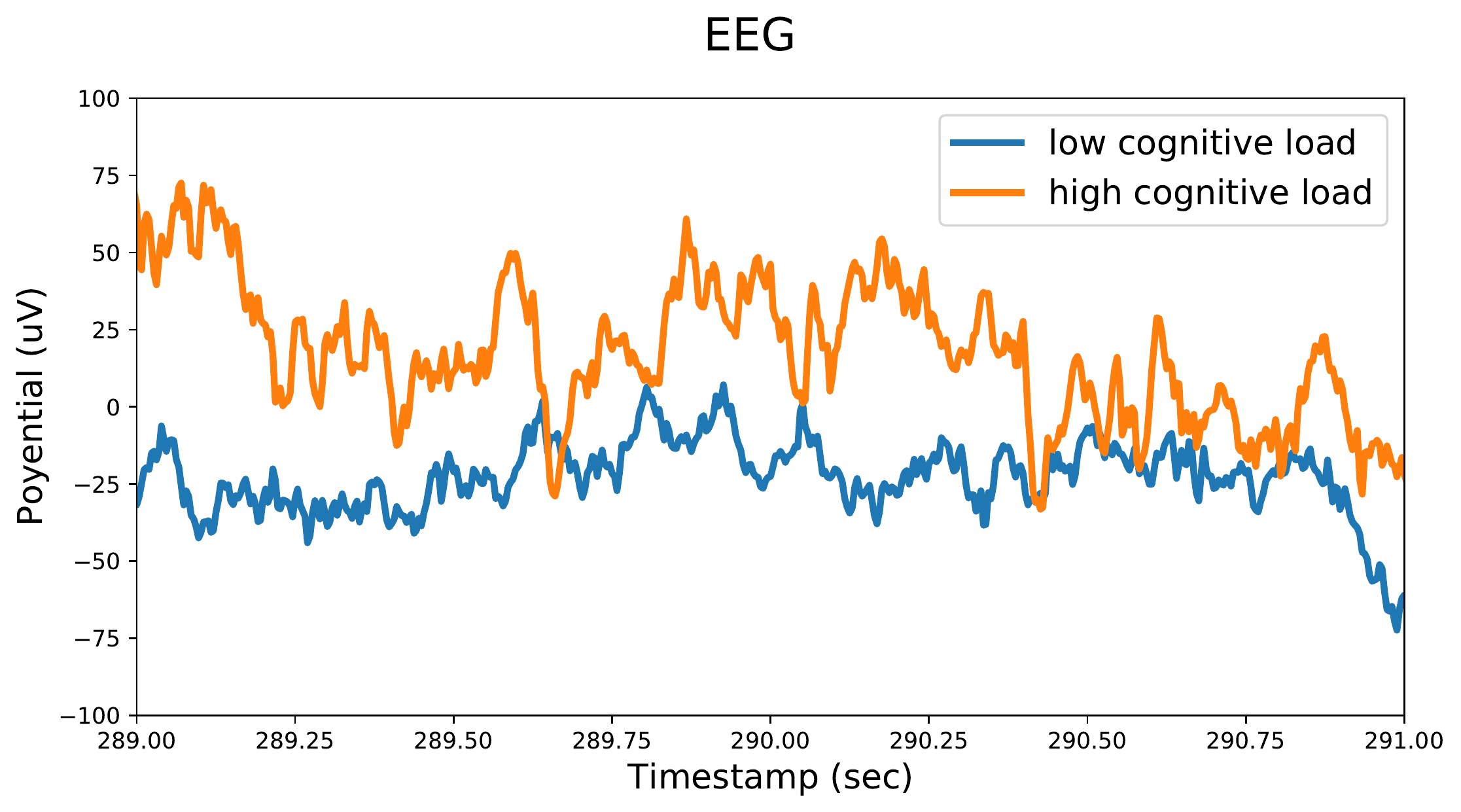}\label{fig:eeg_high_low}}\hskip1ex
% \subfloat[ECG signal for high and low cognitive load]
{\includegraphics[width=0.33\textwidth]{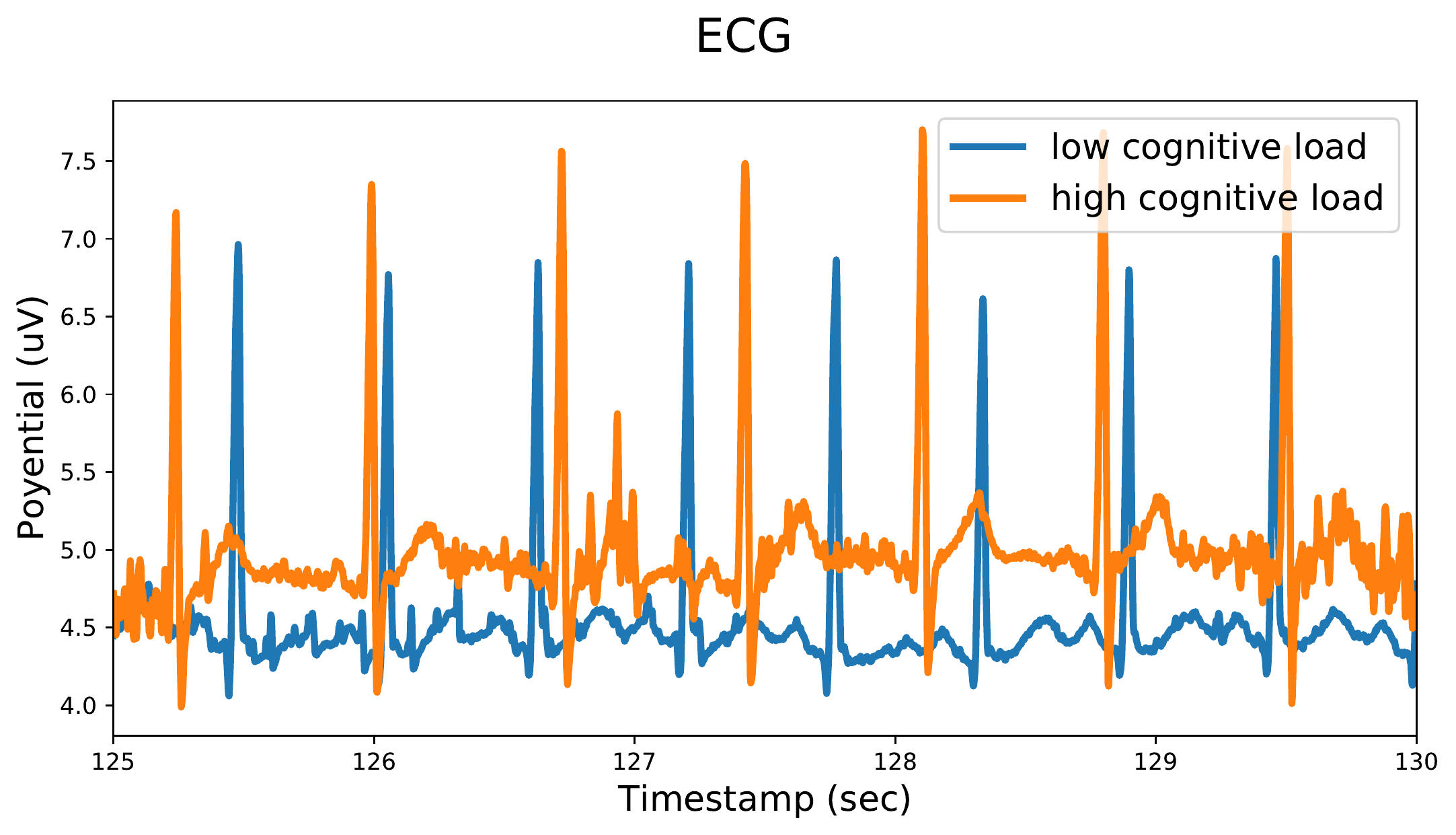}\label{fig:ecg_high_low}}\hskip1ex
% \subfloat[EDA signal for high and low cognitive load]
{\includegraphics[width=0.33\textwidth]{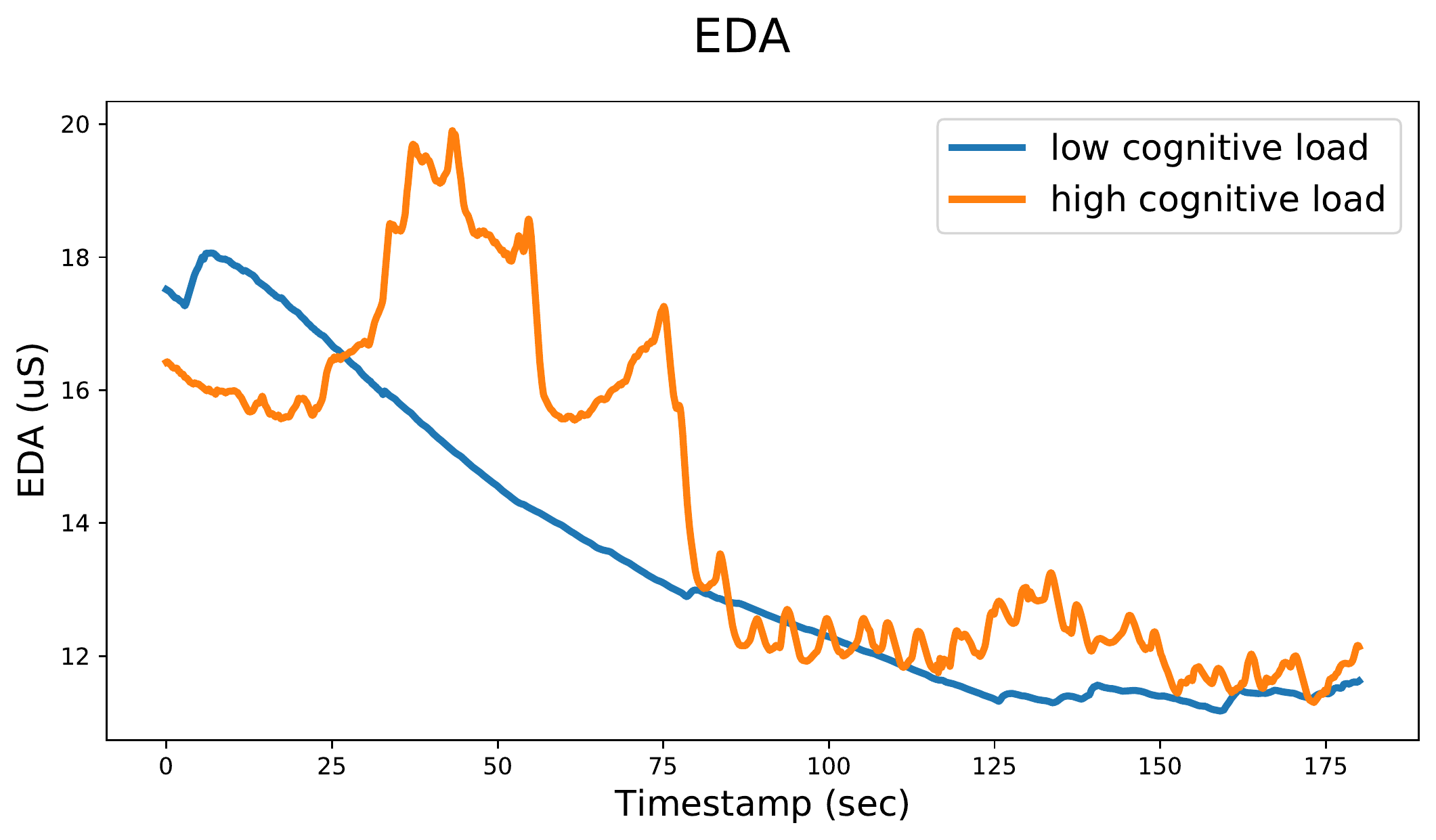}\label{fig:eda_high_low}}
% \subfloat[Pupil size signal for high and low cognitive load]
{\includegraphics[width=0.33\textwidth]{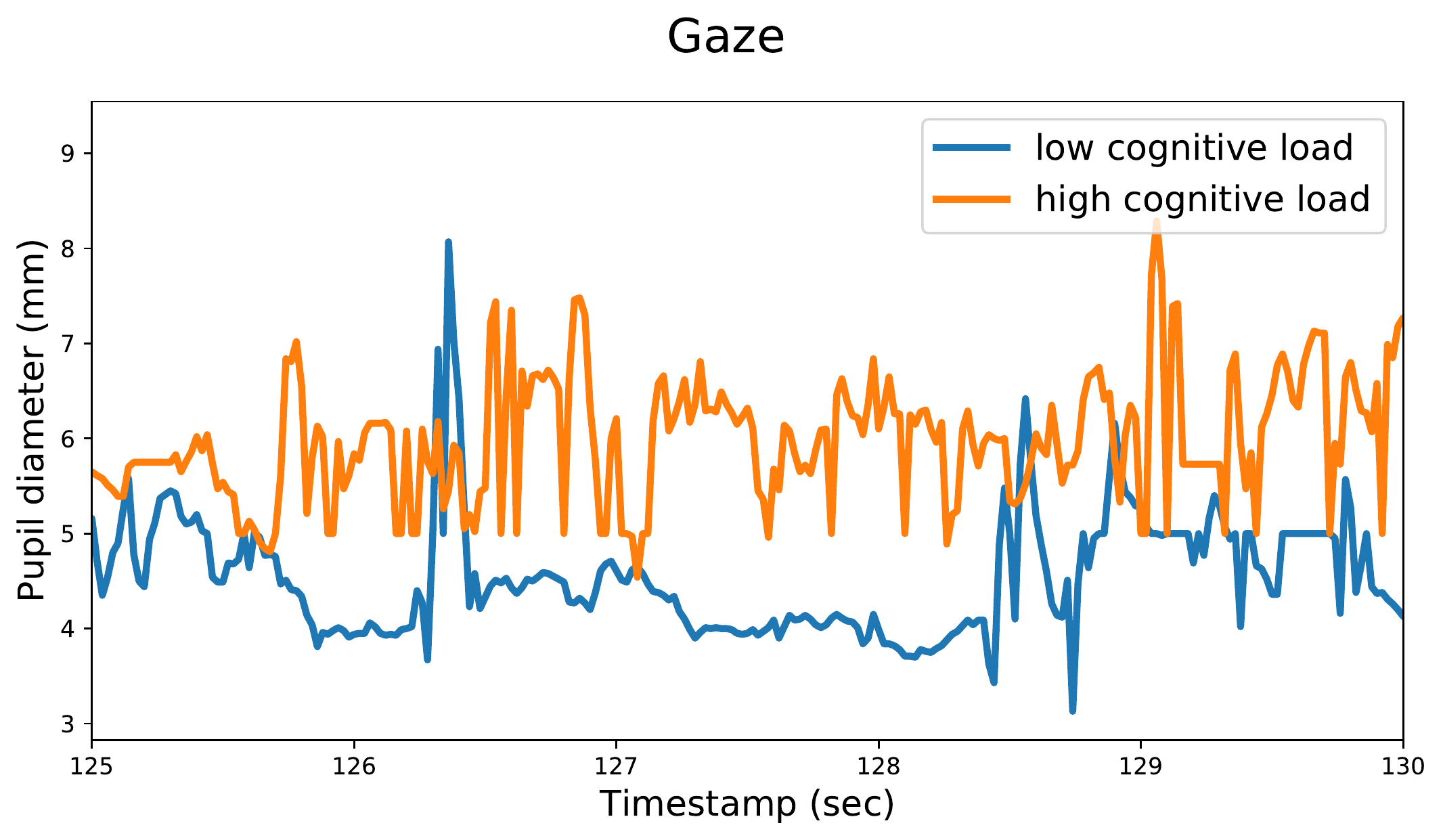}\label{fig:gaze_high_low}}
\caption{Examples of different signals in high and low cognitive load scenarios.}
\label{fig:high_low}
\end{figure*}

\subsection{Experiment Protocol}
\label{section:experiment_protocol}
% On the day of the experiment the 
Participants were given clear descriptions about the data collection protocol and equipment. 
% They were introduced to each sensor and how to wear them with the aid of photographs. 
% Participants were required to wear the sensors themselves due to Covid-19 restrictions, however, after all the sensors were attached, a team member had a look to ensure all the sensors were worn correctly. 
After careful sensor placement, participants were asked to sit in the driving seat of the simulator and the sensors were connected via Bluetooth to a data collection station. 
% To further ensure that the quality of the EEG, ECG, and EDA data were checked by visualizing. At this point, the reception of the data from each sensor was also checked.    
% The participants were then introduced to the driving simulator and explained how it works. 
% They were introduced to the looping auditory cue where they needed to verbally report their cognitive load every 10 seconds.
% The driving simulator was then calibrated and made sure all the components such as horn, steering wheel, brake, accelerator, indicators, hazard, gear etc. were working correctly. 
% In order to remove artifacts from EEG data, the participants were asked to perform the following eye movements 10 times each: left-right, up-down and blink. 
% After that, 
First, 3 minutes of baseline data was collected from each participant which could be used for future normalization of the signals. To become adapted to the simulated driving environment and make sure that participants had a clear understanding about the `low', `medium', and `high' complexity levels, and also to reduce the SAS level, participants were asked to perform the \textit{Orientation} scenario mentioned earlier in Table~\ref{table_virage_tasks}. 
Following every 3-minute driving scenario (see Table ~\ref{table_virage_tasks}), the participants were given a resting time of 2 minutes to allow them to rest, reduce the possibility of SAS, and come back to a relatively lower cognitive load state. Followed by the resting period, a 2-minute baseline was collected before each new scenario. The experiment flow is shown in Figure~\ref{fig:exp_flow}. During all these experiments, the wearable sensors discussed earlier in Section \ref{section:sensors} were used to record the respective signals from the participants. Figure \ref{fig:high_low} we illustrate a sample from each captured modality in both high and low cognitive load scenarios.

{\renewcommand{\arraystretch}{1.5}
\begin{table*}[!t]
\caption{Extracted features from each modality.}
\label{tbl:extracted-features}
\centering
\begin{tabular}{p{0.1\linewidth} p{0.65\linewidth} c}
\hline
\textbf{Modalities} & \textbf{Extracted features} & \textbf{Number of features} \\
 \hline \hline
EEG & 
% FFT (mean, maximum, minimum, and median), 
PSD (absolute, mean, maximum, minimum, median power), Spectral Entropy, Hjorth mobility and complexity, Lempel-Ziv Complexity, Higuchi fractal dimension, raw signal (mean, minimum, maximum, median, variance, and standard deviation) & 40 \\ 
\hline
ECG & RMSSD, MeanNN, SDNN, SDSD, CVNN, CVSD, MedianNN, MadNN, MCVNN, IQRNN, pNN50, pNN20, TINN, HTI, SD1, SD2 SD1/SD2, S, CSI, CSI\_Modifies, CVI, PIP, IALS, PSS, PAS, GI, SI, AI, PI, C1d, C1a, SD1d, SD1a, C2d, C2a, SD2d, SD2a, Cd, Ca, SDNNd, SDNNa, ApEn, SampEn, mean, median, standard deviation, skewness, kurtosis, entropy, interquartile range, area under curve, squared area under the curve, median absolute deviation & 53 \\
\hline
EDA & Mean, median, standard deviation, skewness, kurtosis, entropy, interquartile range, area under curve, squared area under the curve, median absolute deviation for raw data as well as phasic and tonic response & 30 \\
 \hline
Gaze & Pupil diameter (max, min, mean), Blink count, duration (max, mean), Fixation count, duration (max, min, mean), dispersion (max, min, mean), Saccade count, duration (max, min, mean), amplitude (max, min, mean), peak velocity (max, min, mean), peak acceleration (max, min, mean), peak deceleration (max, min, mean), direction (max, min, mean) & 32\\
\hline
\end{tabular}
\end{table*}}

\subsection{Dataset Release}
We make the dataset public at:
% \href{https://github.com/Prithila05/CL-Drive.git}{https://github.com/Prithila05/CL-Drive.git}
\href{https://github.com/Prithila05/CL-Drive}{https://github.com/Prithila05/CL-Drive}

% \href{https://borealisdata.ca/privateurl.xhtml?token=505ef1f5-18f8-407b-a9aa-31c964208005}{https://borealisdata.ca/privateurl.xhtml?token=505ef1f5-18f8-407b-a9aa-31c964208005}

% \subsection{\hl{Post-Hoc Questionnaire}}

\section{Data Processing} \label{section:data_processing}
In this section, we explain the data pre-processing steps for each signal type, followed by feature extraction. Next, we describe data normalization, which is followed by a description of the baseline classifiers used for benchmarking. 

\subsection{Pre-processing}
% In this section, we will discuss the data pre-processing steps for all the modalities and how we segmented our data. 
The cognitive load scores were collected at 10-second intervals during the 3-minute driving scenarios. We segment each recording into 18 segments of 10 seconds each. These segments will later be used for feature extraction or fed directly into the deep learning models. 
% \textcolor{black}{The initial signals undergo pre-processing, and any instances of missing values are appropriately handled. Missing data problems in datasets can be efficiently handled by latent factor analysis \cite{wu2022prediction, wu2022double}. It identifies the underlying structure in the data by capturing the shared patterns and relationships among the observed variables assuming that the observed variables are influenced by a smaller number of latent factors or unobservable variables. By estimating these latent factors, missing values can be predicted. Other simpler interpolation techniques such as polynomial interpolation can also be used to approximate the values in between known data points, effectively filling in the gaps caused by missing data.}

% , and features were collected for each segment. 
% Among the 23??? participants, 3 participants data is incomplete due to wifi and device issues. For participant 13, we only have data for complexity level 3, 4, 5, 6, and 9. For participant 16, we have data for complexity level 1, 5, 6 and 9. And for participant 18, we have data for complexity level 1 to 6. We have included all the participants data in for training the classifiers.    

\noindent \textbf{EEG.} 
% The EEG data collected using the EEG headband experienced some noise, artifacts, and interruption. 
To remove noise and artifacts from EEG, we used 
% signal from 4 channels namely TP9, AF7, AF8 and TP10, were filtered using the 
a Butterworth 2\textsuperscript{nd} order bandpass filter with a passband frequency of 0.4 to 75 \textit{Hz}. A notch filter with a quality factor of 30, was used to remove the powerline noise at a frequency of 60 \textit{Hz}. 
\textcolor{black}{
% For EEG data, there were no noteworthy instances of data packet loss. However, 
Due to some Bluetooth problems, the device experienced a few disconnections lasting approximately 30 seconds during the experiments. Given the duration of these gaps, using imputation methods would not be suitable.
% , as they would not generate accurate values which might adversely affect the accuracy of the model's performance. Hence, 
We therefore excluded segments with missing data from our dataset.}
% Figure~\ref{fig:eeg_raw_vs_filtered} shows the EEG signal, before and after applying the butterworth bandpass filter and removing powerline noise.  

% \begin{figure}
%     \begin{center}
%     \includegraphics[width=0.7\columnwidth]{Figures/eeg_raw_vs_filtered.pdf} 
%     \caption{Raw EEG signal before and after applying butterworth bandpass filter}
%     \label{fig:eeg_raw_vs_filtered}
%     \end{center}
% \end{figure}

% \hl{Confirm with Anubhav}

\noindent \textbf{ECG.} Artifacts such as high-frequency noise, EMG noise, T noise interference, etc., were then filtered out using a Butterworth bandpass filter with passband frequency of 5 to 15 \textit{Hz}, which also enables us to obtain maximum \textit{QRS} energy \cite{thakor1983optimal, goovaerts1976digital}. \textcolor{black}{The ECG signals experienced missing values occasionally, which we imputed using simple 5\textsuperscript{th} order polynomial interpolation.} 

\noindent \textbf{EDA.} We then used a lowpass butterworth filter with a cut-off frequency of 3 \textit{Hz} to remove the unwanted noise. A highpass butterworth filter with a cut-off frequency of 0.05 \textit{Hz} was used to decompose the filtered EDA signal to tonic skin conductance level and phasic skin conductance response to isolate the slow changing levels and rapid changing peaks in the signal \cite{lim1997decomposing}. \textcolor{black}{
% For EDA 
There were some missing values which we replaced with a sample-and-hold strategy given the simplicity of EDA signals in comparison to ECG.} 

% \hl{EDA if possible include diagram of phasic and tonic}

\noindent \textbf{Gaze.} The device measures saccade, fixation, pupil diameter, blink count, and blink duration based on 2D gaze coordinates ($x$,$y$ pixel coordinates in screen space) for both left and right eyes, 3D gaze coordinates ($x$,$y$,$z$ coordinates in mm in camera space), 3D gaze direction (vector units), gaze velocity in degrees per second ($^\circ$/s), and gaze acceleration in degrees per second squared ($^\circ$/$s^2$). We directly use the high-level metrics in our study. \textcolor{black}{For the missing values in Gaze data, we used the sample-and-hold method similar to that of EDA due to the straightforward nature of the signals.}

% The device measures pupil diameter, 2D gaze coordinates ($x$,$y$ pixel coordinates in screen space) for both left and right eyes, 3D gaze coordinates ($x$,$y$,$z$ coordinates in mm in camera space), 3D gaze direction (vector units), gaze velocity in degrees per second ($^\circ$/s), and gaze acceleration in degrees per second squared ($^\circ$/$s^2$). The events such as fixations, and saccades are extracted from the raw data using the Tobii I-VT (Velocity Threshold Identification) filter \cite{olsen2012tobii}. The blinks were measured using a blink detection algorithm\footnote{https://imotions.com/blog/learning/product-guides/what-are-r-notebooks-in-imotions/}.  The events such as fixations, saccades are extracted from the raw data using the Tobii I-VT (Velocity Threshold Identification) filter \cite{olsen2012tobii}.  

% \begin{figure}
%     \begin{center}
%     \includegraphics[width=0.7\columnwidth]{Figures/high_low_cogload.pdf} 
%     \caption{Signal representation of high and low cognitive load}
%     \label{fig:high_low_cogload}
%     \end{center}
% \end{figure}

\begin{figure*}
    \begin{center}
    \includegraphics[width=0.8\linewidth]{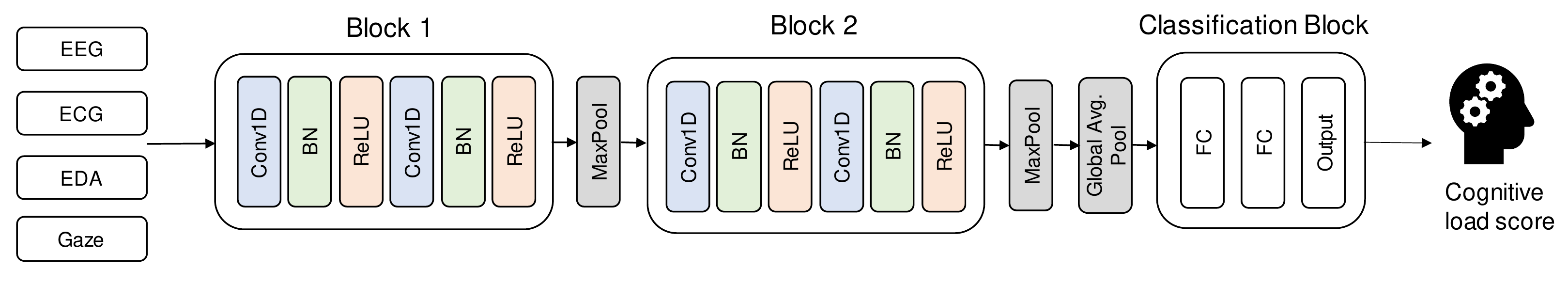}
    \caption{VGG-style network used as for benchmarking in this study.}
    \label{fig:vgg-like_network}
    \end{center}
\end{figure*}

\begin{figure*}
    \begin{center}
    \includegraphics[width=0.8\linewidth]{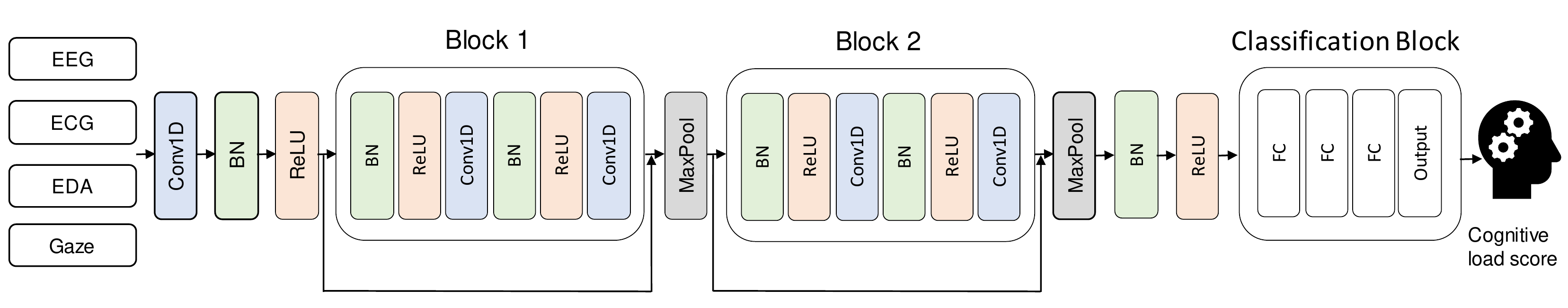}
    \caption{ResNet-style network used as for benchmarking in this study.}
    \label{fig:resnet-like_network}
    \end{center}
\end{figure*}

\subsection{Feature Extraction} \label{section:feature_extraction}
Different features were extracted to train our machine learning algorithms. In this section, we describe the features extracted from each modality. 
% \textcolor{black}{In order to not add a new layer of complexity in our pipeline and to do a fair comparison among all the machine learning models, we did not use any feature selection method and used all the extracted features in our training.} 
\textcolor{black}{No feature selection methods were used.} 
% Please note that 
Given the focus of our work on BCI (other modalities play auxiliary roles in the multimodal setups), we describe the EEG-related features in more depth below.
% The features are described as follows.

\noindent \textbf{EEG.} 
% There are lots of datapoints in EEG signal which makes it  difficult to interpret meaningful information. Therefore, it is important to extract useful features from the signal for better classification performance \cite{SUBASI2019193, de2020feature}. 
% For our experiment, 
We extract 40 features from both time and frequency domains from each channel for each 10 seconds segment. 
% Some features are extracted for 5 EEG frequency bands: Delta (0.5-4 \textit{Hz}), Theta (4-8\textit{Hz}), Alpha (8-12 \textit{Hz}), Beta (12-31 \textit{Hz}) and Gamma (31-75 \textit{Hz}) while other features are extracted for the entire length of signal for each channel. 
The details of the features are given below:
\begin{enumerate}
  \item \textbf{Power Spectral Density (PSD):} PSD measures the power of the EEG signal. To calculate this feature we use the Welch's method from 0.5 \textit{Hz} to 75 \textit{Hz} frequency, for each frequency band, Delta (0.5-4 \textit{Hz}), Theta (4-8\textit{Hz}), Alpha (8-12 \textit{Hz}), Beta (12-31 \textit{Hz}), and Gamma (31-75 \textit{Hz}). We then measure the absolute, mean, maximum, minimum, and median power of the measured PSD. 
%   \hl{include a diagram before and after notch filter}
%   \hl{Figure XYZ shows the PSD of EEG signal before and after applying the notch filter.}
%   \item \textbf{Fast Fourier Transform (FFT):} Time series signal can be represented as a collection of sinusoids and plotted in the frequency-power spectrum \cite{nussbaumer1981fast}. We performed fast Fourier transform to convert the signal in frequency domain from where we extracted features like mean, maximum, minimum and median of each frequency band. 
  \item \textbf{Spectral entropy:} Spectral entropy (SE) of a time series signal is derived from normalized Shannon's entropy \cite{shannon1948mathematical} and can be used to determine the complexity of a signal. The formula of SE can be derived from normalized PSD or probability distribution \textit{p(i)} of the signal as
  \begin{equation}
  \label{eqn:spectral_entropy}
    SE = \sum_{i=1}^n p(i)\ln p(i).
  \end{equation} 
  We calculate $SE$ for all 5 bands of EEG.
  \item \textbf{Hjorth mobility and complexity:} 
%   Hjorth parameters are measured in time domain. 
  % Hjorth mobility returns the square root of the variance of the first derivative of the signal to the variance of the signal itself. 
  % Hjorth complexity on the other hand is the ratio of the mobility of the first derivative of the signal to the mobility of the signal itself which determines how much the signal corresponds to a sine wave \cite{hjorth1970eeg}. 
  Both Hjorth mobility and complexity of a time series signal determine aspects of the signal complexity \cite{hjorth1970eeg}, where the variations in signal frequency and amplitude are represented by Hjorth mobility and complexity respectively. These two measurements can be jointly used to capture the dynamic behavior of signals. Hjorth mobility and complexity are respectively calculated as:
%   \begin{equation}
%   \label{eqn:hjorth_mobility}
%     Hjorth_M = \sqrt{\frac{var(\frac{dy(t)}{d(t)})}{var(y(t))}}
%   \end{equation} 
  \begin{equation}
  \label{eqn:hjorth_mobility}
    Hjorth_M = \sqrt{\frac{\sigma^2(\frac{dy(t)}{d(t)})}{\sigma^2(y(t))}}
  \end{equation}   
and
  \begin{equation}
  \label{eqn:hjorth_complexity}
    \textit{\text{Hjorth}}_C = \frac{\textit{\text{Hjorth}}_M (\frac{dy(t)}{d(t)})}{\textit{\text{Hjorth}}_M (y(t))},
  \end{equation}
where $y(t)$ represents the signal and $\sigma^2$ is the variance operator.
%   We calculate both Hjorth mobility and complexity for the entire signal length for each channel.
  \item \textbf{Lempel-Ziv complexity:} 
%   Lempel-Ziv complexity is calculated for each channel. 
  % This feature aims to measure the diversity of the pattern of the signal \cite{lempel1976complexity}.
  Lempel-Ziv complexity (LZC) is a measure that also determines the complexity of a signal \cite{lempel1976complexity}. To apply the LZC algorithm to an EEG signal, the signal first needs to be binarized by the median or mean value of the entire signal. The resulting binary sequence can then be analyzed the using LZC algorithm to find any randomness \cite{kaspar1987easily}. 
  \item \textbf{Higuchi fractal dimension:} 
  % This calculates the fractal dimension of the signal in time domain which is provides the complexity of the signal \cite{higuchi1988approach}.
  Higuchi fractal dimension (HFD) is a non-linear method that can capture changes in time-series signals by measuring the complexity in time domain \cite{higuchi1988approach}. Prior studies have shown promising result using HFD with EEG signal in the past \cite{al2017higuchi, shamsi2021higuchi}. 
%   It is calculated for each EEG channel.
  \item \textbf{Statistical features:} In addition to the more sophisticated features mentioned above, we also extract simple statistical features namely mean, minimum, maximum, and median from the signal in time domain.% for each channel.  
\end{enumerate}
The complete list of EEG features is summarized in Table \ref{tbl:extracted-features}.
% \hl{confirm with Anubhav}

\noindent \textbf{ECG.} 
% Various features were extracted from both time and frequency domains for each 10 second segment of ECG. Statistical features such as minimum, maximum, standard deviation, etc. are extracted along with time domain features such as heart-rate variability (HRV), R-R intervals, NN20, NN50, etc. Some of the frequency domain features for EDA that we used to train our classifier were ultra-low-frequency (ULF), very-low-frequency (VLF), low-frequency (LF), etc. 
We extract various commonly used 
% 53
% time domain and non-linear 
% features along with statistical 
features from ECG \cite{shaffer2017overview}. 
% These include RR-based features 
% (i.e., based on the time between two successive R-waves), 
% HRV-based, and general statistical features. Examples include the popular pNN50 (proportion of RR intervals greater than 50ms, out of the total number of RR intervals), pNN20 (the proportion of RR intervals greater than 20ms, out of the total number of RR intervals), and others. 
The full list of features extracted from ECG is presented in Table \ref{tbl:extracted-features}.  
\textcolor{black}{All the features are extracted using the 
Neurokit2\footnote{\textcolor{black}{https://neuropsychology.github.io/NeuroKit/functions/hrv.html}} library. Please visit the library for further details.
% and the detail description of all the features can be found in there website.
}

\noindent \textbf{EDA.} 
% The tonic skin conductance level and phasic skin conductance response were segmented into segmented using 10 second window and features were extracted from each window. Minimum, maximum, standard deviation, skin-conductance response (SCR) peaks, SCR rise- and recovery-time, etc., were extracted as statistical and time domain features. 
% Similar to ECG, 
We extract a number of features from EDA.
% using the same euroKit2 library mentioned above. 
% A total of 30 statistical features are collected from raw EDA data 
These include statistical features from the raw data as well as phasic and tonic responses which are calculated by decomposing the EDA signal. The complete list of features extracted from EDA is presented in Table \ref{tbl:extracted-features}. 
% The features are given below:
% \begin{enumerate}
%     \item \textbf{Statistical features:} Mean, median, standard deviation, skewness, kurtosis, entropy, interquartile range, area under the curve, squared area under the curve, and median absolute deviation are calculated from raw EDA signal, as well as phasic and tonic responses.   
% \end{enumerate}

\noindent \textbf{Gaze.} For gaze analysis, we extract statistical features for each 10-second segment. The details of all the features are given in Table \ref{tbl:extracted-features}.

\subsection{Normalization} \label{section:normalization}

% \subsubsection{Data}
To reduce the variability \textit{between subjects}, which is a common phenomenon when recording such data, we divide each feature value with its corresponding average value from the baseline. Second, to reduce the variability \textit{within subjects}, we perform z-score normalization \cite{sarkar2020self} \textcolor{black}{following prior works such as \cite{kalamaras2017interactive, li2020multi, chen2022graph}}.

% We notice a considerable "variance among the data collected from the participants". To minimize that, we have divided each features with the feature extracted from the baseline. Furthermore, for each modality, we applied z-score normalization on the features after splitting the data into train and test set using the average and standard deviation of the train set. 

% \subsubsection{Labels}
% As discussed earlier, our experiment was designed to cover the range of 9 cognitive load complexity levels to match the PAAS subjective cognitive load scale. Even though participants were made aware of the entire scale as shown in Table~\ref{table_paas_level}, the tails of the range (e.g., 1 or 9) were rarely used.
% % often the cognitive load scores reported by the participants were clustered around the middle range 3 to 8. 
% % To resolve this issue, the verbally stated cognitive load scores were scaled to the scale of 1 to 9 for all participants. 
% To normalize the collected output labels and reduce the inter-participant subjectivity???, we scale each individual's scores to cover the entire range.
% Figure~\ref{fig:scaled_label} shows the verbally stated cognitive load scores of all the participants after scaling. 
% % Figure~\ref{fig:label_comparison} shows the comparison of scaled and unscaled label for one participant. 
% The data high vs. low cognitive load data from all modalities are shown in Figure \ref{fig:high_low}

\subsection{Classifiers} \label{section:classifiers}

To evaluate the dataset and to experiment the efficacy of building an automated cognitive load detection system using the collected data, we train several classical machine learning and deep learning classifiers on the extracted features or raw data. In this section, we describe these models in detail.

% We have also trained the VGG-style and ResNet-style models on raw data
% ???

% \subsubsection{Trained on features}
\subsubsection{Classical machine learning}

We train a total of 9 machine learning classifiers namely AdaBoost (AB), Decision Tree (DT), Naive Bayes (NB), K-Nearest Neighbor (KNN), Linear Discriminant Analysis (LDA), Random Forest (RF), Support Vector Machine (SVM), Extreme Gradient Boosting (XGB), and Multi-Layer Perceptron (MLP). The details of the parameters of these classifiers are presented in Table~\ref{ml_model_parameters}. 

{\renewcommand{\arraystretch}{1.8}
\begin{table}[!t]
\caption{The machine learning model parameters used in this study.}
\centering
\begin{tabular}{p{0.1\linewidth} p{0.78\linewidth}}
\hline
% 		\rowcolor{lightgray} 
		\textbf{Models} & \textbf{Parameters} \\
		\hline \hline
		AB & number of estimators: 70, learning rate: 0.1, algorithm: SAMME.R \\
	    \hline
		DT & criterion: gini, random state: 42, maximum depth: 3, minimum samples to be at leaf node: 5 \\
		\hline
		NB & variance smoothing: 1e$^{-09}$ \\
		\hline
% 		Gradient Process Classifier & n\_restarts\_optimizer: 10, warm\_start: True \\
%         \hline
% 		Gradient Boosting & learning\_rate: 0.1, n\_estimators: 100, max\_depth: 20, tol: 0.0001, ccp\_alpha: 0.2 \\
% 		\hline
		KNN & number of neighbors: 20, weights: distance, algorithm: 'auto' \\
		\hline
% 		Light Gradiant Boosting Machine & reg\_lambda: 0.01, num\_leaves: 100, n\_estimators: 1000, learning\_rate: 0.1, importance\_type: gains, boosting\_type: gbdt, class\_weight: balanced, random\_state: 24 \\
% 		\hline
		LDA & solver: least squares solution\\
		\hline
% 		Logistic Regression & C: 0.1, max\_iter: 500, class\_weight: balanced\\
% 		\hline
% 		Multinomial Naive Bayes & alpha: 1.0, fit\_prior: True\\
% 		\hline
% 		Passive Aggressive & C: 0.5, max\_iter: 500, early\_stopping: True, tol: 0.001, class\_weight: 'balanced'\\
% 		\hline
% 		Quadratic Vector Machine & reg\_param: 0.5, tol: 0.0001\\
% 		\hline
		RF & maximum depth: 50, number of estimators: 1000, number of jobs: -1, random state: 42, class weight: balanced\\
% 		\hline
% 		Ridge Classifier & alpha: 1.0, class\_weight: balanced, solver: svd\\
% 		\hline
% 		Stochastic Gradient Descent & loss: hinge, penalty: elasticnet, alpha: 0.0001, learning\_rate: optimal\\
		\hline
		SVM & regularization parameter: 0.1, kernel: polynomial\\
		\hline
		XGB & maximum depth: 20, number of estimators: 1000, learning rate: 0.001, use label encoder: False, subsample: 0.5, verbose eval: 200, booster: dart, number of jobs: -1, number of leaves: 50, regularization lambda: 0.0001, class weight: balanced\\
		\hline 
		MLP & hidden layer sizes: (100, 50), learning rate: adaptive, maximum iteration: 1000\\
		\hline
	\end{tabular}
	\label{ml_model_parameters}
\end{table}}

\subsubsection{Deep learning}
% model names + details (architectures, losses, etc), raw and features
% \textcolor{black}{Numerous innovative deep learning methodologies have emerged, accompanied by efforts aimed at addressing challenges and enhancing the effectiveness of deep learning. In \cite{liu2021activated}, a Gradient Activation Function (GAF) is proposed which can solve the vanishing and exploding gradient problem.} 
% \textcolor{black}{A Gamma Deep Belief Network (GDBN) was proposed in \cite{wu2020nonparametric} which showed highly accurate recognition capabilities and efficient learning proficiency on pilot fatigue recognition with EEG data. Deep Sum-Logarithmic-Multinomial Mixture (DSLMM) \cite{wu2021fatigue} derived from a Poisson-gamma process and Scalable Gamma Non-Negative
% Matrix Network (SGNMN) \cite{wu2021scalable} which uses Poisson randomized Gamma factor analysis are two other networks that achieved encouraging results on pilot fatigue detection. Although these new and promising models hold potential for future exploration, we have opted to use two standard networks similar to VGG and ResNet for our benchmarking work.}
For deep learning network
% trained on the extracted features, 
we use two deep Convolutional Neural Networks (CNNs), a VGG-style network as shown in Figure \ref{fig:vgg-like_network}, and a ResNet-style network which is shown in Figure \ref{fig:resnet-like_network}. The VGG-style network has two main blocks where each block consist of two Conv1D layers, batch normalization, ReLU activation, and maximum pooling operation. 
These blocks are followed by two fully connected layers and a classification layer.
% sigmoid or softmax for binary or ternary classification respectively.
% The two blocks have filter sizes of 64 and 128 respectively, and a kernel size of 3 with a stride 1. The output from the second block is fed to two fully connected layers of 128 units.
% and 1 or 3 unit respectively. 
Cross-entropy loss with a learning rate of 0.001 was used for training. ADAM was used as the optimizer for this network \cite{adam}.
% We have used binary cross-entropy for binary and cross-entropy loss for ternary classification. 

The ResNet-style network, similar to the VGG model above, consists of two blocks containing two Conv1D layers, batch normalization, and ReLU activation function in each block. The classification block contains three fully connected layer followed by an output classification layer.
% which can either be sigmoid or softmax depending on the classification type.
% The resnet blocks have base filter of size 32, kernel 3 and and stride 2. The output from the last resnet block is fed to 3 fully connected layers of unit 256, 128 and 1 respectively. 
Similar to the VGG network, we use cross-entropy loss and train the model with ADAM optimizer. Here, a learning rate of 0.01 is used. \textcolor{black}{For both networks trained with features, a batch size of 32 is used.}

It should be noted that most studies on BCI and EEG in particular use extracted features to train deep learning models \cite{zheng2018emotionmeter, zhang2022parse}, which is the approach we took with the networks described above. However, for completeness, we also train the deep networks with raw data (after pre-processing). 
% Apart from training on extracted features, we have also trained the VGG-style and ResNet-style models on raw data. 
For this purpose we design a separate encoder for each modality and use feature-level fusion.
% which learned the features from four individual modalities. 
We expectedly notice that the optimum network depth used when utilizing the extracted features (2 blocks) is not sufficient when using the raw data. We therefore increase the depth by adding a third block to obtain better results.
Accordingly each encoder for the raw data contains 3 blocks for both the VGG and ResNet-style models. The details of the architecture remain mostly the same. We present all the details of the deep networks in Table \ref{table:features_architecture}, for both VGG and ResNet-style models, when trained with features or raw signals. \textcolor{black}{For both networks trained with raw data, a batch size of 256 is used.}
% Similar to the models for deep learning with features, each VGG block contains two Conv1D layer, batch normalization and ReLU activation function and maxpool. 
% The filter sizes of the 3 VGG blocks are 64, 128 and 256 respectively. Correspondingly, the kernel sizes are 32, 16 and 8 for the 3 blocks. The outputs from each encoder are concatenated at a feature level and fed into 2 fully connected layers of units 128 and 1 or 3 depending on binary or ternary training scheme.   

% Identically, each ResNet block in the individual encoders for the ResNet-style model contains 3 ResNet blocks. The ResNet blocks have filter sizes 64, 128 and 256 and kernel sizes of 32, 16 and 8 respectively. Like the VGG-style model, each encoder's outputs are concatenated at a feature level. We have used 3 fully connected layers for the ResNet-style with 256, 128 and 1 or 3 units depending on binary or ternary evaluation criteria. For both VGG-style and ResNet-style models, we have used sigmoid or softmax function for binary or ternary classification respectively. 

\subsubsection{\textcolor{black}{Multimodal}}
% For multimodal training with raw data, the model performs feature-level fusion by extracting features from the modalities independently. After extracting features from the required modalities, global average pooling is computed of each feature map along the time axis (1D) to obtain fixed-sized feature representations. Next, the feature representations from the modalities, are concatenated along the feature dimension to create a fused feature representation. This concatenated feature vector combines information from all the modalities.

\textcolor{black}{For the multimodal setup using the classical machine learning methods, we simply concatenate the hand-crafted features, and feed the concatenated features to each classical classifier. For multimodal learning with deep neural networks (VGG and ResNet), we first feed the raw data from each modality to a separate network and 
% . We then obtain the features of each modality from the networks and 
apply MaxPool followed by global average pooling on the outcome.
% for VGG-style network and only MaxPool for ResNet-style network along the time axis to obtain fixed sized features.
The features are then fused together through concatenation and fed to the classifier block.}

% \textcolor{black}{For training the model with multiple data modalities in their raw form, feature-level fusion is performed by independently extracting features from each modality. Subsequently, MaxPool is applied for VGG-style network and both MaxPool and global average pooling is applied for ResNet-style network to each feature map along the 1D time axis to obtain fixed size features. These features from all modalities are then concatenated along the feature dimension, resulting in a fused representation that integrates information from all the modalities. In multimodal training using machine learning classifiers, we concatenate the modality's features and input them into the model. Similarly, for multimodal training with features for deep learning models, we adopt a similar approach by concatenating the modalitie's features and feeding them to the model.
% }

\subsection{Training scheme}
% 10-fold 
% LOSO
% binary
% ternary

We train all the models (classical machine learning and deep networks) in both 10-fold cross validation and the more rigorous Leave-One-Subject Out (LOSO) scheme.
% We have followed 10-fold and Leave-One-Subject-Out (LOSO) cross validation evaluation scheme during our training. 
\textcolor{black}{We also explore both binary and ternary classification of cognitive load. Certain individuals 
% might find it challenging to differentiate between cognitive load scores at a precise level such as between 3 and 4. This is the reason we converted the scores into two categories, namely `high' and `low', as well as three categories, namely `high', `medium', and `low'.
may not be able to distinguish cognitive load scores to that level of detail. This is precisely why we converted the scores to `binary' (high/low) and `ternary' (high/medium/low) levels using grouping of the scores. This initial high-resolution scheme, however, allows for future research to focus on more detailed classification schemes if necessary.} For binary, we group the cognitive load ratings from 1 to 4 as `low' cognitive load and 5 to 9 as `high' cognitive load. For ternary, we divide the cognitive load ratings into 3 groups, 1 to 3, 4 to 6, and 7 to 9 which corresponds to `low', `medium', and `high' cognitive load classes respectively.

\begin{table*}
% \footnotesize
\centering
% \tiny
\caption{Architectural details of the VGG-style and ResNet-style networks used in this study for both features and raw signals.}
\label{table:features_architecture}

\begin{tabular}{l|l|l|l|l|l}
 \hline
 \textbf{Modules} & \textbf{Parameters} & \textbf{VGG (feat.)} & \textbf{ResNet (feat.)} & \textbf{VGG (raw)} & \textbf{ResNet (raw)} \\ %[0.5ex] 
 \hline\hline
 Conv1D & Kernel size & - & 1$\times$3  & - & 1$\times$32\\
 & Filter size & - & 32 & - & 64 \\
%  & Padding type & - & Zero-padding  & - & Zero-padding \\
%  & Output feature size & - &  & - &  \\
 \hline
 Conv Block 1 & Architecture  & VGG & ResNet & VGG & ResNet \\
%  & Input feature size & 1$\times$feature size & &  &  \\
 & Activation & ReLU & ReLU & ReLU & ReLU \\
 & Kernel size & 1$\times$3 & 1$\times$3 & 1$\times$32 & 1$\times$32 \\
 & Filter size & 64 & 32 & 64 & 64 \\
 & Dropout rate & - & 0.5 & - & 0.5 \\
%  & Output feature size & & & &  \\
 \hline
 Conv Block 2 & Architecture & VGG & ResNet & VGG & ResNet \\
%  & Input feature size & & & &  \\
 & Activation & ReLU & ReLU & ReLU & ReLU \\
 & Kernel size & 1$\times$3 & 1$\times$3 & 1$\times$16 & 1$\times$16\\
 & Filter size & 128 & 32 & 128 & 128\\
 & Dropout rate & - & 0.5 & - & 0.5 \\
%  & Output feature size & & & &  \\
 \hline
 Conv Block 3 & Architecture  & - & - & VGG & ResNet \\
%  & Input feature size & - & - & &  \\
 & Activation & - & - & ReLU & ReLU \\
 & Kernel size & - & - & 1$\times$8 & 1$\times$8 \\
 & Filter size & - & - & 256 & 256 \\
 & Dropout rate & - & - & - & 0.5 \\
%  & Output feature size & - & - & &  \\
 \hline
%  Conv Block 4 & Architecture  & - & - & VGG & ResNet \\
%  & Input feature size & - & - & &  \\
%  & Activation & - & - & ReLU & ReLU \\
%  & Dropout rate & - & - & 0.5 & 0.5 \\
%  & Output feature size & - & - & &  \\
%  \hline
 Classification Block & Layer type & FC & FC & FC & FC \\
 & Number of layers & 2 & 3 & 2 & 3 \\
 & Dropout rate & 0.25 & - & 0.25 & 0.25 \\
 & Activation & ReLU & ReLU & ReLU & ReLU \\
%  & Metric & Angular error ($^{\circ}$) \\
 \hline
\end{tabular}
\end{table*}

% {\renewcommand{\arraystretch}{1.1}
\begin{table*}
% \caption{Classifiers Accuracy for binary classes for k-fold, the numbers are in accuracy (f1 score) format: AB, DT, NB, KNN, LDA, 
% RF, SVM, XGB, MLP, VGG (features), ResNet (features), VGG (raw data), ResNet (raw data)}
\caption{The accuracy and F1 scores for the classifiers in 10-fold binary setup.}
% SGD (Stochastic Gradient Descent), LR (Logistic Regression), MNB (Multinomial Naive Bayes), PA (Passive Aggressive), QDA (Quadratic Discriminant Analysis)
\label{tbl: binary_k-fold_results}
\centering
\scriptsize
\begin{tabular}{l | l | l l l | l l l | l | l}
% \begin{tabular}{l | c | c c c | c c c | c | c}
% \cmidrule{2-9}
\cline{2-9}
\multicolumn {10}{c}{Modalities} \\
% \cmidrule{2-9}
\cline{2-9}
% \hline
 % \textbf{Models} & EEG & EEG,ECG & EEG,EDA & EEG,Gaze & EEG,ECG,EDA & EEG,ECG,Gaze & EEG,EDA,Gaze & EEG,ECG,EDA,Gaze & Mean\\
\multirow{ 2}{*}{\textbf{Models}} & \multirow{ 2}{*}{EEG} & \multirow{ 2}{*}{EEG, ECG} & \multirow{ 2}{*}{EEG, EDA} & \multirow{ 2}{*}{EEG, Gaze} & EEG, ECG & EEG, ECG & EEG, EDA & EEG, ECG & \multirow{ 2}{*}{Mean}\\
& & & & & EDA & Gaze & Gaze & EDA, Gaze & \\ 
\hline\hline
AB &
% 68.51	(58.29) &
% 73.15	(68.04) &
% 72.36	(67.87) &
% 70.37	(61.39) &
% 75.04	(71.25) &
% 74.46	(69.63) &
% 73.94	(69.38) &
% \textbf{75.56	(71.94)} 
67.17	(55.37) &
73.26	(68.43) &
71.09	(66.53) &
70.13	(61.29) &
73.84	(69.85) &
74.36	(69.75) &
73.36	(69.07) &
75.35	(71.80) &
72.32 (66.51)
\\
% \hline 
DT & 
% 65.69	(61.41) &
% 68.44	(63.54) &
% 68.00	(64.01) &
% 67.72	(59.66) &
% 71.23	(66.84) &
% 70.95	(68.9) &
% 68.20	(65.75) &
% \textbf{72.26	(68.52)} 
65.31	(63.04) &
72.77	(71.15) &
69.61	(67.93) &
68.17	(66.48) &
73.02	(71.41) &
73.12	(71.52) &
72.39	(70.75) &
73.98	(72.40) &
71.05	(69.34)

\\
% \hline 
NB & 
% 47.92	(45.83) &
% 50.81	(49.33) &
% 50.05	(48.61) &
% 49.09	(47.32) &
% 52.32	(51.25) &
% 51.84	(50.64) &
% 51.15	(49.96) &
% \textbf{53.08	(52.2)} 
48.68	(46.54) &
51.80	(50.48) &
51.39	(49.97) &
49.95	(48.22) &
53.73	(52.87) &
52.46	(51.35) &
52.39	(51.23) &
54.42	(53.65) &
51.85   (50.54)
\\
% \hline 
KNN &  
% 71.47	(69.47) &
% 69.95	(67.96) &
% 72.6	(70.97) &
% 77.24	(75.04) &
% 71.60	(70.03) &
% 75.52	(72.89) &
% \textbf{77.52	(75.38)} &
% 76.28	(74.04) 
70.61	(68.49) &
69.34	(67.40) &
71.78	(70.16) &
77.00	(74.65) &
70.64	(69.16) &
74.60	(71.97) &
77.72	(75.46) &
74.97	(72.50) &
73.33	(71.22)
\\
% \hline 
LDA & 
% 68.1	(64.52) &
% 72.67	(70.62) &
% 71.95	(69.53) &
% 69.89	(66.88) &
% 75.42	(73.75) &
% 73.53	(71.53) &
% 73.91	(71.83) &
% \textbf{76.52	(75.00)} 
66.83	(62.45) &
72.74	(70.65) &
71.19	(68.45) &
69.65	(66.24) &
74.73	(72.94) &
73.70	(71.61) &
72.95	(70.62) &
75.83	(74.04) &
72.20	(69.63)
\\
% \hline 
RF  & 
% 76.97	(72.84) &
% 79.41	(76.2) &
% 78.96	(75.83) &
% 79.37	(75.68) &
% 80.72	(78.04) &
% 80.92	(78.05) &
% 80.61	(77.75) &
% \textbf{81.78	(79.29)} 
\textbf{77.41	(73.39)} &
\underline{79.34	(76.27)} &
\underline{79.48	(76.47)} &
\underline{79.89	(76.31)} &
\underline{81.26	(78.8)} &
\underline{80.82	(77.94)} &
\underline{80.65	(77.83)} &
\underline{81.71	(79.23)} &
80.07	(77.03)
\\
% \hline 
SVM & 
% 61.95	(38.48) &
% 62.19	(39.42) &
% 62.02	(38.67) &
% 64.59	(46.94) &
% 63.87	(45.33) &
% 72.05	(66.11) &
% 68.03	(56.52) &
% \textbf{73.74	(68.96)} 
61.88	(38.29) &
62.08	(38.89) &
61.88	(38.29) &
63.46	(43.86) &
64.35	(46.59) &
71.54	(65.23) &
67.14	(54.32) &
73.70	(68.98) &
65.75	(49.31)
\\
% \hline 
XGB & 
% 76.79	(72.93) &
% 83.22	(81.5) &
% 79.68	(77.24) &
% 80.65	(77.88) &
% 82.61	(80.96) &
% 83.19	(81.38) &
% 81.85	(79.74) &
% \textbf{83.57	(81.93)} 
\underline{77.38	(73.72)} &
\textbf{82.95	(81.25)} &
\textbf{80.06	(77.67)} &
\textbf{80.75	(78.06)} &
\textbf{82.61	(80.94)} &
\textbf{83.02	(81.22)} &
\textbf{82.12	(80.08)} &
\textbf{83.67	(82.05)} &
81.57	(79.37)
\\
% \hline 
MLP & 
% 73.6	(71.63) &
% 75.32	(73.83) &
% 75.90	(73.75) &
% 77.45	(75.62) &
% 77.24	(75.55) &
% 77.59	(75.94) &
% \textbf{78.48	(77.14)} &
% 77.76	(76.49) 
74.32	(72.36) &
74.22	(72.31) &
76.31	(74.02) &
75.18	(73.46) &
76.00	(74.54) &
75.83	(74.55) &
77.11	(75.47) &
77.69	(76.19) &
75.83	(74.11)
\\
% \hline 
VGG (feat.) &  
% 74.90	(72.86) &
% 77.43	(75.73) &
% 78.09	(75.92) &
% 79.27	(77.16) &
% 79.10	(77.11) &
% 79.51	(77.64) &
% \textbf{80.49	(78.82)} &
% 80.24	(78.44) 
% 75.38	(69.36) &
% 77.99	(70.85) &
% 78.82	(73.56) &
% 79.69	(73.88) &
% 79.79	(72.93) &
% 79.72	(74.1) &
% 80.52	(73.64) &
% 80.66	(75.27) 
75.56	(73.21) &
77.57	(75.8) &
78.99	(76.94) &
78.78	(76.74) &
78.78	(77.22) &
78.82	(77.23) &
80.17	(78.39) &
80.66	(79.17) &
78.67	(68.31)
\\
% \hline 
ResNet (feat.) &
% 68.65	(64.44) &
% 74.03	(71.3) &
% 72.50	(68.94) &
% 72.74	(68.88) &
% 75.00	(72.3) &
% 75.07	(72.38) &
% 75.31	(72.1) &
% \textbf{75.73	(72.83)} 
% 70.10	(59.28) &
% 73.89	(65.4) &
% 72.33	(63.3) &
% 73.78	(64.8) &
% 74.83	(66.98) &
% 74.65	(67.03) &
% 74.72	(65.75) &
% 75.28	(68.98) 
69.38	(65.26) &
74.27	(71.48) &
71.74	(68.46) &
72.85	(69.15) &
75.49	(72.71) &
74.65	(71.83) &
73.61	(70.67) &
76.39	(74.28) &
73.55	(70.48)
\\
% \hline 
VGG (raw) &
63.83	(63.23) &
67.73	(66.97) &
66.95	(66.11) &
67.62	(66.95) &
70.12	(69.2) &
71.45	(70.5) &
71.76	(71.07) &
73.87	(73.00) &
69.17	(68.38)
\\
% \hline 
ResNet (raw) &
61.95	(59.75) &
64.49	(62.14) &
60.90	(57.45) &
66.68	(64.85) &
64.41	(62.82) &
70.04	(67.69) &
68.71	(66.37) &
69.96	(67.04) &
65.89	(63.51)
\\ \hline
Mean &
67.72	(62.70) &
70.97	(61.92) &
70.11	(66.03) &
70.78	(66.64) &
72.23	(58.91) &
73.42	(70.95) &
73.08	(70.10) &
74.78	(72.64)
\\ 
% \hline
\cline{1-9}
\end{tabular}
\end{table*}
% }

% {\renewcommand{\arraystretch}{1.1}
\begin{table*}
% \caption{Classifiers Accuracy for binary classes for LOSO, the numbers are in accuracy (f1 score) format: AB (AdaBoost), DT (Decision Tree), GNB (Gaussian Naive Bayes), GP (Gaussian Process), GB (Gradient Boosting), KNN (K-Nearest Neighbor), LGBM (Light Gradient Boosting Machine), LDA (Linear Discriminant Analysis), 
% RF (Random Forest), SVM (Support Vector Machine), XGB (Extreme Gradient Boosting), MLP (Multi-Layer Perceptron)}
\caption{The accuracy and F1 scores for the classifiers in LOSO binary setup.}
% SGD (Stochastic Gradient Descent), LR (Logistic Regression), MNB (Multinomial Naive Bayes), PA (Passive Aggressive), QDA (Quadratic Discriminant Analysis)
\label{tbl: binary_loso_results}
\centering
\scriptsize
\begin{tabular}{l | l | l l l | l l l | l | l}
\cline{2-9}
\multicolumn{10}{c}{Modalities} \\
% \cmidrule{2-9}
\cline{2-9}
\hline
 % \textbf{Models} & EEG & EEG,ECG & EEG,EDA & EEG,Gaze & EEG,ECG,EDA & EEG,ECG,Gaze & EEG,EDA,Gaze & EEG,ECG,EDA,Gaze & Mean\\
 \multirow{ 2}{*}{\textbf{Models}} & \multirow{ 2}{*}{EEG} & \multirow{ 2}{*}{EEG, ECG} & \multirow{ 2}{*}{EEG, EDA} & \multirow{ 2}{*}{EEG, Gaze} & EEG, ECG & EEG, ECG & EEG, EDA & EEG, ECG & \multirow{ 2}{*}{Mean}\\
& & & & & EDA & Gaze & Gaze & EDA, Gaze & \\ 
\hline\hline
AB &
% 63.15	(47.62) &
% 66.33	(59.07) &
% 65.36	(56.7) &
% 65.00	(53.97) &
% 68.34	(62.59) &
% 66.53	(59.72) &
% 67.68	(60.29) &
% \textbf{69.21	(63.51)} 
62.30	(46.81) &
66.58	(59.47) &
63.01	(54.57) &
64.81	(53.53) &
67.86	(62.22) &
66.80	(60.05) &
66.92	(59.68) &
69.14	(63.60) &
65.93 (57.49)
\\
% \hline 
DT & 
% 62.44	(54.87) &
% 64.41	(57.50) &
% 65.52	(59.11) &
% 64.14	(55.22) &
% 66.01	(59.00) &
% \textbf{69.47	(65.17)} &
% 64.18	(59.05) &
% 67.11	(60.00) 
54.63	(49.73) &
60.33	(54.82) &
57.94	(53.07) &
57.57	(53.77) &
60.97	(56.91) &
62.13	(57.25) &
61.19	(56.73) &
62.00	(57.02) &
59.60 (54.91)
\\
% \hline 
NB & 
% 47.07	(42.85) &
% 48.49	(45.32) &
% 48.04	(44.49) &
% 47.84	(43.82) &
% 49.04	(46.46) &
% 49.30	(46.47) &
% 48.59	(45.03) &
% \textbf{49.71	(47.22)} 
47.80	(43.54) &
48.94	(45.80) &
48.16	(44.71) &
49.00	(45.06) &
49.85	(47.52) &
49.85	(47.11) &
49.15	(45.71) &
50.35	(48.06) &
49.14 (45.94)
\\
% \hline 
KNN & 
% 58.37	(53.28) &
% 60.89	(57.08) &
% 60.76	(55.76) &
% 63.75	(58.82) &
% 61.94	(58.25) &
% 64.48	(59.86) &
% \textbf{64.90	(59.86)} &
% 64.47	(60.04) 
58.21	(53.11) &
61.45	(58.09) &
60.83	(56.10) &
66.03	(61.42) &
62.51	(59.51) &
65.55	(61.10) &
67.06	(62.48) &
65.60	(61.36) &
63.40	(59.15)

\\
% \hline 
LDA & 
% 57.87	(51.74) &
% 58.92	(55.14) &
% 62.28	(56.87) &
% 60.08	(54.74) &
% 62.32	(58.24) &
% 60.27	(56.42) &
% 63.77	(58.61) &
% \textbf{63.80	(59.53)} 
57.06	(49.61) &
59.87	(55.67) &
62.95	(56.99) &
60.45	(54.25) &
63.15	(58.60) &
61.88	(57.75) &
64.45	(58.97) &
64.73	(60.26) &
61.82	(56.51)

\\
% \hline 
RF  & 
% 63.83	(51.01) &
% 65.51	(56.22) &
% 66.60	(56.67) &
% 65.68	(54.46) &
% 68.63	(60.38) &
% 66.54	(57.72) &
% 69.03	(60.05) &
% \textbf{70.14	(62.05)} 
63.82	(50.97) &
65.76	(56.84) &
66.64	(56.73) &
65.79	(54.20) &
67.95	(59.92) &
66.50	(57.90) &
68.94	(60.48) &
70.15	(62.39) &
66.94   (57.43)

\\
% \hline 
SVM & 
% 62.04	(37.66) &
% 59.86	(38.32) &
% 61.73	(37.96) &
% 63.02	(44.60) &
% 61.72	(46.53) &
% 66.16	(59.22) &
% 65.82	(52.47) &
% \textbf{68.38	(62.43)} 
62.01	(37.65) &
59.85	(37.84) &
61.80	(37.91) &
62.56	(42.96) &
61.48	(45.71) &
66.26	(59.40) &
65.94	(51.89) &
68.53	(62.79) &
63.55	(47.02)
\\
% \hline 
XGB & 
% 64.57	(54.52) &
% 66.52	(60.62) &
% 66.84	(60.27) &
% 67.24	(60.70) &
% 69.42	(64.07) &
% 68.60	(62.90) &
% 69.70	(63.98) &
% \textbf{71.30	(65.87)} 
62.98	(52.34) &
66.61	(60.53) &
66.39	(59.34) &
66.67	(59.96) &
69.37	(64.01) &
68.81	(63.20) &
69.73	(64.01) &
71.48	(66.07) &
67.76	(61.18)
\\
% \hline 
MLP & 
% 61.41	(56.82) &
% 60.72	(55.90) &
% 63.52	(57.61) &
% 62.25	(58.21) &
% 62.64	(58.76) &
% 61.45	(57.44) &
% \textbf{63.85	(59.04)} &
% 63.33	(59.34) 
57.86	(51.98) &
63.64	(57.90) &
63.44	(58.13) &
61.45	(57.43) &
64.48	(60.33) &
61.32	(57.15) &
65.72	(60.94) &
64.51	(59.41) &
62.80	(57.91)
\\
% \hline 
VGG (feat.) & 
% 71.96	(52.59) &
% 75.21	(56.87) &
% 73.93	(54.47) &
% 74.11	(54.86) &
% 75.77	(57.5) &
% 75.22	(54.68) &
% 75.56	(55.7) &
% 76.82	(57.29) 
% 70.12	(64.51) &
% 71.67	(67.18) &
% 73.37	(68.57) &
% 73.44	(69.39) &
% 74.53	(70.13) &
% 74.54	(70.43) &
% 74.83	(70.83) &
% \textbf{75.65	(71.50)} 
\textbf{70.70	(64.22)} &
\underline{74.72	(70.68)} &
\textbf{73.01	(68.08)} &
\textbf{74.68	(69.91)} &
\textbf{76.17	(71.72)} &
\textbf{76.04	(71.83)} &
\textbf{75.49	(71.34)} &
\textbf{75.52	(72.44)} &
74.54	(70.03)
\\
% \hline 
ResNet (feat.) & 
% 67.38	(61.42) &
% 74.65	(70.09) &
% 71.22	(65.6) &
% 73.28	(67.81) &
% 74.41	(69.79) &
% 74.71	(70.65) &
% 73.73	(69.04) &
% \textbf{73.95	(69.71)} 
% 69.2	(48.51) &
% 76.45	(58.67) &
% 71.87	(52.64) &
% 73.51	(56.06) &
% 75.21	(54.75) &
% 76.23	(59.77) &
% 75.26	(57.36) &
% 76.21	(57.22) 
\underline{67.45	(61.39)} &
\textbf{75.90	(71.62)} &
\underline{72.20	(66.48)} &
\underline{72.85	(67.53)} &
\underline{74.23	(69.28)} &
\underline{74.42	(70.32)} &
\underline{74.89	(69.49)} &
\underline{74.59	(69.55)} &
73.32	(68.21)
\\
% \hline 
VGG (raw) &
65.00	(58.92) &
63.67	(57.59) &
67.18	(60.78) &
65.37	(59.81) &
67.37	(60.84) &
64.97	(58.51) &
66.47	(61.79) &
67.18	(61.37) &
65.90	(59.95)
\\
% \hline 
ResNet (raw) &
65.79	(57.58) &
63.99	(56.73) &
69.03	(61.67) &
68.03	(61.12) &
68.74	(61.88) &
66.67	(59.63) &
66.50	(62.16) &
67.69	(61.88) &
67.06	(60.33)
\\
\hline
Mean &
61.20	(52.14) &
63.95	(57.20) &
64.04	(56.50) &
64.25	(57.00) &
65.70	(59.88) &
65.48	(60.09) &
66.34	(60.44) &
67.04	(62.02)

\\ 
% \hline
\cline{1-9}
\end{tabular}
\end{table*}
% }

% {\renewcommand{\arraystretch}{1.1}
\begin{table*}
% \caption{Classifiers Accuracy for ternary classes for k-fold, the numbers are in accuracy (f1 score) format: AB, DT, NB, KNN, LDA, 
% RF, SVM, XGB, MLP, VGG (features), ResNet (features), VGG (raw data), ResNet (raw data)}
\caption{The accuracy and F1 scores for the classifiers in 10-fold ternary setup.}
% SGD (Stochastic Gradient Descent), LR (Logistic Regression), MNB (Multinomial Naive Bayes), PA (Passive Aggressive), QDA (Quadratic Discriminant Analysis)
\label{tbl: ternary_k-fold_results}
\centering
\scriptsize
\begin{tabular}{l | l | l l l | l l l | l | l}
\cline{2-9}
\multicolumn {10}{c}{Modalities} \\
% \cmidrule{2-9}
\cline{2-9}
\hline
 % \textbf{Models} & EEG & EEG,ECG & EEG,EDA & EEG,Gaze & EEG,ECG,EDA & EEG,ECG,Gaze & EEG,EDA,Gaze & EEG,ECG,EDA,Gaze & Mean\\
 \multirow{ 2}{*}{\textbf{Models}} & \multirow{ 2}{*}{EEG} & \multirow{ 2}{*}{EEG, ECG} & \multirow{ 2}{*}{EEG, EDA} & \multirow{ 2}{*}{EEG, Gaze} & EEG, ECG & EEG, ECG & EEG, EDA & EEG, ECG & \multirow{ 2}{*}{Mean}\\
& & & & & EDA & Gaze & Gaze & EDA, Gaze & \\ 
\hline\hline
AB &
% 46.06	(37.96) &
% 52.29	(49.35) &
% 51.46	(48.42) &
% 49.78	(44.53) &
% 53.18	(51.29) &
% 54.38	(51.59) &
% 53.18	(50.16) &
% \textbf{55.62	(53.68)} 
46.20	(38.35) &
51.84	(48.65) &
51.56	(48.78) &
50.12	(44.48) &
53.52	(51.62) &
54.66	(51.69) &
52.56	(49.66) &
55.31	(53.15) &
51.97	(48.30)
\\
% \hline 
DT & 
% 44.66	(42.26) &
% 47.06	(43.06) &
% 46.17	(44.01) &
% 47.82	(45.02) &
% 48.40	(47.70) &
% 48.06	(42.87) &
% \textbf{48.88	(46.16)} &
% 47.99	(47.18) 
48.95	(48.77) &
56.03	(56.08) &
52.08	(51.85) &
52.70	(52.26) &
57.48	(57.36) &
58.82	(58.88) &
56.14	(55.85) &
58.13	(58.16) &
55.04	(54.90)
\\
% \hline 
NB & 
% 34.48	(29.61) &
% 36.82	(32.82) &
% 37.40	(33.89) &
% 35.51	(30.96) &
% 39.22	(36.15) &
% 37.50	(33.65) &
% 38.19	(34.89) &
% \textbf{40.19	(37.29)} 
34.93	(29.77) &
37.13	(33.00) &
37.47	(33.77) &
36.58	(31.86) &
39.50	(36.55) &
38.78	(35.06) &
38.64	(35.23) &
40.50	(37.74) &
37.94	(34.12)

\\
% \hline 
KNN &  
% 56.27	(55.61) &
% 53.39	(52.66) &
% 57.58	(57.22) &
% 63.15	(62.95) &
% 54.56	(53.88) &
% 58.44	(57.82) &
% \textbf{63.91	(63.71)} &
% 60.26	(59.7) 
54.59	(53.95) &
51.60	(50.81) &
56.34	(55.90) &
62.22	(62.01) &
52.46	(51.67) &
57.06	(56.33) &
63.05	(62.87) &
58.61	(57.93) &
56.99	(56.43)
\\
% \hline 
LDA & 
% 54.49	(53.6) &
% 59.71	(59.57) &
% 58.03	(57.81) &
% 56.45	(55.88) &
% 61.29	(61.34) &
% 60.05	(59.92) &
% 58.82	(58.61) &
% \textbf{62.80	(62.77)} 
53.01	(51.93) &
58.27	(58.09) &
58.13	(57.82) &
54.93	(54.33) &
61.02	(61.00) &
59.43	(59.23) &
58.03	(57.89) &
61.84	(61.84) &
58.08	(57.77)
\\
% \hline 
RF  & 
% 62.5	(62.0) &
% 68.65	(68.59) &
% 68.92	(68.8) &
% 67.41	(67.09) &
% 71.95	(72.07) &
% 69.99	(69.98) &
% 71.23	(71.19) &
% \textbf{72.71	(72.83)} 
\underline{63.56	(63.04)} &
\underline{68.41	(68.42)} &
\underline{69.34	(69.17)} &
\underline{68.30	(68.03)} &
\underline{71.57	(71.63)} &
\underline{70.40	(70.39)} &
\underline{71.67	(71.76)} &
\underline{72.67	(72.86)} &
69.49	(69.41)
\\
% \hline 
SVM & 
% 42.21	(24.74) &
% 46.37	(36.44) &
% 43.21	(26.94) &
% 49.78	(44.16) &
% 47.96	(39.44) &
% 53.18	(50.45) &
% 51.12	(46.61) &
% \textbf{53.87	(51.48)} 
41.01	(21.66) &
46.58	(37.78) &
41.73	(24.17) &
48.75	(42.59) &
48.40	(41.25) &
53.01	(50.34) &
50.39	(45.38) &
53.83	(51.60) &
47.96	(39.35)
\\
% \hline 
XGB & 
% 64.73	(64.39) &
% 70.54	(70.76) &
% 70.64	(70.61) &
% 70.99	(70.98) &
% 73.56	(73.8) &
% 72.74	(73.0) &
% 73.29	(73.32) &
% \textbf{74.42	(74.61)} 
\textbf{64.49	(64.14)} &
\textbf{70.78	(71.01)} &
\textbf{71.74	(71.69)} &
\textbf{71.19	(71.22)} &
\textbf{73.50	(73.76)} &
\textbf{72.91	(73.15)} &
\textbf{73.60	(73.61)} &
\textbf{74.08	(74.26)} &
71.54	(71.61)
\\
% \hline 
MLP & 
% 58.99	(58.16) &
% 59.95	(59.49) &
% 60.81	(60.65) &
% 62.98	(62.88) &
% 63.11	(63.06) &
% 61.33	(61.31) &
% 64.32	(64.32) &
% \textbf{65.21	(65.17)} 
58.44	(57.72) &
61.50	(60.64) &
61.12	(60.90) &
62.63	(62.22) &
62.80	(62.66) &
63.08	(63.09) &
63.84	(63.83) &
63.25	(63.26) &
62.08	(61.79)
\\
% \hline 
VGG (feat.) &  
% 60.59	(59.73) &
% 62.71	(62.23) &
% 63.61	(63.05) &
% 64.31	(63.84) &
% 64.13	(63.77) &
% 64.97	(64.68) &
% 65.97	(65.81) &
% \textbf{67.15	(66.57)} 
% 60.97	(55.53) &
% 62.94	(57.07) &
% 63.28	(59.38) &
% 66.06	(60.71) &
% 64.19	(59.4) &
% 64.46	(60.44) &
% 67.28	(63.02) &
% 67.07	(62.58) 
62.12	(60.92) &
62.85	(62.21) &
64.44	(63.91) &
65.38	(64.88) &
65.76	(65.37) &
64.03	(63.43) &
67.12	(66.72) &
66.67	(66.04) &
64.80	(64.19)
\\
% \hline 
ResNet (feat.) &
% 47.6	(45.22) &
% 54.17	(53.71) &
% 51.08	(49.67) &
% 51.46	(49.28) &
% 55.69	(55.34) &
% 53.26	(52.6) &
% 52.53	(51.65) &
% \textbf{55.31	(54.48)} 
% 47.33	(32.96) &
% 50.52	(37.99) &
% 48.69	(35.37) &
% 50.87	(36.68) &
% 50.67	(36.7) &
% 50.98	(38.05) &
% 51.78	(43.06) &
% 51.29	(38.79) 
47.19	(44.61) &
55.52	(53.74) &
51.91	(50.66) &
51.39	(49.41) &
56.15	(55.48) &
54.86	(53.88) &
53.19	(52.24) &
55.63	(54.67) &
53.23	(51.84)
\\
% \hline 
VGG (raw) &
47.85	(43.7) &
55.43	(51.37) &
56.48	(51.61) &
56.68	(52.24) &
61.76	(57.96) &
62.89	(58.3) &
63.44	(59.41) &
66.33	(62.93) &
58.86	(54.69)
\\
% \hline 
ResNet (raw) &
50.82	(37.41) &
56.56	(50.09) &
53.67	(44.25) &
56.37	(49.93) &
60.62	(54.07) &
59.38	(54.49) &
61.05	(56.39) &
64.84	(60.60) &
57.91	(50.90)
\\\hline 
Mean &
51.78	(47.38) &
56.35	(53.99) &
55.85	(52.65) &
56.71	(54.27) &
58.81	(56.95) &
59.18	(57.56) &
59.44	(57.76) &
60.90	(59.62) 
\\ 
% \hline
\cline{1-9}
\end{tabular}
\end{table*}
% }

% {\renewcommand{\arraystretch}{1.1}
\begin{table*}
% \caption{Classifiers Accuracy for ternary classes for LOSO, the numbers are in accuracy (f1 score) format: AB, DT, NB, KNN, LDA, 
% RF, SVM, XGB, MLP, VGG (features), ResNet (features), VGG (raw data), ResNet (raw data)}
\caption{The accuracy and F1 scores for the classifiers in LOSO ternary setup.}
% SGD (Stochastic Gradient Descent), LR (Logistic Regression), MNB (Multinomial Naive Bayes), PA (Passive Aggressive), QDA (Quadratic Discriminant Analysis)
\label{tbl: ternary_loso_results}
\centering
\scriptsize
\begin{tabular}{l | l | l l l | l l l | l | l}
\cline{2-9}
\multicolumn {10}{c}{Modalities} \\
% \cmidrule{2-9}
\cline{2-9}
\hline
%  \textbf{Models} & EEG & EEG,ECG & EEG,EDA & EEG,Gaze & EEG,ECG,EDA & EEG,ECG,Gaze & EEG,EDA,Gaze & EEG,ECG,EDA,Gaze & Mean\\
\multirow{ 2}{*}{\textbf{Models}} & \multirow{ 2}{*}{EEG} & \multirow{ 2}{*}{EEG, ECG} & \multirow{ 2}{*}{EEG, EDA} & \multirow{ 2}{*}{EEG, Gaze} & EEG, ECG & EEG, ECG & EEG, EDA & EEG, ECG & \multirow{ 2}{*}{Mean}\\
& & & & & EDA & Gaze & Gaze & EDA, Gaze & \\ 
\hline\hline
AB &
% 37.1	(26.07) &
% 42.78	(35.47) &
% 42.32	(35.27) &
% 43.21	(35.99) &
% 44.35	(38.75) &
% 45.01	(38.03) &
% 44.99	(38.21) &
% \textbf{47.68	(41.97)} 
37.79	(27.09) &
42.13	(34.39) &
42.56	(36.04) &
43.88	(36.63) &
44.26	(38.46) &
45.42	(37.95) &
46.03	(39.76) &
47.65	(41.81) &
43.15	(35.76)
\\
% \hline 
DT & 
% 34.1	(31.31) &
% 39.77	(37.52) &
% 37.7	(35.89) &
% 37.02	(34.55) &
% 37.59	(35.28) &
% \textbf{40.61	(37.7)} &
% 38.29	(36.79) &
% 40.22	(37.93) 
35.83	(33.35) &
40.15	(37.63) &
37.37	(34.68) &
37.69	(34.8) &
35.77	(33.02) &
40.32	(37.76) &
37.58	(35.42) &
39.15	(37.17) &
37.82   (35.24)
\\
% \hline 
NB & 
% 33.22	(26.24) &
% 33.55	(26.82) &
% 33.03	(27.75) &
% 33.64	(27.28) &
% 33.20	(28.27) &
% 34.26	(27.91) &
% 33.42	(28.57) &
% \textbf{33.87	(29.05)} 
33.28	(26.3) &
33.81	(27.37) &
33.02	(27.85) &
34.27	(28.04) &
33.23	(28.40) &
34.71	(28.59) &
33.93	(29.18) &
34.08	(29.40) &
33.75	(27.96)
\\
% \hline 
KNN & 
% 35.97	(33.64) &
% 39.2	(36.93) &
% 36.8	(34.23) &
% \textbf{42.91	(40.6)} &
% 39.94	(37.39) &
% 41.94	(39.63) &
% 42.17	(39.64) &
% 41.18	(39.12) 
35.19	(32.87) &
39.19	(37.06) &
37.40	(34.97) &
44.03	(41.90) &
39.24	(37.31) &
41.51	(39.82) &
44.78	(42.45) &
41.69	(40.04) &
40.19	(38.05)
\\
% \hline 
LDA & 
% 38.49	(35.39) &
% 41.17	(38.00) &
% 42.01	(38.54) &
% 40.95	(38.14) &
% \textbf{43.20	(39.60)} &
% 41.26	(38.60) &
% 41.69	(38.64) &
% 43.10	(40.02) 
36.18	(33.64) &
40.23	(37.02) &
40.33	(37.61) &
38.43	(36.29) &
42.31	(39.07) &
40.02	(37.22) &
40.96	(38.78) &
41.03	(38.45) &
39.78	(37.09)
\\
% \hline 
RF  & 
% 36.79	(32.54) &
% 39.56	(37.29) &
% 39.56	(35.92) &
% 41.93	(39.13) &
% 41.27	(39.36) &
% 42.43	(39.83) &
% 43.73	(41.13) &
% \textbf{44.36	(42.14)} 
37.02	(32.58) &
40.15	(37.54) &
39.97	(36.05) &
43.03	(39.93) &
42.28	(40.15) &
43.11	(40.13) &
45.31	(42.38) &
44.96	(42.76) &
41.55	(38.39)
\\
% \hline 
SVM & 
% 37.71	(20.67) &
% 38.76	(30.32) &
% 38.14	(21.78) &
% 41.21	(33.8) &
% 39.29	(31.56) &
% 43.24	(38.85) &
% 42.62	(35.9) &
% \textbf{44.19	(40.24)} 
38.48	(20.41) &
39.66	(31.73) &
38.42	(21.00) &
42.46	(34.15) &
39.83	(33.22) &
44.69	(40.43) &
43.13	(35.87) &
45.05	(41.22) &
40.95	(30.97)
\\
% \hline 
XGB & 
% 35.91	(32.91) &
% 39.48	(37.49) &
% 41.05	(38.04) &
% 41.94	(40.38) &
% 43.82	(41.57) &
% 43.37	(40.98) &
% 45.73	(43.75) &
% \textbf{46.88	(44.43)} 
36.09	(32.83) &
40.06	(38.11) &
41.63	(38.50) &
43.26	(41.51) &
44.48	(42.31) &
43.44	(40.82) &
46.14	(43.77) &
47.10	(44.68) &
42.16	(39.69)
\\
% \hline 
MLP & 
% 35.8	(32.87) &
% 38.4	(35.98) &
% 40.35	(37.29) &
% 43.18	(40.78) &
% 44.46	(40.8) &
% 41.16	(39.3) &
% \textbf{43.21	(40.39)} &
% 41.97	(38.18) 
38.30	(34.11) &
39.30	(36.70) &
41.44	(36.83) &
44.23	(41.10) &
42.04	(38.88) &
41.68	(38.99) &
46.05	(42.65) &
42.94	(40.38) &
41.86	(38.47)
\\
% \hline 
VGG (feat.) &  
% % % % 48.32	(43.68) &
% % 52.84	(49.39) &
% % 51.01	(47.4) &
% % 53.64	(50.24) &
% % 54.07	(50.45) &
% % \textbf{56.09	(52.37)} &
% % 55.71	(51.7) &
% % 54.86	(52.71) 
% 49.75	(32.95) &
% 53.16	(35.99) &
% 52.31	(34.05) &
% 52.98	(36.25) &
% 54.27	(35.15) &
% 54.84	(38.49) &
% 55.58	(39.68) &
% 56.86	(38.84) 
49.21	(43.75) &
54.34	(48.93) &
51.83	(47.64) &
54.67	(49.85) &
54.39	(50.96) &
53.64	(49.99) &
56.16	(52.17) &
56.88	(52.98) &
53.46	(49.04)
\\
% \hline 
ResNet (feat.) &
% 47.06	(41.22) &
% 54.15	(50.14) &
% 53.21	(47.23) &
% 51.36	(46.67) &
% 56.22	(50.73) &
% 55.09	(50.79) &
% 52.58	(47.79) &
% \textbf{57.44	(51.07)} 
% 51.28	(30.33) &
% 56.57	(34.25) &
% 54.31	(31.03) &
% 54.16	(36.06) &
% 58.91	(37.00) &
% 56.75	(37.93) &
% 55.51	(35.67) &
% 59.34	(37.86) 
47.30	(42.08) &
52.82	(49.44) &
53.44	(47.1) &
50.23	(46.15) &
55.31	(50.65) &
53.81	(49.49) &
53.26	(49.16) &
55.70	(51.02) &
52.31	(47.72) 
\\
% \hline 
VGG (raw) &
\underline{57.91	(44.12)} &
\underline{60.84	(47.61)} &
\textbf{58.86	(44.83)} &
\underline{58.08	(44.08)} &
\textbf{61.29	(49.18)} &
\underline{63.04	(49.47)} &
\underline{57.01	(45.06)} &
\textbf{63.56	(49.39)} &
59.58	(46.34)
\\
% \hline 
ResNet (raw) &
\textbf{58.13	(42.54)} &
\textbf{60.86	(47.30)} &
\underline{58.22	(43.81)} &
\textbf{60.37	(45.90)} &
\underline{60.12	(46.68)} &
\textbf{64.53	(51.40)} &
\textbf{58.65	(46.27)} &
\underline{61.55	(49.93)} &
60.13	(46.27)
\\ \hline 
Mean &
41.59	(34.28) &
44.89	(39.29) &
44.19	(37.45) &
45.74	(40.03) &
45.73	(40.64) &
46.92	(41.70) &
46.85	(41.76) &
47.80	(43.02)
\\ 
% \hline
\cline{1-9}
\end{tabular}
\end{table*}
% }

\section{Benchmarking Results}
\label{section:results}
% ???? overview explanation...

% We have trained our machine learning and deep learning models in 2 different evaluation scheme k-fold and LOSO for both binary as well as ternary label distributions. We have used features and raw data in separate trainings for both evaluation schemes and label distributions. Handcrafted features are commonly used in BCI using machine learning and deep learning \cite{zheng2018emotionmeter, zhang2022parse}. For training with raw data, we have segmented the data in 10 second segments so the input shape of the data would be batch size x channel x (sampling rate * 10). Along with machine learning models in \ref{ml_model_parameters}, we have also performed the VGG-style and ResNet-style model training on the extracted features and achieved better performance than that of raw data in most cases. The detail explanation of our results are given below in this section.

Here, we present the results of the benchmarking study for binary and ternary classification in both validation schemes. We also present a comparison for the different multimodal setups in our experiments. The detailed results are presented in Tables \ref{tbl: binary_k-fold_results}, \ref{tbl: binary_loso_results}, \ref{tbl: ternary_k-fold_results}, and \ref{tbl: ternary_loso_results}.  
In these tables, bold values denote the highest, while underline represents the second-highest. As we observe in Table \ref{tbl: binary_k-fold_results}, for 10-fold cross-validation in the binary setup, we obtain the highest accuracy of 83.67\% with the XGB classifier. This performance is achieved when all 4 modalities are used. This is followed by 83.02\% as the second best obtained with EEG with ECG and Gaze by the same classifier. Comparing the average values obtained for different modality setups indicates that as expected, EEG, ECG, EDA, and Gaze altogether outperform the rest, followed by tri-modal, bi-modal, and uni-modal setups, respectively. Looking at the average values for all the models we observe that the XGB classifier generally outperforms the rest followed by RF. Among the 4 different deep learning variants in this setup, we notice that VGG trained with features from all 4 modalities outperforms the other 3 scenarios.

% Among the 4 different deep learning variants in this setup, we notice that VGG trained with features from all 4 modalities outperforms the other 3 scenarios. Comparing the average values for different classifiers, we observe that XGB classifier generally outperforms the rest, while analyzing the average values obtained for different multimodal setups indicates that as expected, EEG, ECG, EDA, and Gaze altogether outperform the rest. When comparing the bi-modal setups, we observe that EEG along with ECG as an auxiliary modality performs better than the alternatives.

In Table \ref{tbl: binary_loso_results}, for the binary LOSO evaluation scheme, we observe that the highest accuracy of 76.17\% is obtained by the VGG-style network trained with features. This accuracy is obtained using 3 modalities, namely EEG, ECG and EDA. The second best accuracy, 76.04\%, is achieved with the combination of EEG, ECG, and Gaze. Comparing the average values from different modality setups, we observe that the highest accuracy is obtained by the multimodal scenario with all 4 modalities, followed by the tri-modal, bi-modal, and uni-modal setups respectively. From the average values for each model, we can deduce that the VGG-style model trained on features performs the best, followed by the ResNet-style model trained on features. 

% Here, we observe that the VGG-style model trained on features performed the best among the 4 deep learning models. Comparing the average performance of all the models across different multimodal setups, we see that the best performance is achieved when EEG is combined with all other 3 modalities. Moreover, in the majority of cases, the VGG-style classifier trained on features shows superior performance over the other models. Comparing the bi-modal setups, we see that Gaze seems to have the most added value to EEG, followed closely by EDA.

% The highest accuracy of 75.90\% is obtained with ResNet-style model when EEG is trained with ECG in a bi-modal setup. 

We present the results for the ternary 10-fold setup in Table \ref{tbl: ternary_k-fold_results}, and we observe that the best result of 74.08\% is achieved with the machine learning classifier, XGB. 
% The result for the ternary 10-fold setup is shown in Table \ref{tbl: ternary_k-fold_results}. 
This result is obtained when EEG is trained along with all 3 auxiliary modalities. The second best accuracy of 73.60\% is obtained with the same classifier when EEG, EDA, and Gaze are used together. Comparing the average results for each modality setup, we find that as expected, the using all 4 modalities outperforms the rest while the best performing average result is achieved by the XGB classifier. Among the 4 deep learning models, the VGG-style model outperforms the other 3 when trained with features. Here, the highest accuracy of 67.12\% is obtained when trained with 3 modalities EEG, EDA, and Gaze.

% Among the 4 deep learning models, VGG-style model outperforms the other 3 when trained with features. Here, the highest accuracy of 67.12\% is obtained when trained with 3 modalities EEG, EDA and Gaze. In the bi-modal setup, we observe that EEG along with EDA achieves the best performance when trained with XGB classifier. Analyzing the average values, we can observe that, all 4 modalities combined give us the best result and the best-performing classifier is the XGB classifier. 

Lastly, Table \ref{tbl: ternary_loso_results} shows us the result in the LOSO ternary setup. The highest result obtained is 64.53\% when using the ResNet-style network trained with the raw data with 3 modalities namely EEG, ECG, and Gaze. Following, the second highest accuracy of 63.56\% is obtained by the VGG-style network trained on raw data from all 4 modalities. The average results of various modality setups show that the multimodal setup with all 4 modalities achieves the highest accuracy. The average highest accuracy is obtained using the ResNet-style model followed by the VGG-style model both when trained with raw data.

% \textcolor{black}{Cognitive load may be influenced by individual differences. That is why we have done the Leave-one-subject-out (LOSO) evaluation scheme. This approach enables us to assess the performance of each participant individually. While we have reported the average accuracy and F1 scores derived from the LOSO setup, we observed that the k-fold performance surpasses that of LOSO. One of the reasons for this is the higher variability between the train and test set of the LOSO setup.}

In the end, to summarize our findings above, we observe that both classical machine learning and deep learning models possess the ability to distinguish between different levels of driver cognitive loads. As expected, we find that ternary classification is more challenging than binary, while LOSO on the other hand proves more difficult than 10-fold. In terms of modalities, multimodal setups generally provide more information regarding driver cognitive load, with EEG, ECG, EDA and Gaze showing the best performances. 

\subsection{\textcolor{black}{Limitations}}
\textcolor{black}{
% While 
% the driver's cognitive load dataset is valuable and substantial efforts were dedicated to minimize errors and 
While our work makes significant contributions to the area, there exist a few areas in which our work could be improved. 
% limitations, it has come to our attention that certain constraints within our study remained. 
For instance, using a driving simulator versus real vehicles offers a safe alternative, reduces the risk of accidents and injuries, and allows us to design very specific driving scenarios by controlling the weather, time of day, road conditions, obstacles, number of cars on the road, etc., that are applicable across all the participants. On the other hand, the disadvantage of using a simulator is the possibility of participants experiencing SAS, as well as limited motion and the use of generated graphics that may not fully replicate the experience of real-world driving.} 

\textcolor{black}{The number of participants (6M+15F = total of 21) in our study is in line with other datasets such as SEED (7M+8F = total of 15), SEED-IV (7M+8F = total of 15), SEED-VIG (11M+12F = total of 23), and others. However, we acknowledge that adding more participants and further balancing the demographics in terms of factors such as gender can help increase generalization of the findings and improve diversity in the data, which can lead to more effective machine learning models.}

\textcolor{black}{Lastly, another area that could be discussed is the use of participant-reported subjective measurements for cognitive load. It should be noted that while advanced brain scanning technologies could be used to measure cognitive load more objectively, such self-reported methods are consistently relied upon in the literature to obtain labels or scores with which to train machine learning models.
% such approaches are consistently used in the literature to obtain labels/scores with which to train machine learning models. 
Additionally, the fact that the trained models are capable of making strong predictions, points to the reliability of the output labels. To reduce the possible subjectivity of self-reported scores, larger datasets may be used, in addition to deep learning paradigms such as weakly supervised or partial-label learning.}

\section{Conclusion}
In this paper, we presented CL-Drive, a new multimodal cognitive load dataset collected during simulated driving. Our dataset, which we made public, contains EEG, ECG, EDA, and Gaze data from 21 participants in a variety of different driving conditions. Subjective self-reported cognitive load scores were recorded at 10-second intervals throughout the experiment, making it a very rich and dense dataset in terms of both modalities and labels. We also provided benchmarks by evaluating our dataset in both binary and ternary label distributions for both LOSO as well as k-fold evaluation schemes. 
\textcolor{black}{The CL-Drive dataset can have various applications in the field of transportation, driver safety, and human-machine interaction. The dataset can be used to assess the cognitive workload experienced by drivers in various driving scenarios, such as high-traffic conditions, adverse weather, or during complex maneuvers. 
% Furthermore, it can be used to evaluate the cognitive load demand of driving as a primary task. 
Overall, the CL-Drive dataset has the potential to improve road safety, enhance driver experience, and contribute to the development of more intelligent and human-centered transportation systems.}

% The dataset contains data from 4 modalities namely EEG, ECG, EDA and Gaze from 21 subjects for 9 different simulated driving complexity levels. 

\section{Acknowledgement}
We would like to thank the Innovation for Defence Excellence and Security (IDEaS) program under the Department of National Defence (DND) for funding this project. 

\bibliographystyle{IEEEtran}
\scriptsize
\bibliography{References}

% \newpage


\begin{IEEEbiography}[{\includegraphics[width=1in,height=1.25in,clip,keepaspectratio]{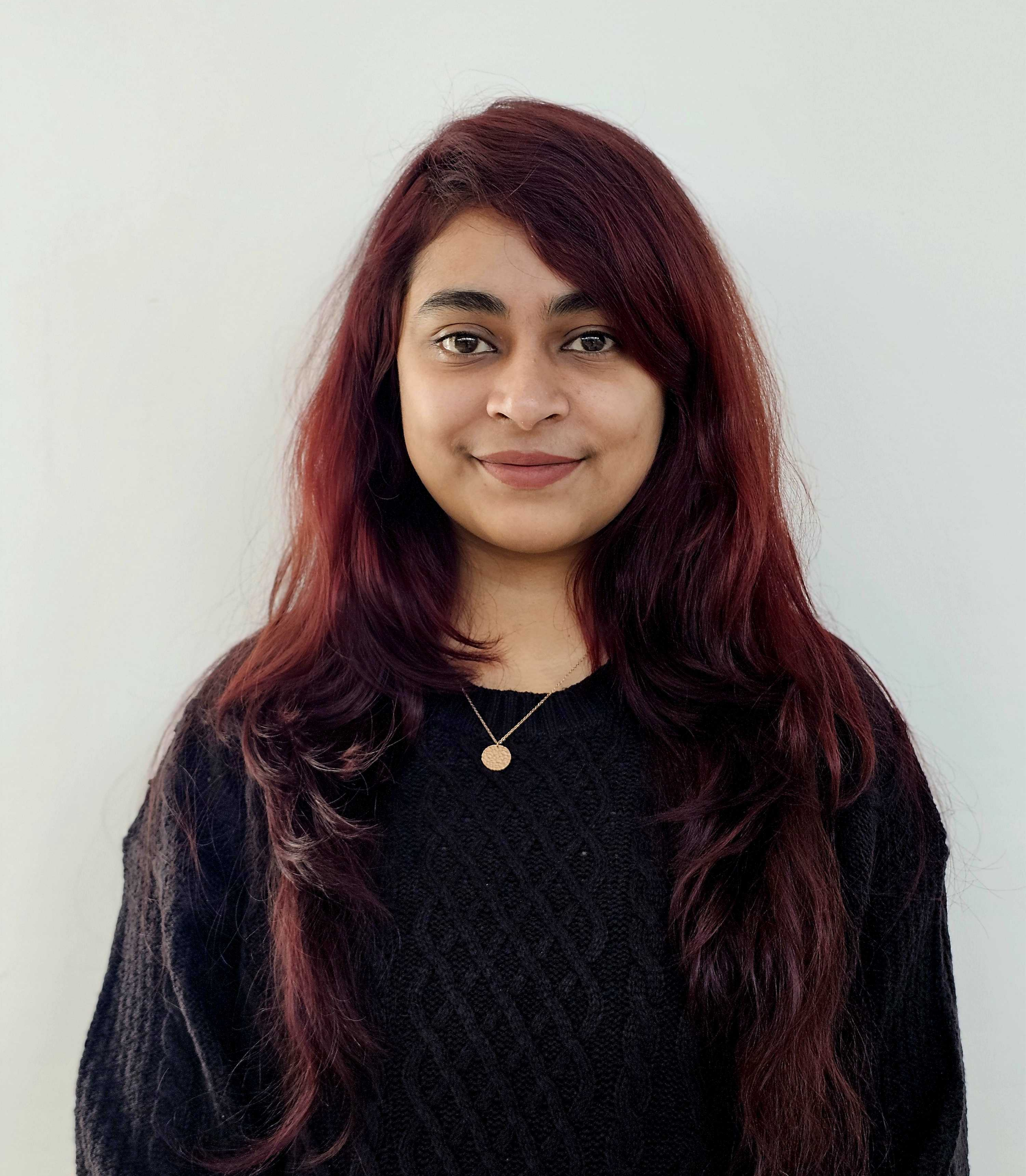}}]{Prithila Angkan}
Prithila Angkan is currently pursuing her Ph.D. at Queen's University in Canada in the Department of Electrical and Computer Engineering. She received her master's degree from the same department at Queen's University and her bachelor's from BRAC University, Bangladesh. Currently, she is a member of Ambient Intelligence and Interactive Machines Laboratory
(Aiim Lab) and Ingenuity Labs Research Institute at Queen's University. Her research focuses on Brain-Computer Interface, Artificial Intelligence, Cognitive Load Analysis, Affective Computing, and Deep Learning utilizing Electroencephalogram (EEG) signals.

\end{IEEEbiography}

\begin{IEEEbiography}[{\includegraphics[width=1in,height=1.25in,clip,keepaspectratio]{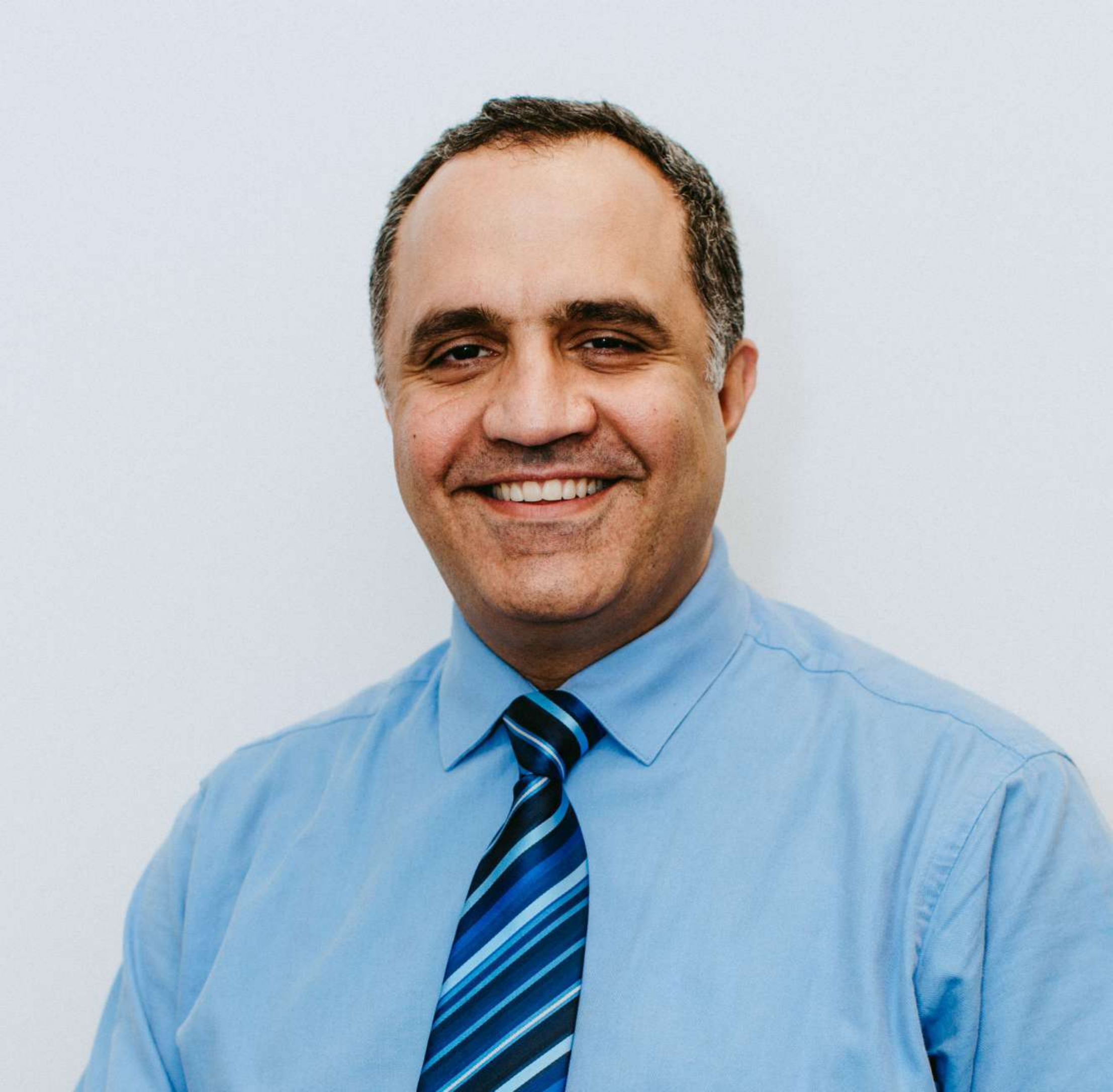}}]{Behnam Behinaein}
Behnam Behinaein received his PhD from Queen’s University in 2016 and was a post-doctoral fellow in QDES and Aiim Labs. He is currently with Huawei Canada. His research interests include affect computing, wearable technologies, and the application of deep learning in analyzing sequential data.
\end{IEEEbiography}

\begin{IEEEbiography}[{\includegraphics[width=1in,height=1.25in,clip,keepaspectratio]{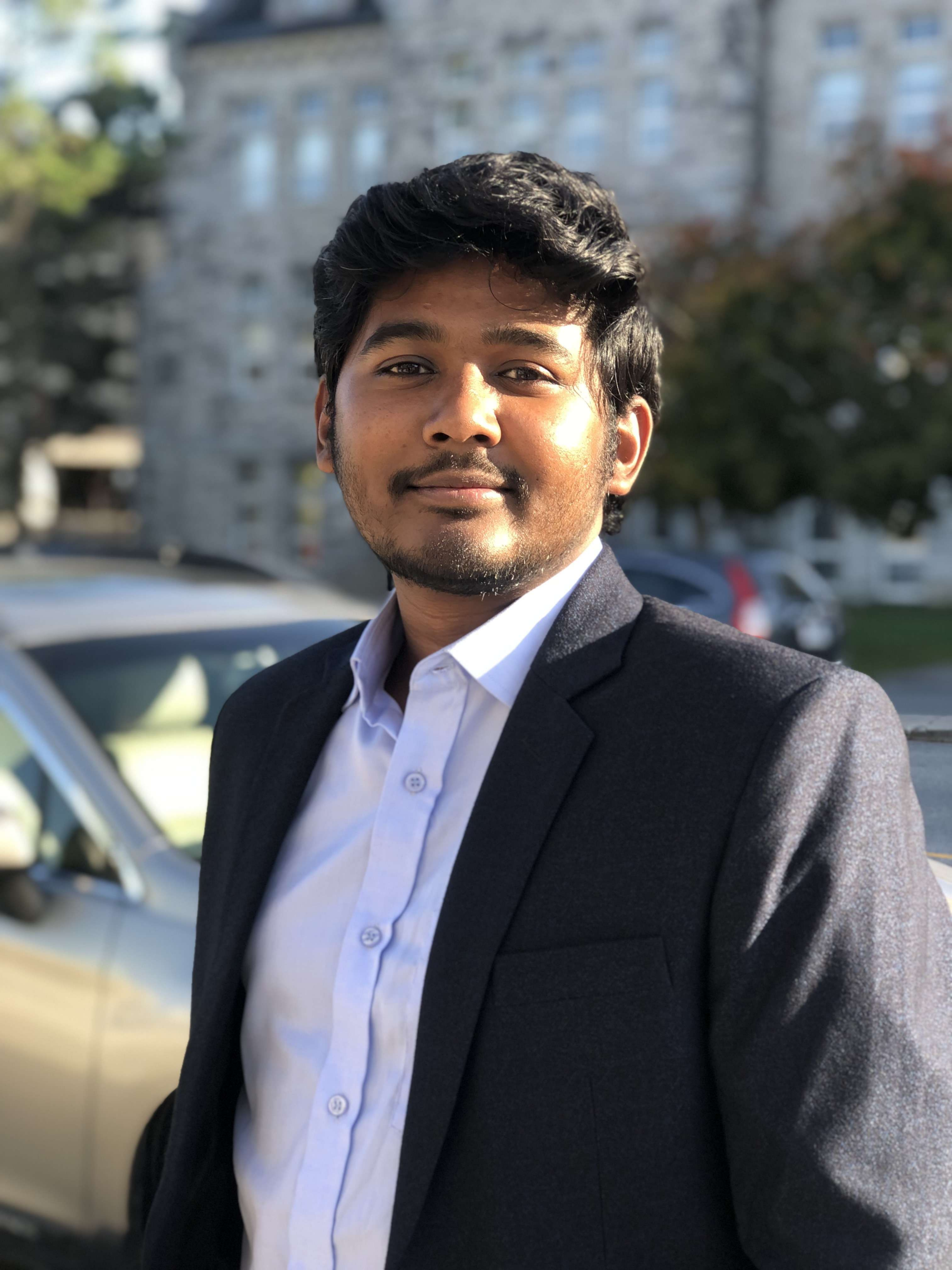}}]{Zunayed Mahmud}
Zunayed Mahmud received the M.A.Sc. degree from the Department of Electrical and Computer Engineering at Queen’s University, Canada, in 2022. Concurrently, he served as a Student Researcher at the Ingenuity Labs Research Institute, Canada. He earned his bachelor's degree from the Department of Electrical and Electronic Engineering at BRAC University, Bangladesh, in 2015. He is currently working as an Associate Researcher at Huawei Technologies Canada. His research interests include computer vision, deep learning, 2D/3D object detection and recognition, 3D reconstruction, and gaze estimation.
\end{IEEEbiography}

\begin{IEEEbiography}[{\includegraphics[width=1in,height=1.25in,clip,keepaspectratio]{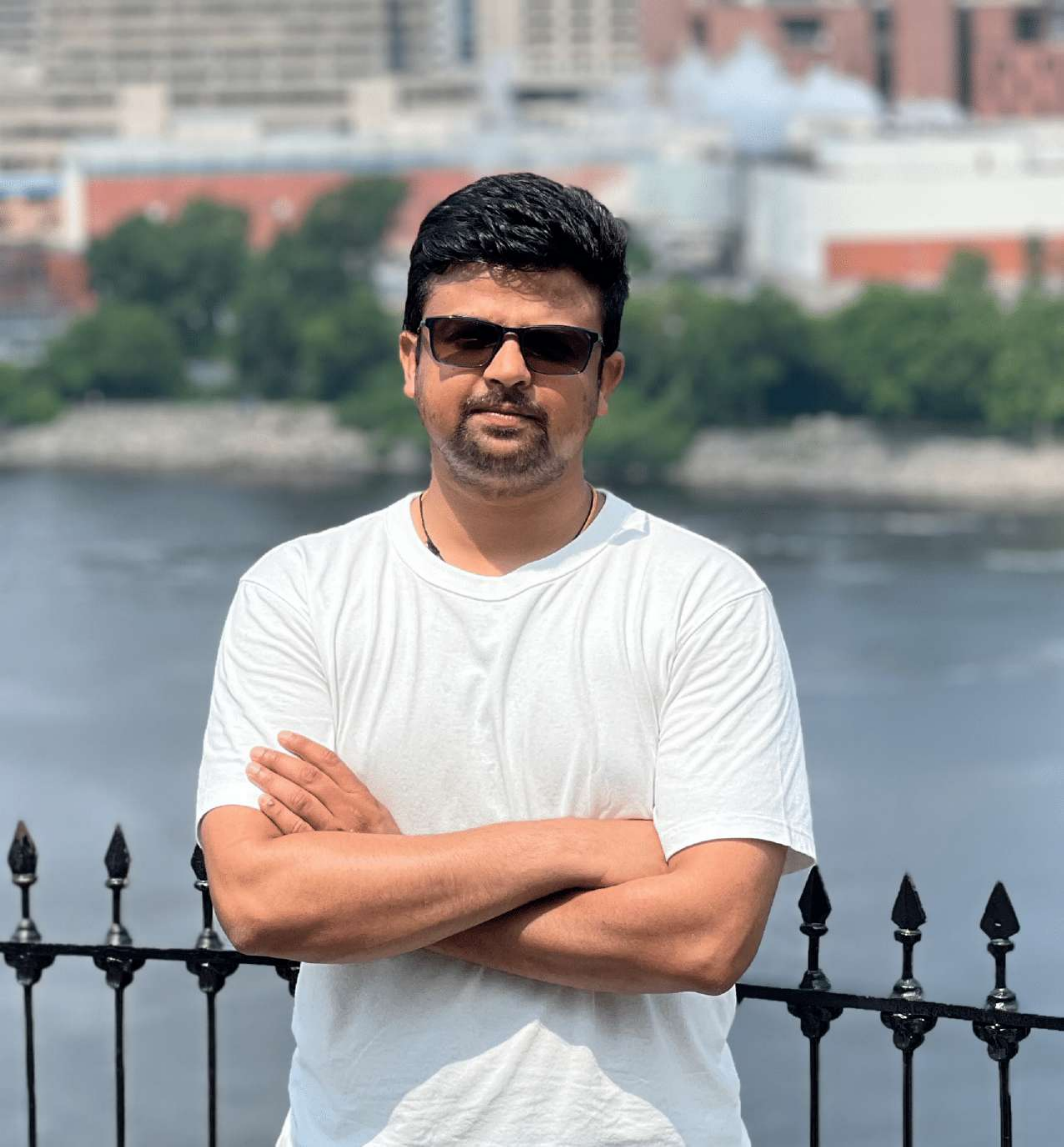}}]{Anubhav Bhatti}
Anubhav Bhatti is a machine learning engineer at SpassMed Inc. His journey in machine learning previously led him to the Vector Institute as a Machine Learning Associate. He received his Master’s in Artificial Intelligence from the Electrical and Computer Engineering Department at Queen's University, Canada, and his bachelor’s in electrical engineering from the National Institute of Technology, India. In his research work at Vector Institute and SpassMed, he leverages his expertise in Large Language Models, Generative AI, and Time Series Forecasting in designing and implementing deep-learning models for the early detection of critical events in patients. While at Queen's University and Ingenuity Labs Research Institute, he extensively researched multimodal time-series physiological data fusion techniques and affective computing.
\end{IEEEbiography}

\begin{IEEEbiography}[{\includegraphics[width=1in,height=1.25in,clip,keepaspectratio]{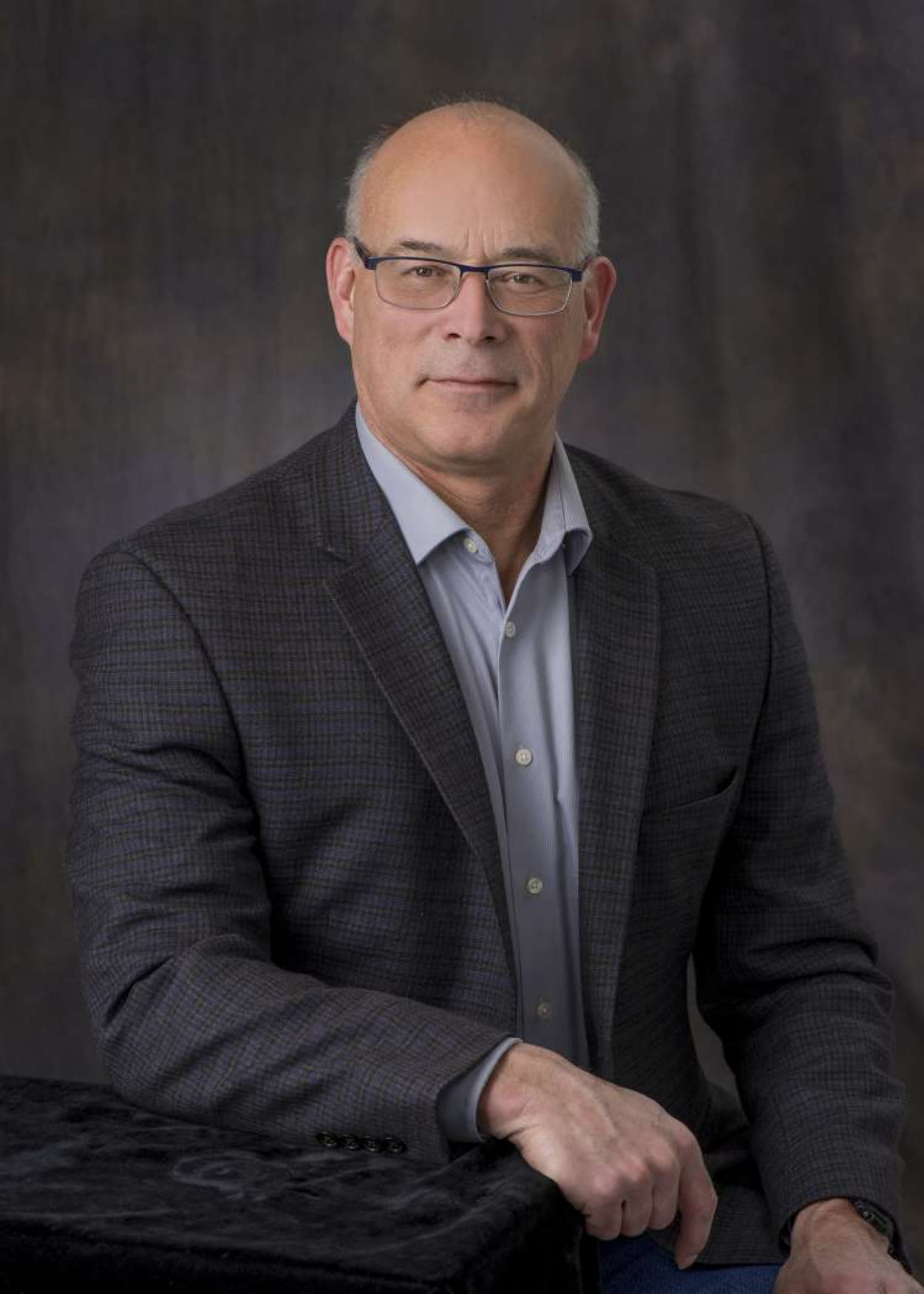}}]{Dirk Rodenburg}
Dr. Rodenburg is an Adjunct Professor with the Queen’s University Smith School of Engineering’s Department of Chemical Engineering, and the Faculty of Arts and Science. Dirk’s research interests include human performance, expertise, cognition, real time data analytics, data feedback, human-computer interaction, and ethics and privacy. In addition to his academic background, he has twenty years of experience as a software entrepreneur and consultant within academia, educational technology, financial services, biotechnology and scientific instruments, and has played an integral role in the launch of three highly innovative startups. Dirk holds an MA in Adult Education from the University of British Columbia and a PhD from the Faculty of Information from the University of Toronto.
\end{IEEEbiography}

\begin{IEEEbiography}[{\includegraphics[width=1in,height=1.25in,clip,keepaspectratio]{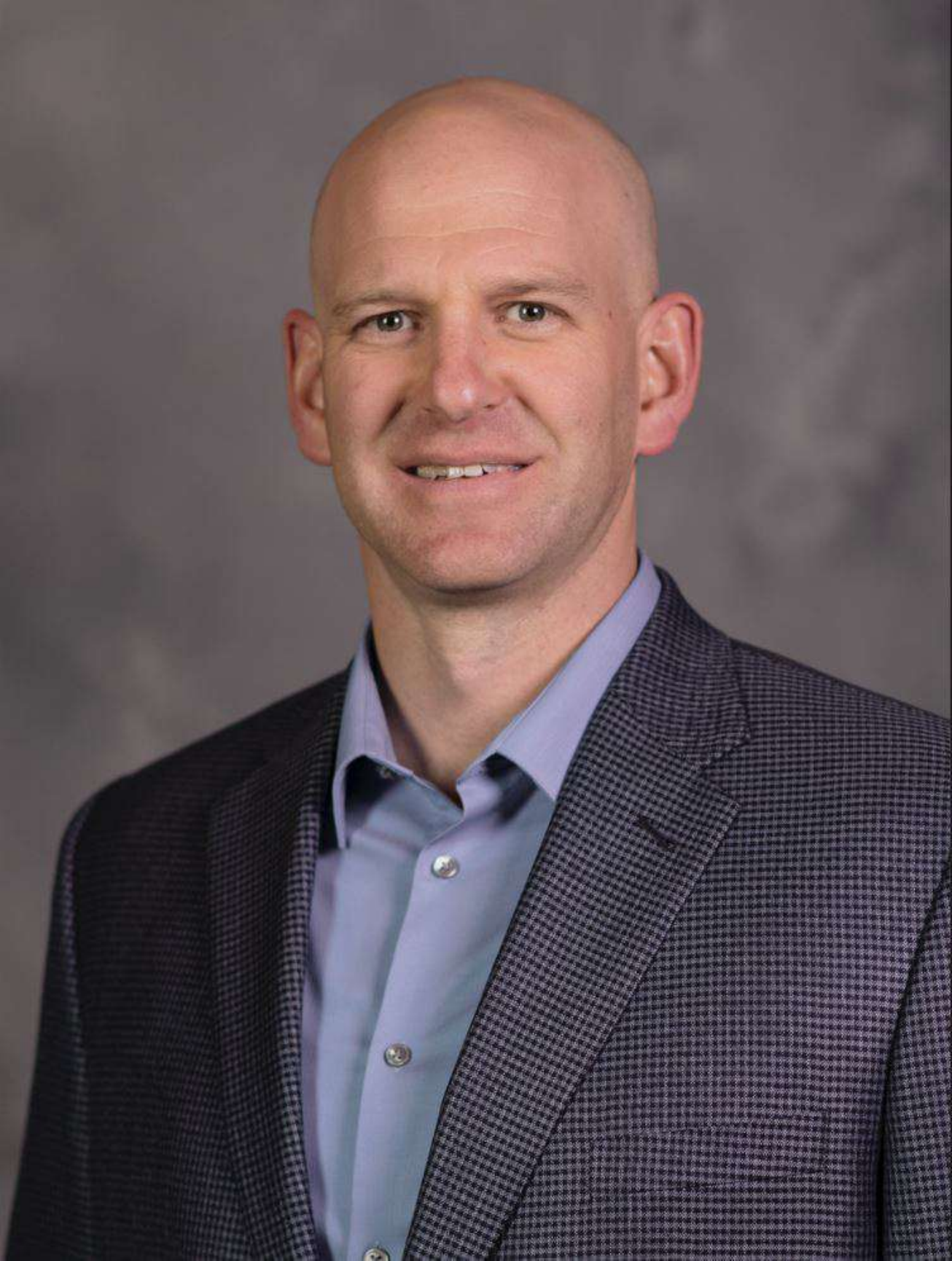}}]{Paul Hungler}
Dr. Paul Hungler is an Associate Professor in the Department of Chemical Engineering and Ingenuity Labs at Smith Engineering, Queen’s University. Dr. Hungler’s research is focused on the development of intelligent, dynamically adaptive simulation to enhance education and training.
\end{IEEEbiography}

\begin{IEEEbiography}[{\includegraphics[width=1in,height=1.25in,clip,keepaspectratio]{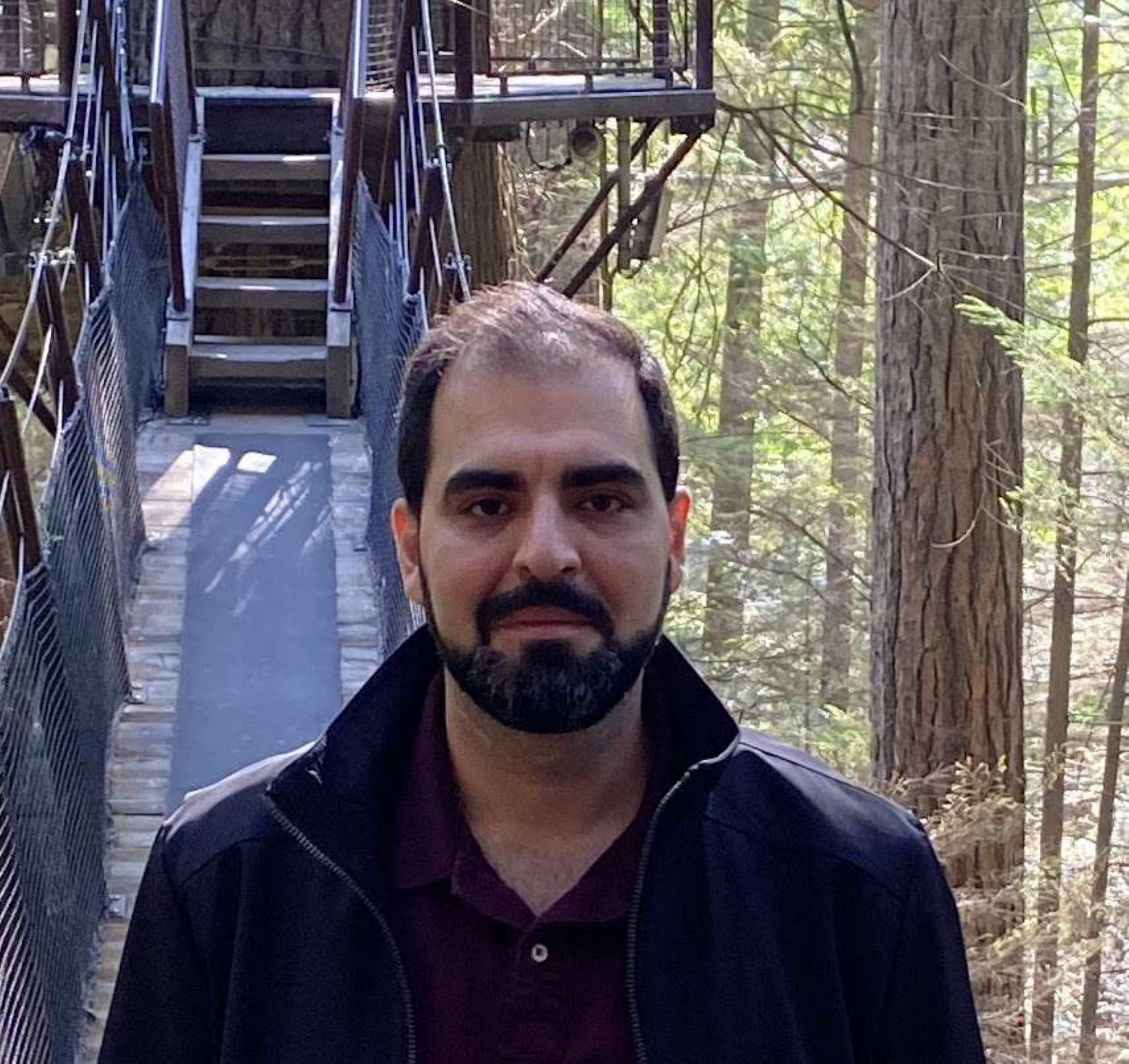}}]{Ali Etemad}
Dr. Etemad is an Associate Professor at the Department of Electrical and Computer Engineering, Queen's University. He holds an endowed professership of Mitchell Professor in AI for Human Sensing and Understanding. He leads the Ambient Intelligence and Interactive Machines (Aiim) lab. He received his M.A.Sc. and Ph.D. degrees in Electrical and Computer Engineering from Carleton University, Ottawa, Canada, in 2009 and 2014, respectively.  His main areas of research are machine learning and deep learning focused on human-centered applications with wearables, smart devices, and smart environments. Prior to joining Queen’s, he held several industrial positions as lead scientist. He has published over 160 papers in top venues in the area, is a co-inventor of 10 patents, and has given over 25 invited talks at different venues.  Dr. Etemad is an Associate Editor for IEEE Transactions on Affective Computing and IEEE Transactions on Artificial Intelligence. He has served as a PC member/reviewer, and has held organizing roles at various venues. He has received a number of awards including Supervisor of the Year Award (at Queen’s), Instructor of the Year Award (at Queen’s), and several Best Paper Awards (e.g., at ACM ICMI'23). Dr. Etemad’s lab and research program have been funded by the Natural Sciences and Engineering Research Council (NSERC) of Canada, Ontario Centers of Excellence (OCE), Canadian Foundation for Innovation (CFI), Mitacs, and other organizations, as well as the private sector.
\end{IEEEbiography}



\end{document}